\documentclass[11pt]{article}

\usepackage{blindtext}

\usepackage[nohyperref]{jmlr2e}
\usepackage[american]{babel}

\usepackage{natbib} 
\usepackage{mathtools} 
\usepackage{booktabs} 
\usepackage{tikz,ifthen} 
\usepackage{amsmath, eufrak}
\usepackage{algorithm}
\usepackage{algpseudocode}
\usepackage{bm}
\usepackage{cancel}
\usepackage{cleveref}
\usepackage[mathcal]{eucal}
\usepackage{subfigure}
\usepackage{graphicx}
\usepackage{xr}

\makeatletter
\let\oldr@@t\r@@t
\def\r@@t#1#2{%
\setbox0=\hbox{$\oldr@@t#1{#2\,}$}\dimen0=\ht0
\advance\dimen0-0.2\ht0
\setbox2=\hbox{\vrule height\ht0 depth -\dimen0}%
{\box0\lower0.4pt\box2}}
\makeatother
\usepackage{letltxmacro}
\LetLtxMacro{\oldsqrt}{\sqrt}
\renewcommand*{\sqrt}[2][\ ]{\oldsqrt[#1]{#2}}

\DeclareMathOperator*{\argmin}{argmin}

\newcommand{\setofnodes}{\mathbf{X}}

\newcommand{\setofedges}{\mathbf{E}}
\newcommand{\alphabet}{\mathcal{X}}
\newcommand{\RV}[1]{X_{#1}}

\newcommand{\nbhi}{\mathcal{N}(i)}

\newcommand{\msg}[4][]{\ifthenelse{\equal{#4}{}} {\mu^{(#1)}_{#2 #3}(x_{#3})} {\mu^{(#1)}_{#2 #3}(\RV{#3}=#4)}}
\newcommand{\msgShort}[4][]{\ifthenelse{\equal{#4}{}} {\mu^{#1}_{#2 #3}(x_{#3})} {\mu^{#1}_{#2 #3}(#4)}}

\newcommand{\mse}[1][]{\text{MSE}}

\newcommand{\FG}{\mathcal{F}}

\newcommand{\ent}{\mathcal{S}}

\newcommand{\FB}{\mathcal{F}_B}

\newcommand{\SB}{\mathcal{S}_{{B}}}

\newcommand{\polytopeLocal}{\mathbb{L}}

\newcommand{\Bbox}{\mathbb{B}}
\newcommand{\qvec}{\bm{q}}
\newcommand{\qi}{q_i}
\newcommand{\qj}{q_j}

\newcommand{\thi}{\theta_i}

\newcommand{\Jij}{J_{ij}}
\newcommand{\Qij}{Q_{ij}}
\newcommand{\aij}{\alpha_{ij}}

\newcommand{\xij}{\xi_{ij}}
\newcommand{\xijopt}{\xi_{ij}^{\ast}}

\newcommand{\cij}{c_{ij}}
\newcommand{\Fc}{\mathcal{F}_{\bm{c}}}
\newcommand{\Fzeta}{\mathcal{F}_{\bm{\zeta}}}
\newcommand{\zij}{\zeta_{ij}}
\newcommand{\zi}{\zeta_{i}}

\usepackage{lastpage}

\ShortHeadings{Adaptive Variational Inference in Probabilistic Graphical Models}{Leisenberger and Pernkopf}
\firstpageno{1}

\begin{document}

\title{Adaptive Variational Inference in \\ Probabilistic Graphical Models: Beyond Bethe, Tree-Reweighted, and Convex Free Energies} 

\author{\name Harald Leisenberger \email harald.leisenberger@tugraz.at \\
       \addr Signal Processing and Speech Communication Laboratory\\
       Graz University of Technology\\
       Graz, Austria
       \AND
       \name Franz Pernkopf \email pernkopf@tugraz.at \\
       \addr Signal Processing and Speech Communication Laboratory\\
       Graz University of Technology\\
       Graz, Austria}

\editor{My editor.}

\maketitle

\begin{abstract}
Variational inference in probabilistic graphi\-cal models aims to approximate fundamental quantities such as marginal distributions and the partition function. Popular approaches are the Bethe approximation, tree-reweighted, and other types of convex free energies. These approximations are efficient but can fail if the model is complex and highly interactive. In this work, we ana\-lyze two classes of approximations that include the above methods as special cases: first, if the model parameters are changed; and second, if the entropy approxi\-mation is changed. We discuss benefits and drawbacks of either approach, and deduce from this analysis how a free energy approximation should ideally be constructed. Based on our observations, we propose approximations that automatically adapt to a given model and demonstrate their effectiveness for a range of difficult problems.
\end{abstract}


\section{INTRODUCTION} \label{sec:introduction}

The computation of the partition function and marginals is fundamental for probabilistic inference in graphical models~\citep{koller2009graphical}. As these problems are NP-hard~\citep{valiant1979permanent,cooper1990computational,dagum1993approximate}, one must -- except for special cases -- rely on approximation methods. \\

Variational free energies
provide a deterministic framework for approximate inference~\citep{yedidia2005constructing}.
One states an auxiliary optimization problem and uses its solutions to estimate the quantities of interest. However, the approxi\-mation accuracy depends on the auxiliary objective whose specific choice is known to be challenging~\citep{wainwright2008exponential}. \\

In this work, we analyze \emph{pairwise} free energy approximations that allow for an efficient optimization. Starting from the Bethe approximation~\citep{mooij2005bethe,weller2014understanding}, we focus on two particular generalizations and analyze their properties by systematically varying their governing parameters. \\

The first generalization, denoted by $\Fc$, modifies the Bethe entropy by weighting (or \emph{counting}) its individual statistics differently. This class includes the tree-reweighted free energies and least-squares convex approximations~\citep{wainwright2005logpartition,hazan2008convergent}. We aim to understand the effects on the approximation accuracy when varing the individual entropy counting numbers $\bm{c} = \{\cij, c_i\}$. \\

The second generalization, denoted by $\Fzeta$, leaves the Bethe entropy unchanged but alters the state energy; i.e., it approximates the model. This class includes self-guided belief propagation and edge deletion methods~\citep{leisenberger2022fixing,knoll2023selfguided}. We analyze their approximation properties if the pairwise model potentials $\Jij$ are scaled by factors $\zeta_{ij}$. \\

In our analysis we make several insightful observations:
\begin{itemize}
 \item In attractive models, convex energies accurately approximate the marginals. Furthermore, the class $\Fc$ can slightly enhance the estimated partition function if $\cij$ is decreased to a certain level.
 \item In mixed models, the class $\Fc$ can drastically improve on estimating the partition function and slightly on estimating pairwise marginals, if we increase the pairwise counting numbers $\cij$. For estimating singleton marginals, either $\cij$ (in class $\Fc$) or $\zij$ (in class $\Fzeta$) should be decreased.
\end{itemize}

As practical conclusion, we propose two \emph{adaptive} free energy approximations that automatically adapt to a model: one of class $\Fzeta$ for attractive mo\-dels (\texttt{ADAPT}-$\zeta$), and one of class $\Fc$ for mixed mo\-dels (\texttt{ADAPT}-$c$). We show by experiments that \texttt{ADAPT}-$\zeta$ is superior in estimating singleton marginals in densely connected models; and that \texttt{ADAPT}-$c$ improves the estimated partition function by several orders of magnitude. \\

This work is structured as follows: Sec.~\ref{sec:background} summarizes background on variational inference in graphical mo\-dels. In Sec.~\ref{sec:adaptive_free_energies} we evaluate approximations of class $\Fc$ and $\Fzeta$, and introduce our algorithms. We present our experiments in Sec.~\ref{sec:experiments}, and conclude our work in Sec.~\ref{sec:conclusion}.

\section{BACKGROUND} \label{sec:background}

In this section we introduce the relevant background of this work. This includes probabilistic graphical models (Sec.~\ref{sec:PGMs}), the variational Gibbs free energy (Sec.~\ref{sec:Gibbs}), and Bethe and related approximations (Sec.~\ref{sec:Bethe}). We also summarize additional related work (Sec.~\ref{sec:related}).

\subsection{Probabilistic Graphical Models} \label{sec:PGMs}

Let  $\setofnodes=\{X_1, \dots, X_N\}$ be a set of binary random variables taking values in $\alphabet=\{+1,-1\}$. Let their joint distribution $p(\bm{x})$ be modeled by an undirected graph $\mathbf{G} = (\setofnodes,\setofedges)$, whose nodes are a one-to-one representation of the variables $\setofnodes$ and whose edges $(i,j) \in \setofedges$ represent all pairs of interacting variables $(X_i,X_j)$. More specifically, we assume that $p(\bm{x})$ has the form
\begin{align} \label{eq:joint_distribution}
 p(\bm{x}) = \frac{1}{Z} e^{-E(\bm{x})},
\end{align}
where $Z$ is the \emph{partition function} (i.e., the normali\-zation constant), and $E(\bm{x})$ is the \emph{state energy} corresponding to a joint realization $\bm{x} = (x_1,\dots, x_N)$. We assume that $E(\bm{x})$ has an \emph{Ising} parameterization\footnote{The Ising model has first been analyzed by \citet{ising1925ising}. In~\citet{eaton2013modelreduction,johnson2016ising} it is analyzed how to transform other models (e.g., factor graphs, multi-state models) to an Ising model.}
\begin{align} \label{eq:state_energy}
 E(\bm{x}) = - \! \! \sum\limits_{(i,j) \in \setofedges} \! \! \Jij x_i x_j- \sum\limits_{i \in \setofnodes} \thi x_i
\end{align}
with the \emph{pairwise potentials} $\Jij$ describing the correlations between connected variables $(i,j)$, and the \emph{local potentials} $\theta_i$ biasing the states of individual variables towards $+1$ if $\theta_i > 0$, or towards $-1$ if $\theta_j < 0$. An edge connecting two variables is called \emph{attractive} if $\Jij >0$ (the variables tend to share the same state), and \emph{repulsive} if $\Jij < 0$ (the variables tend to have opposite states). If a model includes only attractive edges we call it an \emph{attractive model}; if it includes both types of edges we call it a \emph{mixed model}. Finally, we denote by $\nbhi$ the set of all nodes that are connected to node $i$, and by $d_i \coloneqq |\nbhi|$ the \emph{degree} of node $i$. \\

We consider the following fundamental problems:
\begin{enumerate}
	\item[\textbf{(P1)}] The computation of the partition function:
	\begin{align} \label{eq:partition_function}
		Z = \sum\limits_{\bm{x} \in \alphabet^N} e^{- E(\bm{x})}
	\end{align}
	\item[\textbf{(P2)}] The computation of marginal probabilities, primarily of single variables and pairs of variables:
	\begin{align}
		p_i(x_i) = & \sum\limits_{\bm{x}^{\prime} \in \alphabet^N: x_i^{\prime} = x_i} p(\bm{x}^{\prime}) \label{eq:singleton_marginals} \\
		p_{ij}(x_i,x_j) =  & \sum\limits_{\bm{x}^{\prime} \in \alphabet^N: x_i^{\prime} = x_i, x_j^{\prime} = x_j} p(\bm{x}^{\prime}) \label{eq:pairwise_marginals}
	\end{align}
\end{enumerate}

These problems cannot be generally solved efficiently -- except for tree-structured graphs~\citep{pearl1988reasoning} -- and thus require approximation methods.

\subsection{Variational Gibbs Free Energy and Bethe Approximation} \label{sec:Gibbs}

One large class of approximation methods relies on variational inference~\citep{jordan1999variational}. The idea is to convert the inference problems \textbf{(P1)}, \textbf{(P2)} into an optimization problem which, however, will be intractable too. Hence, to reduce the computational complexity, one constructs an auxiliary objective that is easier to optimize. Its solutions are then used to estimate the solutions to the inference problems.

Let $q(\bm{x})$ be any 'trial' distribution over $\mathcal{X}^N$ and let the (variational) \emph{Gibbs free energy} be defined as
\begin{align} \label{eq:Gibbs_energy}
\FG(q) = \mathbb{E}_q(E(\bm{x})) - \ent(q)
\end{align}
where $\mathbb{E}_q(E(\bm{x})) = \sum\limits_{\bm{x} \in \mathcal{X}^N} q(\bm{x}) E(\bm{x}) $ is the \emph{average energy} and $\ent(q) = - \sum\limits_{\bm{x} \in \mathcal{X}^N} q(\bm{x}) \log q(\bm{x})$ is the \emph{entropy}. It has been shown that the functional $\FG(q)$ is convex and has a unique mini\-mum for $q = p$, i.e., the true distribution~\eqref{eq:joint_distribution}~\citep{wainwright2008exponential,mezard2009information}. The associated functional value is the negative log-partition function $\FG(p) = - \log Z$. Note that the evaluation of $\FG(p)$ is intractable, as it requires a summation of $2^N$ terms (in the entropy). \\

\begin{figure}[h]
\hspace{3.5cm} \includegraphics[width=0.5\linewidth]{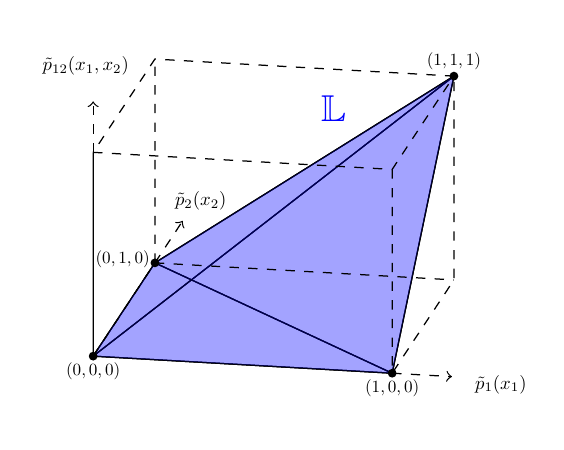}
\caption{Local polytope $\polytopeLocal$ as the linearly constrained domain of the Bethe free energy $\FB$ (simplified illustration on a single-edge graph and two nodes $X_1,X_2$).}
\label{fig:Local_Polytope}
\end{figure}

In the variational framework, $\FG(q)$ is approximated by a simpler objective. The most popular approach is the \emph{Bethe approximation} that makes two relaxations: first, it relaxes the space of feasible distributions $q$ to the space $\polytopeLocal$ of 'pseudo-marginals' $\tilde{p}_i, \tilde{p}_{ij}$ that 
must only satisfy local instead of global probabili\-ty constraints. More precisely, we define $\polytopeLocal$ as the set
\begin{align} \label{eq:local_polytope}
  \begin{split}
  & \polytopeLocal = \{\tilde{p}_i(x_i), \tilde{p}_{ij}(x_i,x_j) \! \in \! (0,1) \!: \! \!  \sum\limits_{x_j \in \alphabet} \! \tilde{p}_{ij}(x_i,x_j) = \tilde{p}_i(x_i), \\ & \hspace{-0.1cm} \sum\limits_{x_i,x_j \in \alphabet} \! \! \! \tilde{p}_{ij}(x_i,x_j) = 1, \sum\limits_{x_i \in \alphabet} \tilde{p}_i(x_i) = 1, (i,j) \in \setofedges, i \in \setofnodes \},
  \end{split}
\end{align}
and call it the \emph{local polytope} (Fig.~\ref{fig:Local_Polytope}). Second, it approximates the entropy $\ent(p)$ by the \emph{Bethe entropy} $\SB$ which is a weighted sum of local entropies $\mathcal{S}_i(\tilde{p}_i) = -\sum\limits_{x_i \in \alphabet} \tilde{p}_{i}(x_i) \log \tilde{p}_{i}(x_i)$ and pairwise entropies $\mathcal{S}_{ij}(\tilde{p}_{ij}) = -\sum\limits_{x_i,x_j \in \alphabet^{2}} \tilde{p}_{ij}(x_i,x_j) \log \tilde{p}_{ij}(x_i,x_j)$, given by
\begin{align} \label{eq:Bethe_entropy}
\SB =   \sum_{(i,j) \in \setofedges} \mathcal{S}_{ij}  - \sum_{i \in \setofnodes} \,  (d_i - 1)\mathcal{S}_i.
\end{align}
More precisely, the Bethe free energy is defined as
\begin{align} \label{eq:Bethe_free_energy}
\FB=  \mathbb{E}_{\tilde{p}_{i},\tilde{p}_{ij}}[E(\bm{x})] - \SB
\end{align}
where the Bethe average energy $\mathbb{E}_{\tilde{p}_{i},\tilde{p}_{ij}}[E(\bm{x})]$ is identical to the average energy in~\eqref{eq:Gibbs_energy} but extended to the local polytope $\polytopeLocal$, while the Bethe entropy $\SB$ is only an approximation to the 'true' entropy $\ent(p)$.
By minimizing the Bethe free energy over $\polytopeLocal$, one can estimate the partition function and marginals according to
\begin{align}
	\hspace{-0.3cm} \min\limits_{\tilde{p}_i, \tilde{p}_{ij} \, \in \, \polytopeLocal} \color{black} \FB \, \, & \approx \, \, - \log Z \quad \quad \text{and} \label{eq:Bethe_approx_partition} \\
	\hspace{-0.3cm} \, \,  \argmin\limits_{\tilde{p}_i, \tilde{p}_{ij} \, \in \, \polytopeLocal} \color{black} \FB \, \, & \approx \, \, \{ \, p_i,p_{ij} \, \, | \, \, (i,j) \in \setofedges \, \, \, \text{and} \, \, \, i \in \setofnodes \}. \label{eq:Bethe_approx_marginals}
\end{align}
The approximations~\eqref{eq:Bethe_approx_partition},~\eqref{eq:Bethe_approx_marginals} are correct if the graph is a tree; i.e., the Bethe entropy approximation is exact in that case~\citep{yedidia2005constructing}.  However, itsquality often degrades with a high connectivity and strong correlations between variables~\citep{meshi2009convexifying, weller2014understanding,leisenberger2024reliability}.
\vspace{-0.3cm}
\subsection{Related Free Energy Approximations} \label{sec:Bethe}
As the Bethe approximation can fail in certain models, various attempts were made to construct alternative types of free energy approximations. In this work we consider the so-called \emph{pairwise approximations}, which are generalizations of the Bethe free energy that preserve two of its favorable properties: first, they include statistics that are defined over at most two variables\footnote{E.g., the Bethe entropy is a sum of statistics that involve either one or two variables. In contrast, the 'true' entropy involves all variables of the model.}; second, they are defined on the local polytope $\polytopeLocal$ and thus bounded by a relatively small number of linear constraints\footnote{More precisely, the convex set $\polytopeLocal$ is bounded by $\mathcal{O}(|\setofnodes| + |\setofedges|)$ linear constraints~\citep{wainwright2008exponential}}. This enables an efficient optimization of the objective function by methods of constrained numerical optimization and using its minima to estimate the partition function and marginals. Among all pairwise approximations we focus on two specific classes: those which make a different entropy approximation than $\FB$ but leave the average energy $\mathbb{E}_{\tilde{p}_{i},\tilde{p}_{ij}}[E(\bm{x})]$ unchanged (Sec.~\ref{subsec:Entropy_approx}); and those which keep the Bethe entropy~\eqref{eq:Bethe_entropy} but modify the energy $E(\bm{x})$ in~\eqref{eq:state_energy} or, in other words, the model parameters $\Jij$, $\theta_i$ (Sec.~\ref{subsec:Energy_approx}).

\subsubsection{Generalizing the Bethe Entropy} \label{subsec:Entropy_approx}
While the Gibbs free energy~\eqref{eq:Gibbs_energy} is convex on its domain, this is not guaranteed for the Bethe free energy; in fact, $\FB$ is convex on $\polytopeLocal$ if and only if the graph contains at most one loop\footnote{A loop -- or cycle -- is a closed walk (i.e., an edge sequence) in the graph that includes any edge at most once.}~\citep{heskes2004uniqueness,watanabe2009graphzeta}. This is because the Bethe entropy $\SB$ is generally not concave which has sometimes been considered the reason why the Bethe approximation fails in loopy graphs~\citep{wainwright2005logpartition,heskes2006convexity}. Hence, alternative approximations use a provably convex entropy approximation and are thus convex on $\polytopeLocal$. \\ 

Let $\bm{c} \coloneqq \{\cij, c_i \,  |  \, (i,j) \in \setofedges, i \in \setofnodes\}$ be a set of \emph{counting numbers} with one pairwise number $\cij$ for each edge and one local number $c_i$ for each node. We define an $\Fc$ - approximation to the Gibbs free energy as
\begin{align} \label{eq:c_approximation}
 \Fc = \mathbb{E}_{\tilde{p}_{i},\tilde{p}_{ij}}[E(\bm{x})] - \tilde{\ent}_{\bm{c}}
\end{align}
with the average energy unmodified as in~\eqref{eq:Bethe_free_energy} and a generalized entropy approximation of the form
\begin{align} \label{eq:c_entropy}
 \tilde{\ent}_{\bm{c}} = \sum_{(i,j) \in \setofedges} \cij \mathcal{S}_{ij}  + \sum_{i \in \setofnodes} \,  c_i \mathcal{S}_i.
\end{align}
Note that, for $\cij = 1$ and $c_i = 1-d_i$ we reobtain the Bethe entropy~\eqref{eq:Bethe_entropy}. The generalization~\eqref{eq:c_entropy} creates additional freedom in choosing the entropy approxi\-mation by weighting the pairwise and local entropies differently than in the Bethe approximation. A result of~\citet{heskes2006convexity} says that $\tilde{\ent}_{\bm{c}}$ is convex if, for all $i \in \setofnodes$ and $(i,j) \in \setofedges$, there exist auxi\-liary numbers $\tilde{c}_{(i,j)},\tilde{c}_{(i,j) \rightarrow i},\tilde{c}_{i} \geq 0$ such that
\begin{align} \label{eq:auxiliary_counting_numbers}
\begin{split}´
 \cij & \, = \, \tilde{c}_{(i,j)} + \sum\limits_{j \in \nbhi} \tilde{c}_{(i,j) \rightarrow i} \\
 c_i & \, = \, \tilde{c}_{i} - \tilde{c}_{(i,j) \rightarrow i} - \tilde{c}_{(i,j) \rightarrow j}
 \end{split}
\end{align}
for all $\cij, c_i \in \bm{c}$. We briefly introduce two particular approximations $\Fc$ that satisfy the properties in~\eqref{eq:auxiliary_counting_numbers} (i.e., that are convex on $\polytopeLocal$) and that we will compare to other methods in our experiments in Sec.~\ref{sec:experiments}.

\paragraph{Tree-Reweighted Free Energies.} The tree-reweighted free energies (TRW) use a weighted sum of entropies over spanning trees in the graph, and are not only convex but also provide an upper bound to the
log-partition function~\citep{wainwright2005logpartition}. One chooses a set $\mathcal{T}$ of spanning trees $T$ and a valid \emph{tree distribution} $\rho(T)$ over all $T \in \mathcal{T}$. The pairwise counting numbers $\cij$ are then computed as \emph{edge occurrence probabilities} representing the proportions how often an edge occurs in the tree set $\mathcal{T}$, weighted by $\rho(T)$. More precisely, we set $\cij = \sum\limits_{T \in \mathcal{T}} \rho(T) \, I_{(i,j)}(T)$, where $I_{(i,j)}$ indicates if an edge $(i,j)$ is contained in a tree $T$ or not. The local counting numbers are then set to $c_i = 1 - \sum\limits_{j \in \nbhi} \cij$. This ensures that the entropy of each node is in sum counted precisely once and hence the entropy approximation is exact on tree graphs. Further details (including the choice of spanning trees) are provided in~\citet{kolmogorov2006convergent, jancsary2011convergent}.

\paragraph{Least-Squares-Convex Free Energies.} The intuition behind Least-Squares (LS) Convex Free Energies proposed by~\citet{hazan2008convergent} is to keep $\Fc$ as close to the Bethe free energy as possible, but with counting numbers $\bm{c}$ that satisfy the convexity conditions~\eqref{eq:auxiliary_counting_numbers}.
Specifically, one solves the LS program
\begin{align} \label{eq:least_squares_convexity}
 \min \sum\limits_{(i,j \in \setofedges)} \big(\tilde{c}_{(i,j)} + \tilde{c}_{(i,j) \rightarrow i}  +\tilde{c}_{(i,j) \rightarrow j} -1 \big)^2
\end{align}
where the minimization is performed with respect to all nonnega\-tive auxiliary numbers \\ $\tilde{c}_{i},\tilde{c}_{(i,j)},\tilde{c}_{(i,j) \rightarrow i},\tilde{c}_{(i,j) \rightarrow j}$ (for all $i \in \setofnodes, (i,j) \in \setofedges$) satisfying the linear constraints \label{eq:least_squares_convexity_modified_constraints}
\begin{align}
\tilde{c}_{i} + \sum\limits_{j \in \nbhi} (\tilde{c}_{(i,j)} + \tilde{c}_{(i,j) \rightarrow j}) = 1, \quad i \in \setofnodes.
\end{align}
Afterwards, one computes $\cij$ and $c_i$ according to~\eqref{eq:auxiliary_counting_numbers}. Other ways to set the counting numbers were, e.g., proposed by~\citet{wiegerinck2002fractional, globerson2007b, meshi2009convexifying}. These approaches are mostly inferior to the above methods.

\subsubsection{Changing the State Energy} \label{subsec:Energy_approx}
The approximations $\Fc$ introduced in Sec.~\ref{subsec:Entropy_approx} share the favorable property that the average energy is computed as in the Gibbs free energy~\eqref{eq:Gibbs_energy} (but extended to $\polytopeLocal$, which may still alter the location of the minimum). Recently, an alternative class of approximations were proposed that leave the entropy approximation as in the Bethe free energy but modify the state energy~\eqref{eq:state_energy} by scaling the model parameters. This may seem unintuitive as now both aspects of the approximation -- the average energy and the entropy -- are incorrect. However, some experimental results show improvements over the Bethe approximation~\citep{knoll2023selfguided}. \\

Let $\bm{\zeta} \coloneqq \{\zij, \zi \,  |  \, (i,j) \in \setofedges, i \in \setofnodes\}$ be a set of \emph{scale factors} with one pairwise scale factor $\zij$ for each edge and one local scale factor $\zi$ for each node. We define an $\Fzeta$ - approximation to the Gibbs free energy as
\begin{align} \label{eq:zeta_approximation}
 \Fzeta =  \mathbb{E}_{\tilde{p}_{i},\tilde{p}_{ij}}[\tilde{E}_{\bm{\zeta}}(\bm{x})] - \SB
\end{align}
with a
$\bm{\zeta}$-scaled state energy of the form
\begin{align} \label{eq:zeta_energy}
 \tilde{E}_{\bm{\zeta}}(\bm{x}) = - \! \! \sum\limits_{(i,j) \in \setofedges} \! \! (\zij \Jij) x_i x_j- \sum\limits_{i \in \setofnodes} (\zi \thi) x_i
\end{align}
and the Bethe entropy $\SB$ from~\eqref{eq:Bethe_entropy}. If all factors $\zij, \zi$ are set to one we reobtain the Bethe approximation. \\

The idea behind this approach is as follows: Strong correlations between variables (represented by high values of $|\Jij|$) can have a detrimental influence on the accuracy of the Bethe approximation~\citep{weller2014understanding,knoll2019accurate}. By decrea\-sing $|\Jij|$, the approximation becomes more reliable -- however, with respect to a different model. Still, the error induced by $\Fzeta$ with res\-pect to the approximated model is often smaller than the error induced by $\FB$ with respect to the original model. In~\citet{leisenberger2024reliability}, this behavior has been explained by the fact that $\Fzeta$ becomes convex on a submanifold of $\polytopeLocal$ if the correlations are sufficiently weakened. There are only few attempts in the literature following this approach. Here we introduce one of them, that we will later use as comparison method in our experiments (Sec.~\ref{sec:experiments}).

\paragraph{Self-Guided Belief Propagation.} Self-Guided Belief Propagation (SBP) exploits the well known relationship between the Bethe approximation and the popular loopy belief propagation (LBP) algorithm\footnote{LBP is an iterative message passing algorithm that aims to solve a fixed point equation system. The fixed points of LBP are in one-to-one correspondence to the local minima of the Bethe free energy~\citep{yedidia2001bethe, heskes2003stable}}. It starts with a completely uncorrelated model (i.e., where all $\Jij$ are set zo zero) and successively increases the correlations $\Jij$ until LBP, which is sequentially applied during this procedure, either fails to converge for some intermediate state of the model or converges to a fixed point of the original model. In our context, this means that we successively increase a pairwise scale factor $\zij = \zeta$ (that is the same for all edges) starting from $\zeta= 0$ and setting the local scale factors $\zi$ to one (i.e., leaving the local potentials $\theta_i$ unmodified), until LBP either fails to find the minimum of $\Fzeta$ for some $\zeta \in (0,1)$ or finds the minimum of $\FB$ for $\zeta = 1$. It uses the minimum of $\Fzeta$ (or $\FB$) associated to the final state of convergence of LBP to estimate the marginals and partition function. The detailed algorithm is described in~\citep{knoll2023selfguided}. \\\

Another approximation of the class $\Fzeta$ successively deletes edges from the model; this corresponds to setting pairwise scale factors $\zij$ associated to deleted edges to zero~\citep{leisenberger2022fixing}. However, SBP proves to be the more flexible algorithm.

\subsection{Other Related Work} \label{sec:related}

\paragraph{Higher-Order Variational Inference.} Increasing the complexity of variational inference can help to achieve a higher accuracy. The exact solution can be computed by the junction-tree algorithm whose complexity increases exponentially with the size of the largest clique\footnote{A clique is a fully connected subgraph.} in the graph~\citep{lauritzen1988junctiontree}. Methods that make higher order entropy approximations were constructed by~\citet{yedidia2005constructing}. Other methods whose efficiency is located 'between' pairwise approximations and exact inference include join graph propagation~\citep{dechter2002joingraph}, loop corrections~\citep{mooij2007loopcorrections}, oriented trees~\citep{globerson2007a}, and variable clamping~\citep{weller2014b}.

\paragraph{Message Passing Algorithms.} Having its origins in statistical mechanics~\citep{bethe1935superlattices,peierls1936superlattices}, the Bethe approximation has gained popularity in the computer science and statistics community when~\citet{yedidia2001bethe} have proven its connection to loopy belief propagation. Inspired by their discovery, other researchers designed alternative free energies that are related to similar message passing algorithms~\citep{yedidia2005constructing,hazan2008convergent,meltzer2009convergent}. These are often efficient but can fail to converge to a fixed point. Also, the convergence behavior of LBP has been subject of research~\citep{tatikonda2003convergence,ihler2005message,mooij2007sufficient,leisenberger2021lyapunov}.

\paragraph{Minimizing Free Energy Approximations.} In practice, it can be difficult to minimize a certain type of free energy approximation, in particular if it is non-convex and has multiple local minima. Various me\-thods were proposed: gradient-based algorithms combined with projection steps~\citep{welling2001belief,shin2012complexity}; a double-loop algorithm that applies a concave-convex decomposition~\citep{yuille2002CCCP}; combinatorial optimization~\citep{weller2014a}; convex optimization~\citep{weller2014understanding}; and projected Quasi-Newton methods~\citep{schmidt2009quasinewton,jancsary2011convergent,leisenberger2024reliability}.


\section{ADAPTIVE FREE ENERGY APPROXIMATIONS} \label{sec:adaptive_free_energies}
In this section we analyze the two classes of pairwise free energy approximations $\Fc$ and $\Fzeta$ introduced in Sec.~\ref{subsec:Entropy_approx} and Sec.~\ref{subsec:Energy_approx} in a broader context.
Our goal is to identify regimes of the counting numbers $\bm{c}$ and scale factors $\bm{\zeta}$ that are related to accurate approximations $\Fc$ and $\Fzeta$.
For that purpose, we vary these parameters systema\-tically in appropriate intervals and, for each realization of $\bm{c}$ and $\bm{\zeta}$, evaluate the approximation accuracy induced by $\Fc$ and $\Fzeta$; that is, we compare the minima of $\Fc$ and $\Fzeta$ to the exact marginals and partition function which were computed with the junction tree algorithm~\citep{lauritzen1988junctiontree}.
For minimizing free energy approximations of class $\Fc$ or $\Fzeta$, we use techniques of constrained numerical optimization~\citep{nocedal2006numerical}. In particular, we apply a projected Quasi-Newton algorithm which ite\-ratively performs second-order parameter updates, by using an approxi\-mation to the inverse Hessian. To ensure that the iterates stay within the local polytope $\polytopeLocal$, a projection step is required. The algorithm stops when it reaches a stationary point (usually a minimum) of the free energy approximation. All details including pseudocode are contained in the Appendix (Alg. 1, named \texttt{F-MIN}). \\

For error evaluation, we measure three kinds of errors: the $\ell^1$-errors between exact and approxi\-mate singleton and pairwise marginals, and the absolute error between the exact and approximate log-partition function. In addition to the error evaluation and analysis, another aspect is of practical importance: do algorithms like Bethe, TRW, LS-Convex, and SBP capture favorable regimes of $\bm{c}$ and $\bm{\zeta}$ well, or do there exist optimal settings of $\bm{c}$ or $\bm{\zeta}$ that are not related to any of the 'exis\-ting' free energy approximations? %
We address this question by comparing the governing parameters $\bm{c}$ and $\bm{\zeta}$ associated to the baseline algorithms to optimal settings of $\bm{c}$ and $\bm{\zeta}$ (i.e., with minimum errors). \\ 

In this section we consider a complete graph on $10$ nodes that allows for tractable exact inference. Further evalu\-ations on different graphs exhibit a similar behavior and are shown in the Appendix. We ana\-lyze both attractive and mixed models (Sec.~\ref{sec:PGMs}). For either type, we consider two different scenarios regarding the pairwise potentials $\Jij$: weak correlations with all $\Jij$ being uniformly sampled\footnote{We will use the notation $\sim \mathcal{U}(a,b)$ to indicate that we sample a parameter uniformly from some interval.} from $(0,0.5)$ resp. $(-0.5,0.5)$, and strong correlations with all $\Jij$ being uniformly sampled from $(0,2)$ resp. $(-2,2)$. We consider three different scenarios regarding the local potentials $\thi$ that are uniformly sampled from $(-0.2,0.2)$, $(-0.6,0.6)$, or $(-1,1)$. For each configuration of the potentials, the results (i.e., the errors and the estimated values of $\bm{c}$ and $\bm{\zeta}$ shown as vertical lines in Fig.~\ref{fig:scaling_c}, Fig.~\ref{fig:scaling_J}) are averaged over $100$ individual models. \\


\subsection{Analysis of Class $\Fc$-Approximations} \label{sec:analysis_c_approx}
To simplify our considerations, we focus on a speci\-fic subclass of $\Fc$ where the pairwise entropies in~\eqref{eq:c_entropy} share the same counting number, i.e., $\cij = c$ for all edges. In~\citet{meshi2009convexifying} it has been argued that free energies of class $\Fc$ should be \emph{variable-valid}, i.e., that the entropy approximation is exact on a tree. This can be achieved by setting the local counting numbers
\vspace{-0.2cm}
\begin{align} \label{eq:variable-valid}
 c_i = 1 - \sum\limits_{j \in \nbhi} \cij
\end{align}
for all nodes. Then the local counting numbers are implicitly parameterized via $c$ too. For error evaluation, we vary $c$ in the interval $(0,3)$. Note that $c=1$ corresponds to the Bethe free energy $\FB$. \\

\begin{figure*}[!t]
\centering
\subfigure{\includegraphics[width=0.87\linewidth]{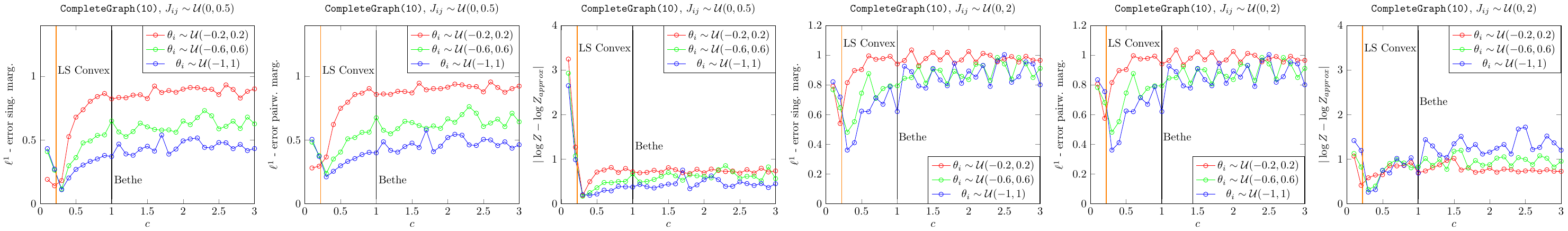}}
\subfigure{\includegraphics[width=0.87\linewidth]{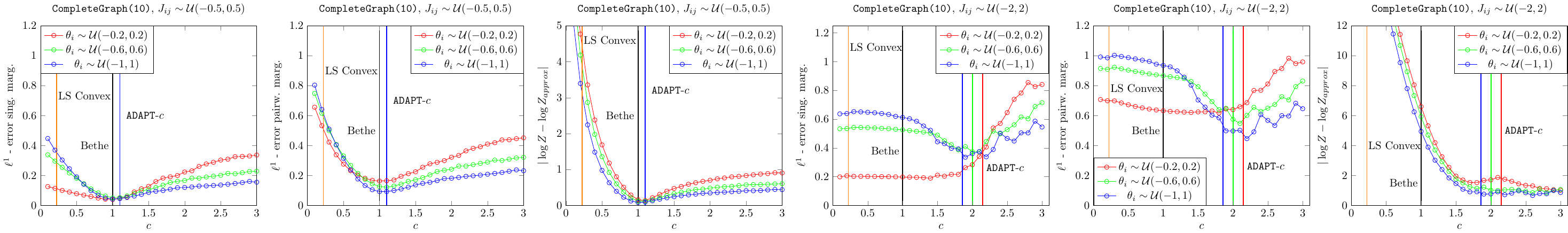}}
\caption{Approximation behavior of $\Fc$. First row: attractive models; second row: mixed models.}
\label{fig:scaling_c}
\end{figure*}

The first row in Fig.~\ref{fig:scaling_c} shows the results for attractive models. The estimates for singleton and pairwise marginals are improved if $c$ decreases. Especially the least-squares (LS) convex energy (Sec.~\ref{subsec:Entropy_approx}) finds a good regime for the counting numbers\footnote{We remark that the LS-convex free energies generally use counting numbers $\cij$ that are not uniform over all edges $(i,j)$. Yet, in the complete graph the assumption $c = \cij$ is valid (due to symmetries) which matches the considerations in this section. For the graph structures considered in the Appendix, the LS-convex energies violate the assumption $c = \cij$. Thus, for visualization of the associated vertical lines, we use an average over all $\cij$, i.e., $c = \frac{1}{|\setofedges|} \sum\limits_{(i,j) \in \setofedges} \cij$.}, as is shown by the orange vertical line, and outperforms the Bethe approximation (represented by the black vertical line); however, this advantage shrinks if $\Jij$ increases. The estimates of the log-partition function are less sensitive to changes in the counting numbers, unless they become too small. In this regard, the LS-convex free energy appears not to be a decent choice and is inferior to Bethe. However, there appears to be an optimal regime of $c$ that is located between $\FB$ and the LS-convex energy. Note that the counting numbers of TRW are often similar to those of LS-convex, and not explicitly shown in this analyses. \\

The second row in Fig.~\ref{fig:scaling_c} shows the results for mixed models. If $\Jij$ is small, $\FB$ finds almost optimal estimates for all quantities; however, its accuracy degrades if pairwise potentials become stronger in which case a minimum error is achieved if $c$ grows beyond one. Beyond a certain threshold (which also depends on the strength of the local potentials $\thi$) the marginal error increases again, while the error in the log-partition function keeps relatively stable at a low level. The red, green, and blue line (adapted to $\thi$) represent the counting numbers $c$ as estimated by our algorithm \texttt{ADAPT}-$c$ that will be introduced in Sec.~\ref{sec:adaptive_c_zeta_approximations}. As \texttt{ADAPT}-$c$ also takes the local potentials into account, three different vertical lines are shown. While, for small $\Jij$, the estimated $c$ is almost identical to the pairwise Bethe counting number (i.e., $c=1$), estimates of $c$ made by \texttt{ADAPT}-$c$ usually increase if $\Jij$ increases. This has a beneficial effect on the partition function (for all settings of $\thi$) and the pairwise marginals (for stronger $\thi$) when estimated by \texttt{ADAPT}-$c$.

\subsection{Analysis of Class $\Fzeta$-Approximations} \label{sec:analysis_zeta_approx}

Again we facilitate our analysis by only modifying the pairwise potentials $\Jij$, while setting the local scale factors $\zeta_i$ in~\eqref{eq:zeta_energy} to one for all nodes (i.e., we leave the local potentials $\thi$ unchanged). We further assume that all edges share the same pairwise scale factor, i.e., $\zij = \zeta$. This enables us to directly compare our results to self-guided belief propagation (SBP, Sec.~\ref{subsec:Energy_approx}) which makes the same assumptions. For error evalu\-ation, we vary $\zeta$ in the interval $(0,1.5)$. \\

\begin{figure*}[!t]
\centering
\subfigure{\includegraphics[width=0.87\linewidth]{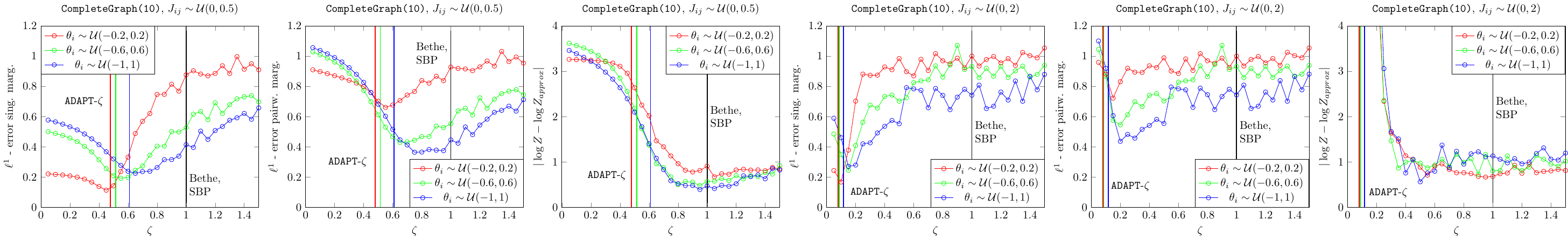}}
\subfigure{\includegraphics[width=0.87\linewidth]{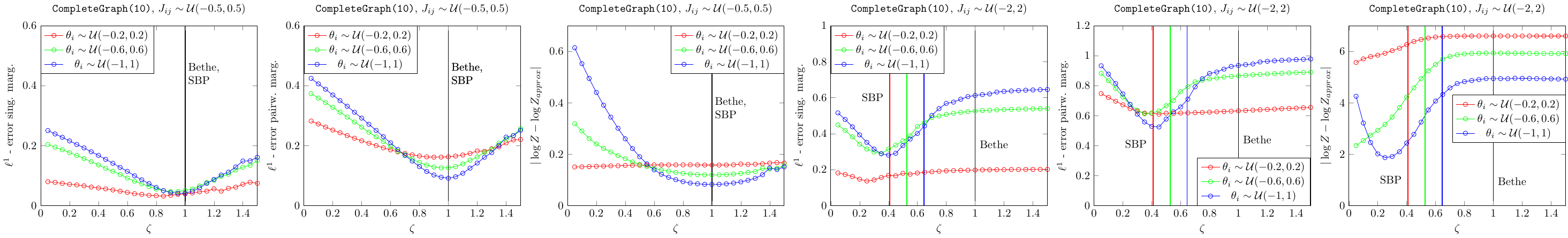}}
\caption{Approximation behavior of $\Fzeta$. First row: attractive models; second row: mixed models.}
\label{fig:scaling_J}
\end{figure*}

The first row in Fig.~\ref{fig:scaling_J} shows the results for attractive models. We observe that the singleton and pairwise marginal error develops similarly as for class $\Fc$; in particular, Bethe fails to approximate the marginals if $\Jij$ is large, while improvements can be made if $\zeta$ is decreased to a certain level. Same as the Bethe approxi\-mation, SBP mostly uses a scale factor of $\zeta=1$ in attractive models\footnote{This is because LBP mostly converges in attractive models (if the messages are updated in random order) and thus SBP continues increasing $\zeta$ until it arrives at the ori\-ginal model (i.e., $\zeta=1$). However, by implicitly selecting more accurate fixed points (i.e., minima of the Bethe free energy), it often finds better estimates of the quantities of interest than $\FB$ (as will be shown in Sec.~\ref{sec:experiments}).}.
Our algorithm \texttt{ADAPT}-$\zeta$ (introduced in Sec.~\ref{sec:adaptive_c_zeta_approximations}) aims to reduce $\zeta$ and affect a 'smoothe\-ning' of $\FB$ until it gets a relatively convex shape. This approach has a positive effect on estimating singleton marginals, while an optimal $\zeta$ for estimating pairwise marginals tends to be underestimated. Moreover, $\Fzeta$-approximations (except for $\FB$) do generally entail unstable estimates of the partition function and should be used with care for that purpose. As \texttt{ADAPT}-$\zeta$ takes the local potentials into account, three lines represent the associated estimates of $\zeta$ (red, green, and blue). \\

The second row in Fig.~\ref{fig:scaling_J} shows the results for gene\-ral models. Usually, $\FB$ outperforms other $\Fzeta$ - approximations if $\Jij$ is small (in which case it is equivalent to SBP). If $\Jij$ increases, the scale factor $\zeta$ should be decreased to a certain optimal level that depends on the local potentials $\thi$. For all quantities of interest, SBP makes clear improvements over $\FB$ in mixed models (drawn by the red, green, and blue line adapted to $\thi$). However, there is some more room for further improvement over SBP regarding the choice of an optimal $\zeta$ (particularly for strong pairwise potentials $\Jij$).

\subsection{Adaptive $\Fc$ - and $\Fzeta$ - approximations} \label{sec:adaptive_c_zeta_approximations}
This subsection includes a conceptional description of our two algorithms \texttt{ADAPT}-$c$ and \texttt{ADAPT}-$\zeta$ that we propose and evaluate in this work. Detailed pseudocodes are contained in the Appendix (Algorithms 3 and 4).

\subsection{Attractive Models: \texttt{ADAPT}-$\zeta$}

For attractive models, we propose an adaptive algorithm of class $\Fzeta$, named \texttt{ADAPT}-$\zeta$.
In these models, the Bethe approximation regularly fails to approximate the quantities of interest; in particular, the estimated marginals are often highly inaccurate if the pairwise potentials are strong.
This behavior has been explained by the non-convexity of the Bethe entropy (and thus the Bethe free energy) and is shared by other free energy approximations that are not provably convex~\citep{wainwright2008exponential,weller2014understanding,leisenberger2024reliability}.
More precisely, the Bethe entropy develops multiple minima which 'pushes' the minima of $\FB$ towards extreme values of the pseudomarginals.
This effect is related to the 'overcounting' of information and thus overconfidence caused by loopy belief propagation in graphs with multiple cycles~\citep{weiss2000correctness,ihler2005message}. \\

Our algorithm \texttt{ADAPT}-$\zeta$ aims to compensate for this unwanted behavior by using a 'close-to-convex' version of $\FB$ within the class $\Fzeta$.
More precisely, it continuously modifies the Bethe free energy by reducing the pairwise potentials $\Jij$ in small steps (or, in other words, the joint pairwise scale factor $\zeta$, starting from $\zeta=1$) until $\Fzeta$ becomes convex or all except one minimum disappears (which we interpret as 'close-to-convexity')\footnote{Usually close-to-convexity is achieved for larger $\zeta$ (and thus earlier) than convexity; in some cases, however, these two properties are equivalent~\citep{mooij2005bethe}.}.
Formally, it searches for the largest $\zeta$ such that $\Fzeta$ has a unique minimum, i.e.,
\begin{align} \label{eq:6_optimal_zeta_scaling}
\begin{split}
 & \zeta^{\ast} = \, \max\limits_{(0,1]} \, \zeta , \\ & \text{s.t.} \, \, \, \Fzeta \, \, \,  \text{has a unique minimum.}
\end{split}
\end{align}
Note that we cannot solve problem~\eqref{eq:6_optimal_zeta_scaling} exactly as there does not exist a practicable result that is equiva\-lent to uniqueness of a minimum.
Thus, to verify that $\Fzeta$ has a unique minimum, we use a sufficient condition of~\citet{mooij2007sufficient} which is based on the spectral radius of the LBP Jacobian (i.e., we usually end up with a lower bound on $\zeta^{\ast}$).\footnote{Note that there always exists a positive scale factor $\zeta$ such that $\Fzeta$ has a unique minimum~\citep{leisenberger2024reliability}; thus, the described procedure is well defined.}
After termination, \texttt{ADAPT}-$\zeta$ requires a single application of \texttt{F-MIN} (Alg.~\ref{algo:projected_quasi_Newton_FMIN} in Appendix~\ref{appendix_pseudocodes}) to minimize the associated free energy approximation $\Fzeta$.

\subsubsection{Mixed Models: \texttt{ADAPT}-$c$}

For mixed models, we propose an adaptive algorithm of class $\Fc$, named \texttt{ADAPT}-$c$, whose design is motivated by the analyses from Sec.~\ref{sec:analysis_c_approx}. In Fig.~\ref{fig:scaling_c} (second row) we have observed that especially the partition function but also the marginals estimated by $\Fc$ become more accurate if $c$ increases beyond one; in particular, there appears to be a certain regime of $c$ in which all errors roughly attain a minimum (with the marginal errors increasing again beyond that point). Note that most existing algorithms of class $\Fc$ such as TRW or LS-convex use counting numbers that are smaller than one to make the free energy approximation convex; however, convexity is sometimes not considered a favorable property in mixed mo\-dels~\citep{weller2014understanding}. \\

The idea of \texttt{ADAPT}-$c$ is to estimate the threshold of $c$ beyond which only small changes in the partition function error occur. The associated free energy approximation $\Fc$ will then also be used to estimate the marginals. Let $\Delta c$ be an incremental change in $c$ and let $c_{tol}$ be some value that defines to what extent changes in the estimated log-partition function (i.e., $- \min \Fc$) are tolerable if $c$ is increased by $\Delta c$. Then we are formally interested in the solution of the following optimization problem:
\begin{align} \label{eq:optimal_c}
\begin{split}
 & c^{\ast} = \, \min\limits_{[1,\infty)} \, c , \\ & \text{s.t.} \, \, \, | \min \, \mathcal{F}_{c + \Delta c} \, - \, \min \, \mathcal{F}_{c}  \, | \, < \, c_{tol}
\end{split}
\end{align}
Note that the solution of~\eqref{eq:optimal_c} depends on the choice of $\Delta c$ and $c_{tol}$. Also, there is no theroretical guarantee on the existence of $c^{\ast}$ for any choice of these two parameters; however, we expect that the existence of a real-valued solution becomes more likely if $c_{tol}$ increases (while $\Delta c$ is kept at a constant level).
Thus, \texttt{ADAPT}-$c$ successively increases $c$ in small steps (starting from one, i.e., from the Bethe approximation), and minimizes the associated free energy approximation $\Fc$ until there are no significant changes in the estimated log-partition function anymore. For each realization of $c$, algorithm \texttt{F-MIN} is applied to find a minimum of $\Fc$. Fortunately, \texttt{F-MIN} usually converges quickly for most mixed models. Although this procedure is not guaranteed to converge\footnote{More precisely, we may both over- and underestimate the solution of~\eqref{eq:optimal_c} (if it exists), because as there is no algorithm that reliably finds the global minimum of $\Fc$.} to the solution $c^{\ast} $ of~\eqref{eq:optimal_c}, we show in our experiments (Sec.~\ref{sec:experiments}) that the results produced by \texttt{ADAPT}-$c$ are fairly promising.


\section{EXPERIMENTS} \label{sec:experiments}
In this section we evaluate and compare
the algorithms
that were discussed in this work. This includes the Bethe approximation ($\FB$), the tree-reweighted (TRW) free energies, the least-squares-convex (LS-convex) free energies, self-guided belief propagation (SBP), and the two adaptive algorithms \texttt{ADAPT}-$\zeta$ (for attractive mo\-dels) and \texttt{ADAPT}-$c$ (for mixed models) introduced and explained in Sec.~\ref{sec:adaptive_free_energies}. Our experimental setup is similar as in Sec.~\ref{sec:adaptive_free_energies}; however, we now vary the size $\hat{J}$ of the intervals $(0,\hat{J})$ (for attractive mo\-dels) resp. $(-\hat{J},\hat{J})$ (for mixed mo\-dels) for sampling $\Jij$ between $0$ and $3$ in steps of $0.1$ (instead of varying the counting numbers $c$ or scale factors $\zeta$). We compare the performance of the algorithms by using
the $\ell^1$-errors w.r.t. the singleton and pairwise marginals, and the absolute error w.r.t. the log-partition function. We consider the same scenari\-os regarding the strength of the local potentials: $\thi \sim \mathcal{U}(-0.2,0.2)$, $\sim \mathcal{U}(-0.6,0.6)$, and $ \sim \mathcal{U}(-1,1)$.
The experiments are performed on a complete graph on 10 vertices. Further experiments on grid graphs and Erdos-Renyi random graphs are presented in the Appendix. For each configuration of the potentials and each algorithm the results are averaged over $100$ models.

\subsection{Attractive Models} \label{subsec:adaptive_experiments_attractive}
\begin{figure*}[h!]
\centering
\subfigure{\includegraphics[width=0.53\linewidth]{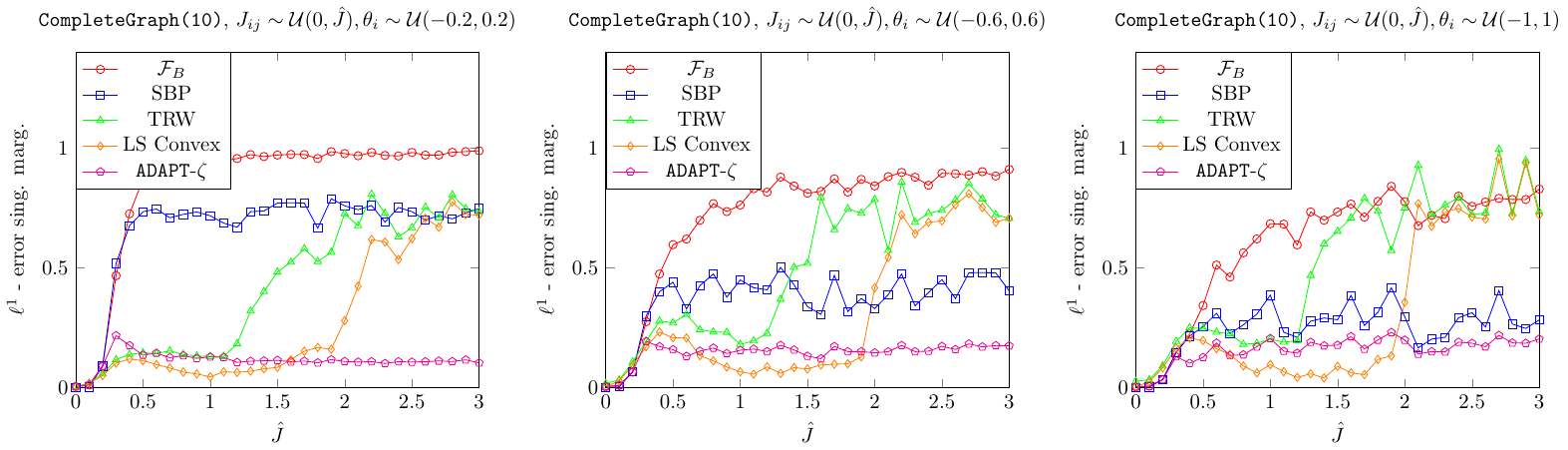}} \\
\subfigure{\includegraphics[width=0.53\linewidth]{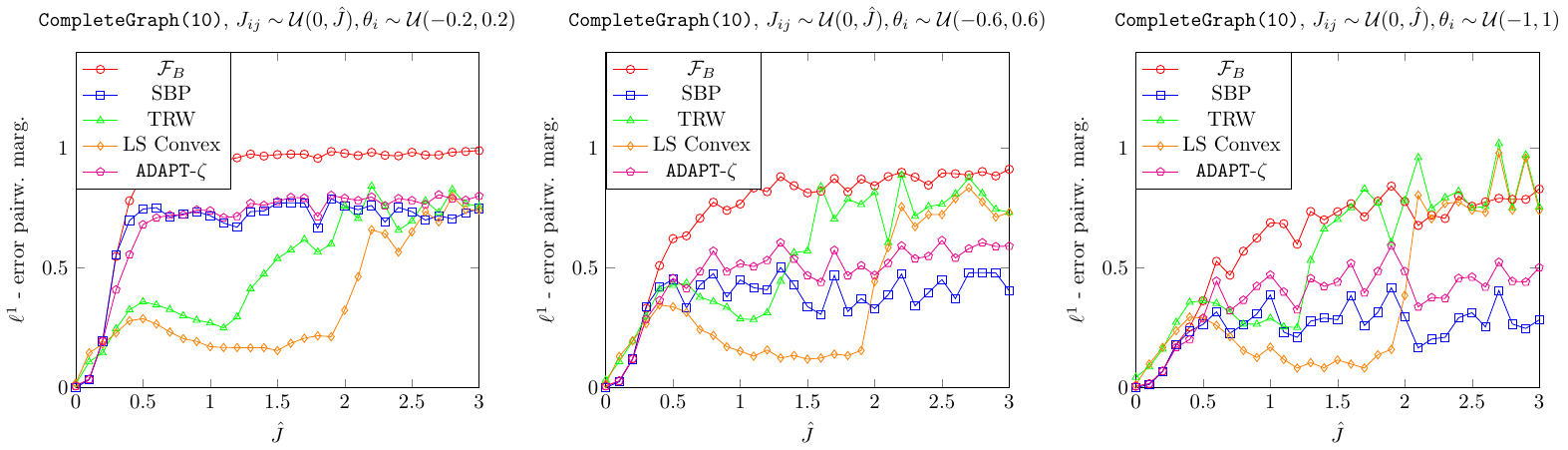}} \\
\subfigure{\includegraphics[width=0.53\linewidth]{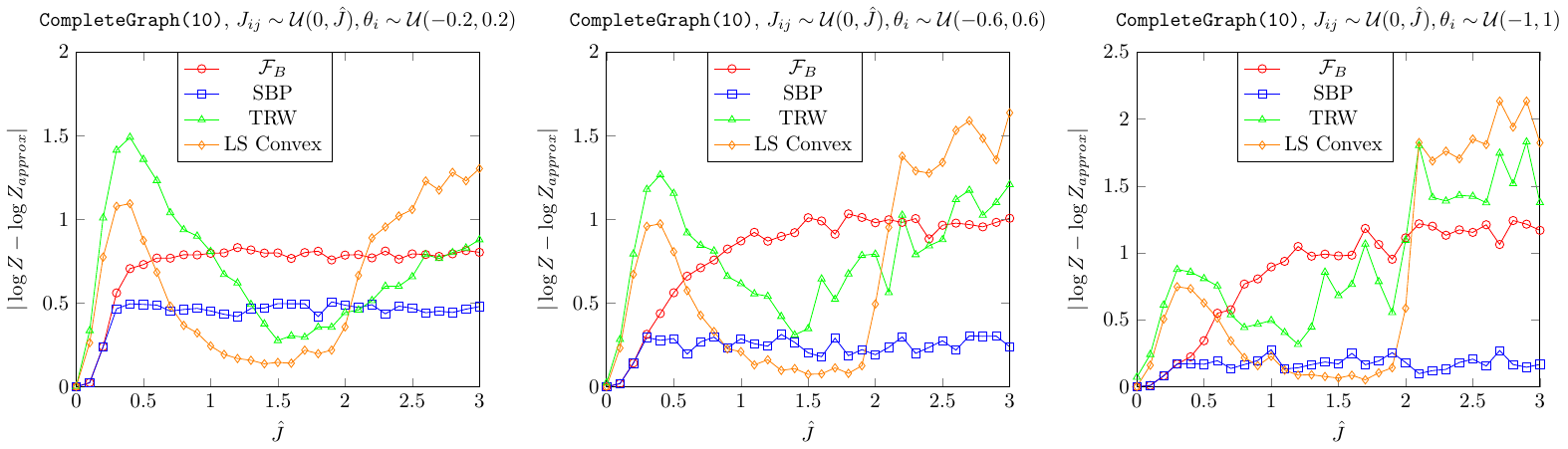}}
\caption{Algorithms Bethe ($\FB$), SBP, TRW, LS-Convex, and \texttt{ADAPT}-$c$'compared on attractive models. First row: $l^1$- error on singleton marginals; second row: $l^1$- error on pairwise marginals; third row: absolute error on log-partition function.}
\label{fig:experiments_attractive}
\end{figure*}

The results for attractive models are shown in Fig.~\ref{fig:experiments_attractive}. We observe that all algorithms perform well if the correlations between model variables are weak (i.e., if the parameters $\Jij$ are small). If $\Jij$ increases beyond a certain threshold (that differs for each method), most algorithms experience a significant degradation
of their approximation accuracy. This effect is sometimes referred to as a \emph{phase transition} in the model, and may be explained by the loss of certain properties such as convexity or uniqueness of a minimum~\citep{mooij2007sufficient,zdeborova2016statistical,leisenberger2024reliability}. In~\citet{wainwright2008exponential} it is argued that convex approximations are more robust to such spontaneous changes in the model dyna\-mics. For the marginals this agrees with our observations, but for the partition function this is less obvious. \\

For estimating singleton marginals, \texttt{ADAPT}-$\zeta$ proves to be the most stable alternative that only suffers a slight loss of accuracy during the increase of $\Jij$. This is especially distinct for weak local potentials $\thi$, while for stronger local potentials also SBP improves its performance considerably.\footnote{Note that SBP mostly outperforms the Bethe approximation, although it minimizes the same free energy approximation in attractive models (namely $\Fzeta$ with $\zeta = 1$, which is precisely $\FB$); however, while the shown results for $\FB$ are based on a 'naive' random initialization of the pseudomarginals, SBP applies an adaptive initialization stratety and thus converges to more accurate minima (Sec.~\ref{sec:analysis_zeta_approx}).} For estimating pairwise marginals, \texttt{ADAPT}-$\zeta$ is less robust and slightly outperformed by SBP. For strong local and pairwise potentials, both methods are superior to the convex approximations. For estimating the partition function, SBP clearly shows the most accurate and stable performance while \texttt{ADAPT}-$\zeta$ is not competitive (thus the corresponding results are omitted at this point).\footnote{As mentioned in Sec.~\ref{sec:analysis_zeta_approx}, $\Fzeta$-approximations should be used with care if we aim to estimate the partition function, because varying $\zeta$ changes the model parameters and thus the magnitude of the Bethe free energy. However, if the deviations from the original model are moderate, one may use the estimated marginals from the $\Fzeta$-approximation and insert them into the Bethe free energy to circumvent problematic aspects of changing the model (i.e., we evaluate $\FB$ in a point that is a minimum with respect to the modified approximation $\Fzeta$, and not of $\FB$, to estimate $- \log Z$). We have used this approach for estimating the partition function with SBP in mixed models in which it mostly uses a $\zeta$ that is different from one (Sec.~\ref{subsec:adaptive_experiments_general}).}

\subsection{Mixed Models} \label{subsec:adaptive_experiments_general}
\begin{figure*}[h!]
\centering
\subfigure{\includegraphics[width=0.53\linewidth]{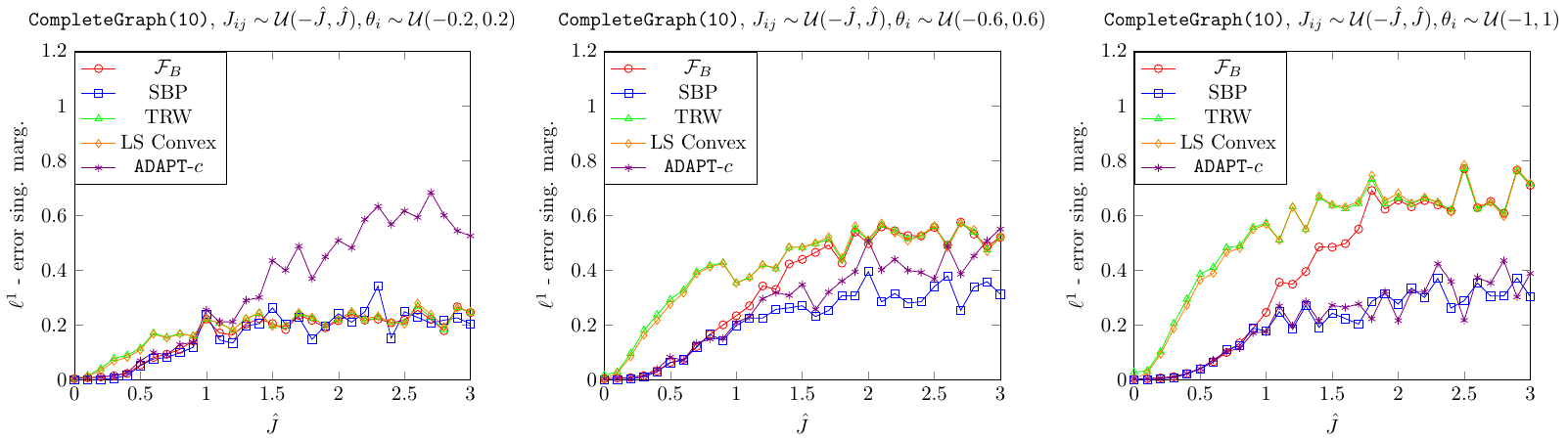}} \\
\subfigure{\includegraphics[width=0.53\linewidth]{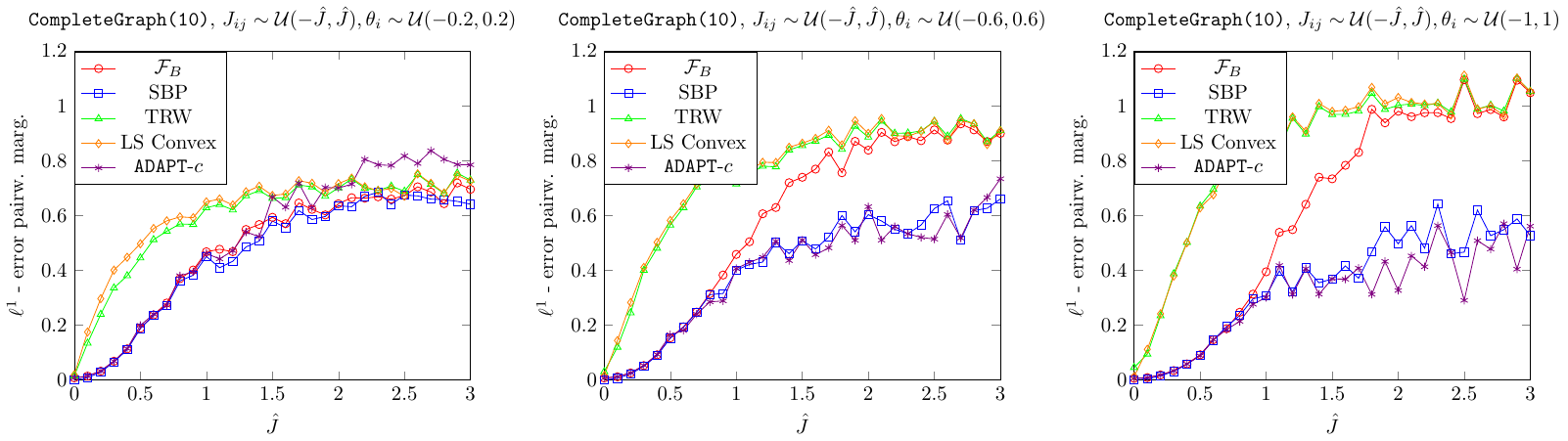}} \\
\subfigure{\includegraphics[width=0.53\linewidth]{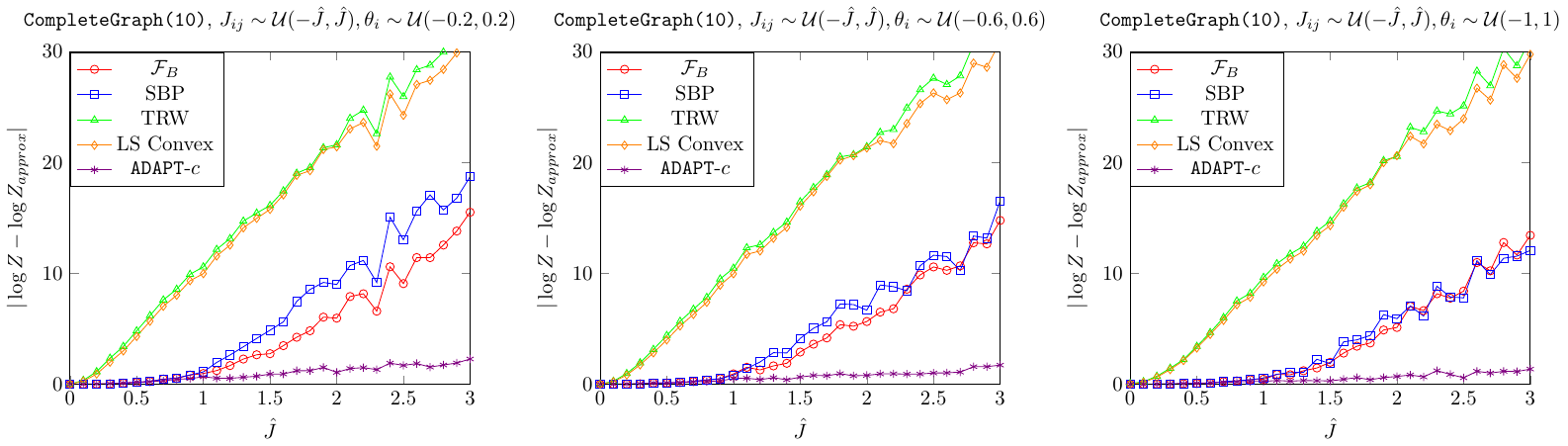}}
\caption{Algorithms Bethe ($\FB$), SBP, TRW, LS-Convex, and \texttt{ADAPT}-$c$ compared on mixed models. First row: $l^1$- error on singleton marginals; second row: $l^1$- error on pairwise marginals; third row: absolute error on log-partition function.}
\label{fig:experiments_general}
\end{figure*}

The results for mixed models are shown in Fig.~\ref{fig:experiments_general}. Similar as for attractive models, the performance of most algorithms deteriorates if $\Jij$ increases. However, this degradation happens rather slowly and steadily than spontanously; in particular, an acceptable approximation accuracy for the marginals is achieved for even higher values of $\Jij$. Two possible explanations for this behavior are the following: either properties such as convexity or uniqueness of a minimum are less relevant in mixed models~\citep{weller2014understanding}; or these properties are relevant but preserved in mo\-dels with stronger correlations~\citep{leisenberger2024reliability}. \\

For estimating singleton marginals SBP is the most stable alternative, while \texttt{ADAPT}-$c$ is inferior for weak local potentials $\thi$. If $\thi$ increases, \texttt{ADAPT}-$c$ achieves a similar performance as SBP. For pairwise potentials the situation is similiar with \texttt{ADAPT}-$c$ even having a slight advantage over SBP if local potentials are strong. For estimating the partition function, \texttt{ADAPT}-$c$ is by far the most powerful alternative. While any other method fails to approximate $Z$ by several orders of magnitude, \texttt{ADAPT}-$c$ proves to be accurate and stable. These observations are in agreement with the discussion of $\Fc$-approximations in Sec.~\ref{sec:analysis_c_approx}. This indicates that \texttt{ADAPT}-$c$ performs well in approximating the solution of problem~\eqref{eq:optimal_c}.

\section{CONCLUSION} \label{sec:conclusion}
In this work we have considered pairwise free energy approximations of two particular classes $\Fc$ and $\Fzeta$, which both generalize the Bethe free ener\-gy.
We have analyzed them in a broader context, by systematically varying their governing parameters (the counting numbers $\cij$ for $\Fc$ and the scale factors $\zij$ for $\Fzeta$) and evaluating the impact on their approximation accuracy.
We have drawn practical conclusions, and proposed two adaptive free energy approximations -- one for attractive and one for mixed models -- that automatically adapt to a model.
Our experiments show that our algorithms slightly improve on estima\-ting singleton marginals in attractive models and pairwise marginals in mixed models, and drastically improve on estimating the partition function in mixed models. \newpage

\appendix


\section{Results from related work} \label{sec:appendix_related_results}

For implementing our algorithms, we require a few results from related work. First, we use from~\citet{welling2001belief} that we can reparameterize the local polytope $\polytopeLocal$ to get a simpler description of pairwise free energy approximations (Sec. 2 in the main paper). \\

Let $X_i, X_j$ be any pair of connected variables and let $\tilde{p}_i, \tilde{p}_j, \tilde{p}_{ij}$ be a set of associated pseudo-marginal vectors\footnote{More precisely, $\tilde{p}_{i} = \begin{pmatrix}\tilde{p}_{i}(X_{i} = +1)\\\tilde{p}_{i}(X_{i} = -1)\end{pmatrix}$ and $\tilde{p}_{ij} = \begin{pmatrix}\tilde{p}_{ij}(X_i = +1,X_j = +1)\\\tilde{p}_{ij}(X_i = +1,X_j = -1)\\\tilde{p}_{ij}(X_i = -1,X_j = +1)\\\tilde{p}_{ij}(X_i = -1,X_j = -1)\end{pmatrix}$.} from the local polytope $\polytopeLocal$ (i.e., with all entries satisfying the constraints in the definition of $\polytopeLocal$). Let us denote the individual pseudo-marginal probabilities for $X_i$ and $X_j$ being in state $+1$ by $q_i \coloneqq \tilde{p}_i(X_i = +1)$ and $q_j \coloneqq \tilde{p}_j(X_j = +1)$ and let us further denote the joint pseudo-marginal probability for $X_i,X_j$ being both in state $+1$ by $\xij \coloneqq \tilde{p}_{ij}(X_i = +1,X_j = +1)$. Then, as $\tilde{p}_i, \tilde{p}_j, \tilde{p}_{ij}$ must satisfy the constraints of $\polytopeLocal$, all five remaining entries of these vectors can already be expressed in terms of $q_i, q_j, \xij$ according to the following table:

\begin{table}[h]
\centering
\caption{Joint probability table of two binary variables $X_i$ and $X_j$.} \label{tab:prob_table}
\begin{tabular}{ c||c|c||c }
 $\tilde{p}_{ij}(X_i, X_j)$ & $X_j = +1$ & $X_j = -1$ & \\ \hline \hline
 $X_i = +1$ & $\xi_{ij}$ & $q_i - \xi_{ij}$ & $q_i$ \\ \hline
 $X_i = -1$ & $q_j - \xi_{ij}$ & $1 + \xi_{ij} - q_i - q_j$ & $1-q_i$ \\ \hline \hline
 & $q_j$ & $1 - q_j$ &  \\
\end{tabular}
\end{table}

Additionally, we have constraints on the independent parameters $q_i, q_j, \xij$ of the form

\begin{align*}
 0 < & \, \, q_i < 1, \\
 0 < & \, \, q_j < 1, \\
 \max(0,q_i + q_j - 1) < & \, \, \xi_{ij}< \min(q_i,q_j).
\end{align*}

This step reparameterizes the local polytope by exploiting probabilistic dependencies between variables and removing a considerable number of redundant parameters. If we collect all node-specific variables $q_i$ in a vector $\bm{q}$ and all edge-specific parameters $\xij$ in a vector $\bm{\xi}$, we can compactly rewrite the local polytope $\polytopeLocal$ as the set

\begin{align} \label{eq:local_polytope_reparameterized}
 \begin{split}
\hspace{-0.5cm} \polytopeLocal = \{& (\bm{q}; \bm{\xi}) \in \mathbb{R}^{\lvert \setofnodes \rvert + \lvert \setofedges \rvert}: 0 <  q_i  < 1, i \in \setofnodes; \\ & \max(0, q_i + q_j - 1) <  \xi_{ij}  < \min(q_i,q_j), (i,j) \in \setofedges \}.
 \end{split}
\end{align}

This reparameterization of $\polytopeLocal$ also reparameterizes the Bethe free energy: By substituting the expressions from Table~\ref{tab:prob_table} into the statistics of $\FB$, it takes the simpler form
\begin{align} \label{eq:Bethe_reparameterized}
 \begin{split}
 \FB(\bm{q}; \bm{\xi}) = & - \sum_{(i,j) \in \setofedges} \, (1+  2 \; (2 \, \xi_{ij} - q_i - q_j)) \, J_{ij} + \, \, \sum_{i \in \setofnodes} (1 - 2 q_i) \, \theta_i \\  & - \Big(\sum_{(i,j) \in \setofedges} \mathcal{S}_{ij} \; - \; \sum_{i \in \setofnodes} (d_i - 1) \, \mathcal{S}_{i} \Big) ,
 \end{split}
\end{align}
with the pairwise entropies
\begin{align} \label{eq:Bethe_entropy_PW}
 \begin{split}
 \hspace{-0.15cm} \mathcal{S}_{ij} = & -\xi_{ij} \log \xi_{ij} - (1+\xi_{ij}-q_i-q_j) \log (1+\xi_{ij}-q_i-q_j) \\ & - (q_i - \xi_{ij}) \log (q_i - \xi_{ij}) - (q_j - \xi_{ij}) \log (q_j - \xi_{ij})
   \end{split}
\end{align}
and the local entropies
\begin{align} \label{eq:Bethe_entropy_Loc}
 \begin{split}
 \mathcal{S}_{i} = -q_i \log q_i - (1-q_i) \log (1-q_i).
   \end{split}
\end{align}

By introducing counting numbers and/or scale factors as in Sec. 2.3 of the main paper, it is straightforward to adapt this reparameterization steps to free energy approximations of class $\Fc$ or $\Fzeta$. Next, we show that we can directly parameterize pairwise free energy approximations via the singleton pseudo-marginals~\citep{welling2001belief,weller2015pairwise}. Without loss of generality, we assume that we modify the counting numbers $\cij, c_i$ and $\zij, \zi$ at the same time, i.e., obtain a combination $\mathcal{F}_{\bm{c},\bm{\zeta}}$ of class $\Fc$ and $\Fzeta$. Then, by differentiating $\mathcal{F}_{\bm{c},\bm{\zeta}}$ with respect to $\xi_{ij}$ and setting the derivative to zero, one obtains a quadratic equation in $\qi,\qj,\xi_{ij}$ whose unique solution is given by
 \begin{align} \label{eq:xi_optimal}
 \begin{split}
   & \xijopt(q_i,q_j) =  \frac{1}{2\alpha_{ij}} \Big( \Qij - \sqrt{\Qij^2 - 4 \alpha_{ij}(1+\alpha_{ij}) q_i q_j  } \, \, \Big),  \\
   & \text{where} \, \,  \quad \aij = e^{4 \frac{\zij \Jij}{\cij}} - 1 \quad \, \, \text{and} \quad \, \, \Qij = 1 + \aij(\qi + \qj).
 \end{split}
\end{align}
In other words, for any $\bm{q}$ there exists at most one $\bm{\xi}^{\ast}(\bm{q})$ (defined in~\eqref{eq:xi_optimal}) such that $\big(\bm{q}; \bm{\xi}^{\ast}(\bm{q})\big)$ can be a minimum of $\mathcal{F}_{\bm{c},\bm{\zeta}}$. As a consequence, all minima of $\mathcal{F}_{\bm{c},\bm{\zeta}}$ must lie on a $\lvert \setofnodes \rvert$- dimensional submanifold $\Bbox$ of $\polytopeLocal$ that we define as
 \begin{align} \label{eq:Bethe_box}
  \begin{split}
\Bbox \coloneqq \{ & (\bm{q}; \bm{\xi}^{\ast} (\bm{q})) \in \polytopeLocal: \, \, 0 <  q_i  < 1, i \in \setofnodes; \\
  & \, \, \xijopt(q_i,q_j) \, \, \, \text{given by} \, \, \, \eqref{eq:xi_optimal}, \, (i,j) \in \setofedges \}.
  \end{split}
 \end{align}

Finally, we require the gradient of free energy approximations of class $\Fc$ and $\Fzeta$~\citep{weller2015pairwise}. Again, we assume without loss of generality that we modify the counting numbers $\cij, c_i$ and $\zij, \zi$ at the same time, i.e., obtain a combination $\mathcal{F}_{\bm{c},\bm{\zeta}}$ of class $\Fc$ and $\Fzeta$. Then the first-order partial derivatives of $\mathcal{F}_{\bm{c},\bm{\zeta}}$ on $\Bbox$ are
 \begin{align}
 \begin{split} \label{eq:Bethe_first_derivatives}
  \frac{\partial} {\partial q_{i}} \mathcal{F}_{\bm{c},\bm{\zeta}} =  -2 \, (\zi \theta_i) + 2 \sum\limits_{j \in \nbhi} (\zij J_{ij})  + \log \Big( \frac{q_i^{c_i}}{(1-q_i)^{c_i}} \prod\limits_{j \in \nbhi} \frac{(q_i - \xijopt)^{\cij} }{(1 + \xijopt - q_i - q_j)^{\cij}}  \Big).
 \end{split}
 \end{align}

\section{Details on Algorithms} \label{appendix_pseudocodes}

This section includes the pseudocodes of the algorithms that we have proposed and applied in our work:
\begin{itemize}
\item Algorithm 1 \texttt{F-MIN} was used in Sec. 3 and 4 of the main paper to minimize an arbitrary pairwise free energy approximation of class $\Fc$ or $\Fzeta$ (or a combination of both) to obtain estimates to the exact marginals and partition function.
\item Algorithm 2 is used as a subroutine in \texttt{F-MIN} for an adaptive choice of the step size in the projected Quasi-Newton iterations.
\item Algorithm 3 is \texttt{ADAPT}-$c$ which was proposed in the main paper for approximate inference in mixed models.
\item Algorithm 4 is \texttt{ADAPT}-$\zeta$ which was proposed in the main paper for approximate inference in attractive models.
\end{itemize}

Algorithm~\ref{algo:projected_quasi_Newton_FMIN} describes our proposed method \texttt{F-MIN} for minimizing a free energy approximation $\mathcal{F}_{\bm{c},\bm{\zeta}}$ of class $\Fc$ or $\Fzeta$ (or a combination of both) on $\Bbox$ that we have applied in our experiments in Sec. 3 and 4 in the main paper. First, one initializes the singleton pseudomarginals $\qvec^{(0)}$. Until convergence, \texttt{F-MIN} iterates through the following steps: Update of the search direction according a quasi-Newton scheme, i.e., with the inverse Hessian being replaced by an approximation; projecting the largest possible step back into $\Bbox$; optimizing the step size via an adaptive Wolfe line search (\texttt{WOLFE-LS}, described in Algorithm~\ref{algo:line_search_Wolfe}); updating the parameter vector according to
\begin{align} \label{eq:quasi_newton_update}
\qvec^{(t+1)} = \qvec^{(t)} - \rho^{(t)} \cdot (\textbf{B}^{(t)})^T \, \nabla \mathcal{F}_{\bm{c},\bm{\zeta}}(\qvec^{(t)}),
\end{align}
and updating the current approximation to the inverse Hessian (we use the BFGS update rules for approximating the inverse Hessian,~\citet{nocedal2006numerical}). For checking convergence, we compute the gradient norm with respect to the current parameter vector $\qvec^{(t)}$. After \texttt{F-MIN} has converged, it returns a stationary point $\qvec^{\ast}$ (usually a minimum) of $\mathcal{F}_{\bm{c},\bm{\zeta}}$.

\begin{algorithm}
\begin{algorithmic}
\caption{\texttt{F-MIN} (Projected Quasi-Newton for minimizing free energy approximations of class $\Fc$ or $\Fzeta$)} \label{algo:projected_quasi_Newton_FMIN}
\Require A free energy approximation $\mathcal{F}_{\bm{c},\bm{\zeta}}$ of class $\Fc$ or $\Fzeta$ (or their combination)
\Ensure Stationary point $\qvec^{\ast}$ of $\mathcal{F}_{\bm{c},\bm{\zeta}}$
\State Initialize $\qvec^{(0)}$ randomly in $\Bbox$ \Comment{Initialization of singleton pseudo-marginals}
\State Initialize estimate of singleton marginals $\qvec^{(t)} \gets \qvec^{(0)}$
\State Initialize $\textbf{B}^{(t)}$ randomly as a matrix of size $|\setofnodes|\times|\setofnodes|$
\While{$\lVert \nabla \mathcal{F}_{\bm{c},\bm{\zeta}}\big( \qvec^{(t)} \big)\rVert > \epsilon$} \Comment{Check for convergence}
\State $\qvec^{\text{old}} \gets \qvec^{(t)}$ \Comment{Store current parameter vector}
\State $\bm{d}^{(t)} \gets - \big(\textbf{B}^{(t)} \big)^T \, \nabla \mathcal{F}_{\bm{c},\bm{\zeta}}\big(\qvec^{(t)} \big) $ \Comment{Update search direction}
\State $\qvec^{\pi} \gets \qvec^{(t)} + \bm{d}^{(t)}$
\While{$\qvec^{\pi} \notin \Bbox$} \Comment{Project $\qvec^{\pi}$ back into $\Bbox$}
\State $\rho^{\max} \gets 0.9 \cdot \rho^{\max}$
\State $\qvec^{\pi} \gets \qvec^{(t)} + \rho^{\max} \cdot \bm{d}^{(t)}$
\EndWhile
\State $\rho^{(t)} \gets \texttt{WOLFE-LS}(\qvec^{(t)},\qvec^{\pi})$ \Comment{Compute step size via Algorithm~\ref{algo:line_search_Wolfe}}
\State $\qvec^{(t)} \gets \qvec^{(t)} + \rho^{(t)} \cdot (\qvec^{\pi} - \qvec^{(t)})$ \Comment{Update parameter vector}
\State $\bm{s}^{(t)} \gets \qvec^{(t)} - \qvec^{\text{old}}$
\State $\bm{y}^{(t)} \gets \nabla \mathcal{F}_{\bm{c},\bm{\zeta}}\big(\qvec^{(t)} \big) - \nabla \mathcal{F}_{\bm{c},\bm{\zeta}}\big(\qvec^{\text{old}} \big)$ \Comment{BFGS update rules}
\State $\gamma^{(t)} \gets ( \bm{s}^{(t)})^T \bm{y}^{(t)}$ \Comment{\hspace{0.08cm} for approximating}
\State $\textbf{B}^{(t)} \gets \textbf{B}^{(t)} + \frac{1}{\big(\gamma^{(t)}\big)^2} \big(\gamma^{(t)} + (\bm{y}^{(t)})^T \textbf{B}^{(t)} \bm{y}^{(t)} \big)$ \Comment{\hspace{-0.12cm} the inverse Hessian} \\
\hspace{1.8cm} $+ \frac{1}{\gamma^{(t)}} \big(\textbf{B}^{(t)} \bm{y}^{(t)} ( \bm{s}^{(t)})^T + \bm{s}^{(t)} (\bm{y}^{(t)})^T \textbf{B}^{(t)} \big)$
\EndWhile
\State \Return $\qvec^{(t)}$ \Comment{Return a stationary point of the provided free energy approximation}
\end{algorithmic}
\end{algorithm}

The Wolfe conditions used in the adaptive line search stategy are as follows:

\begin{align}
  & \hspace{-0.3cm} \mathcal{F}_{\bm{c},\bm{\zeta}}\big(\qvec^{\text{tail}} + \rho^{\text{W}} \cdot (\qvec^{\text{head}} - \qvec^{\text{tail}})\big) \, \leq \, \mathcal{F}_{\bm{c},\bm{\zeta}}\big(\qvec^{\text{tail}}\big) + \tau_1 \cdot \rho^{\text{W}} \cdot \bm{d}^T \, \nabla \mathcal{F}_{\bm{c},\bm{\zeta}}\big(\qvec^{\text{tail}}\big) \tag{W1} \label{eq:Wolfe-Powell_1} \\
  & \hspace{-0.3cm} \bm{d}^T \, \nabla \mathcal{F}_{\bm{c},\bm{\zeta}}\big(\qvec^{\text{tail}} + \rho^{\text{W}} \cdot (\qvec^{\text{head}} - \qvec^{\text{tail}} )\big) \, \geq \, \tau_2 \cdot \bm{d}^T \, \nabla \mathcal{F}_{\bm{c},\bm{\zeta}}\big(\qvec^{\text{tail}} \big) \tag{W2} \label{eq:Wolfe-Powell_2}
\end{align}

The parameter vectors $\qvec^{\text{tail}}$  and $\qvec^{\text{head}}$ specify the search interval. Condition~\eqref{eq:Wolfe-Powell_1} ensures that there is a sufficient decrease of the energy function after each iteration. This can be achieved by sufficiently reducing the step size $\rho^{\text{W}}$ to prevent $\mathcal{F}_{approx}$ from increasing again (which may happen if the step is too large). Conversely, condition~\eqref{eq:Wolfe-Powell_2} ensures that we do not stop moving along the search direction $\bm{d}$, as long as the energy function descends sufficiently steeply. Hence, the step size must not be chosen too small. Both conditions together guarantee that the step size is neither chosen too great nor too small. The parameters $\tau_1$ and $\tau_2$ control how 'strict' the individual conditions are. Following a recommendation of~\citet{nocedal2006numerical}, we have selected numerical values of $\tau_1=10^{-4}$ and $\tau_2=0.9$ for our experiments in Sec. 3 and 4 in the main paper. The detailed procedure of computing a step size satisfying the Wolfe conditions is described in Algorithm~\ref{algo:line_search_Wolfe}. It is used as subroutine in Algorithm~\ref{algo:projected_quasi_Newton_FMIN}.

\begin{algorithm}
\begin{algorithmic}
 \caption{\texttt{WOLFE-LS} (Adaptive line search with Wolfe conditions)} \label{algo:line_search_Wolfe}
\Require Endpoints $\qvec^{\text{tail}}$, $\qvec^{\text{head}}$ of search interval
\Ensure Step size $\rho^\text{W}$ satisfying the Wolfe conditions~\eqref{eq:Wolfe-Powell_1} and~\eqref{eq:Wolfe-Powell_2}
\State $\bm{d} \gets \qvec^{\text{head}} - \qvec^{\text{tail}}$ \Comment{Search direction}
\State Initialize step size $\rho^\text{W}$ randomly between zero and one
\While{\eqref{eq:Wolfe-Powell_1} is satisfied and~\eqref{eq:Wolfe-Powell_2} is not satisfied}
\State $\rho^\text{W} = 1.1 \cdot \rho^\text{W}$ \Comment{Expand interval of potential step sizes}
\EndWhile
\If{\eqref{eq:Wolfe-Powell_1} and~\eqref{eq:Wolfe-Powell_2} are satisfied}
\State \Return $\rho^{\text{W}}$ \Comment{STOP. Feasible step size has been found }
\Else
\State $l \gets 0$ \Comment{Define new search interval for feasible step size}
\State $r \gets \rho^\text{W}$ \Comment{Upper bound of new search interval}
\While{\eqref{eq:Wolfe-Powell_1} or~\eqref{eq:Wolfe-Powell_2} is not satisfied} \Comment{Contraction phase}
\State Pick $\rho^\text{W}$ randomly from interval $(l,r)$
\If{\eqref{eq:Wolfe-Powell_1} is not satisfied}
\State $r \gets \rho^\text{W}$ \Comment{Upper bound of search interval is reduced}
\Else
\State $l \gets \rho^\text{W}$ \Comment{Lower bound of search interval is increased}
\EndIf
\EndWhile
\EndIf
\State \Return $\rho^\text{W}$
\end{algorithmic}
\end{algorithm}

Next we present pseudocodes on our algorithms \texttt{ADAPT}-$c$ and \texttt{ADAPT}-$\zeta$ that we have explained in Sec. 3 and compared to other algorithms in Sec. 4 of the main paper.

\begin{algorithm}
\begin{algorithmic}
\caption{\texttt{Adapt}-$c$ (Adaptive algorithm for optimizing a class $\Fc$-approximation in mixed models.} \label{algo:adapt_c}
\Require $\FB$ on an undirected graphical model. \Comment{Start with the Bethe approximation}
\State $c \gets 1$ \Comment{Initialize the joint pairwise counting number}
\State $\log{Z_{approx}^{old}} \gets 0$ \Comment{Initialize current estimate of $-\log{Z}$}
\State $\qvec^{(t)} \gets \texttt{F-MIN}(\FB)$ \Comment{Minimize the current free energy approximation $\FB$ with Algorithm~\ref{algo:projected_quasi_Newton_FMIN}}
\State $\log{Z_{approx}} \gets \FB(\qvec^{(t)})$ \Comment{Update current estimate of $-\log{Z}$}
\While{$| \log{Z_{approx}} - \log{Z_{approx}^{old}} | <  \epsilon_Z$ and  $c < c_{max}$} \Comment{$c_{max}$ specifies an upper bound on $c$}
\State $\log{Z_{approx}^{old}} \gets \log{Z_{approx}}$ \Comment{Update minimum of $\FB$ from previous iteration}
\State $c + \Delta c$ \Comment{Increase counting number $c$ by some step size $\Delta {c}$}
\State $\FB \gets \Fc$ \Comment{Update the free energy approximation based on the current counting number $c$}
\State $\qvec^{(t)} \gets \texttt{F-MIN}(\FB)$
\State $-\log{Z_{approx}} \gets \FB(\qvec^{(t)})$ \Comment{Update current estimate of $-\log{Z}$}
\EndWhile
\Ensure \State \Return $\qvec^{(t)}$
\end{algorithmic}
\end{algorithm}

\begin{algorithm}
\begin{algorithmic}
\caption{\texttt{Adapt-$\zeta$} (Adaptive algorithm for optimizing class $\Fzeta$-approximation in attractive models} \label{algo:adapt_zeta}
\Require $\FB$ on an undirected graphical model. \Comment{Start with the Bethe approximation}
\State $\zeta \gets 1$ \Comment{Initialize the joint pairwise scale factor}
\While{$\FB$ has multiple minima and $\zeta > 0$} \Comment{Use Theorem 4 from~\citet{mooij2007sufficient}}
\State \Comment{to check if $\FB$ has a unique minimum}
\State $\zeta - \Delta \zeta$ \Comment{Decrease scale factor $\zeta$ by some step size $\Delta \zeta$}
\State $\FB \gets \Fzeta$ \Comment{Update the free energy approximation based on the current scale factor $\zeta$}
\EndWhile
\State $\qvec^{(t)} \gets \texttt{F-MIN}(\FB)$ \Comment{Minimize the current free energy approximation $\FB$ with Algorithm~\ref{algo:projected_quasi_Newton_FMIN}}
\Ensure \State \Return $\qvec^{(t)}$
\end{algorithmic}
\end{algorithm}

\section{Additional analysis of $\Fc$ - and $\Fzeta$ - approximations} \label{sec:additional_analysis_Fc_Fzeta}

This section includes additional experiments. In Sec.~\ref{sec:additional_class_c}, we perform an analogous analysis of class $\Fc$ approximations as in Sec. 3.1 of the main paper; in Sec.~\ref{sec:additional_class_zeta}, we perform an analogous analysis of class $\Fzeta$ approximations as in Sec. 3.2 of the main paper. We show some more evaluations on the complete graph on $10$ nodes, and additionally on a grid graph on $5 \times 5$ vertices, and Erdos-Renyi random graphs~\citep{erdos1959random} on $25$ nodes and an edge probability of $0.2$ (i.e., the probability that a pair of nodes is connected by an edge). The experimental setup is the same as explained in Sec. 3 of the main paper; in particular, the results are averaged over $100$ individual models for each specific configuration of the potentials.

\subsection{Additional evaluation of $\Fc$ - approximations} \label{sec:additional_class_c}
Figure 1-2 include analyses on the complete graph (attractive and mixed models). Figure 3-4 include analyses on the grid graph (attractive and mixed models). Figure 5-6 include analyses on the Erdos-Renyi graphs (attractive and mixed models). The results are similar as in the main paper.

\begin{figure*}[!t]
\centering
\subfigure{\includegraphics[width=0.9\linewidth]{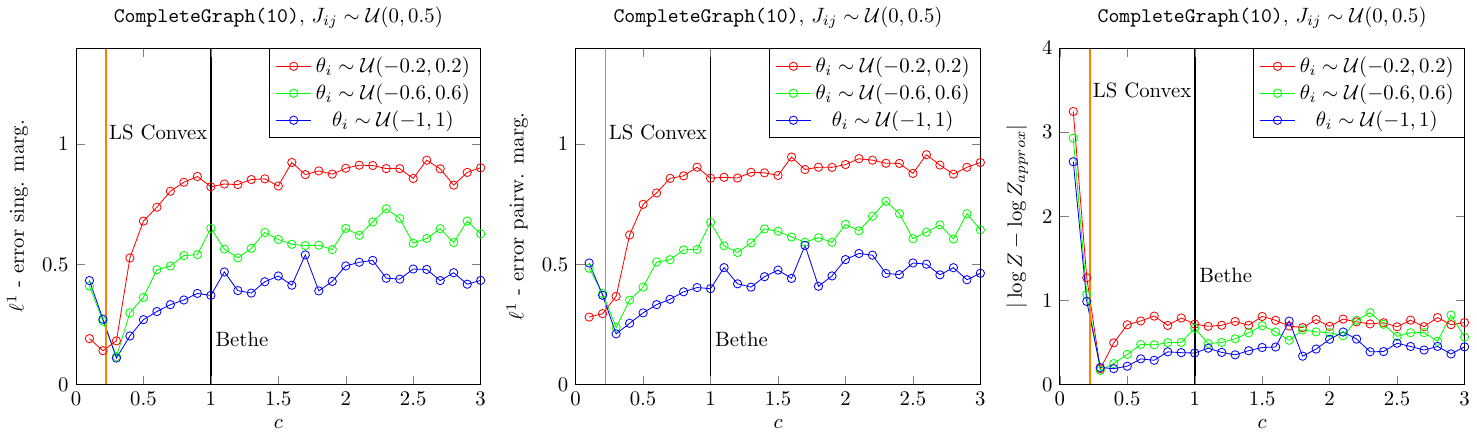}} \\
\subfigure{\includegraphics[width=0.9\linewidth]{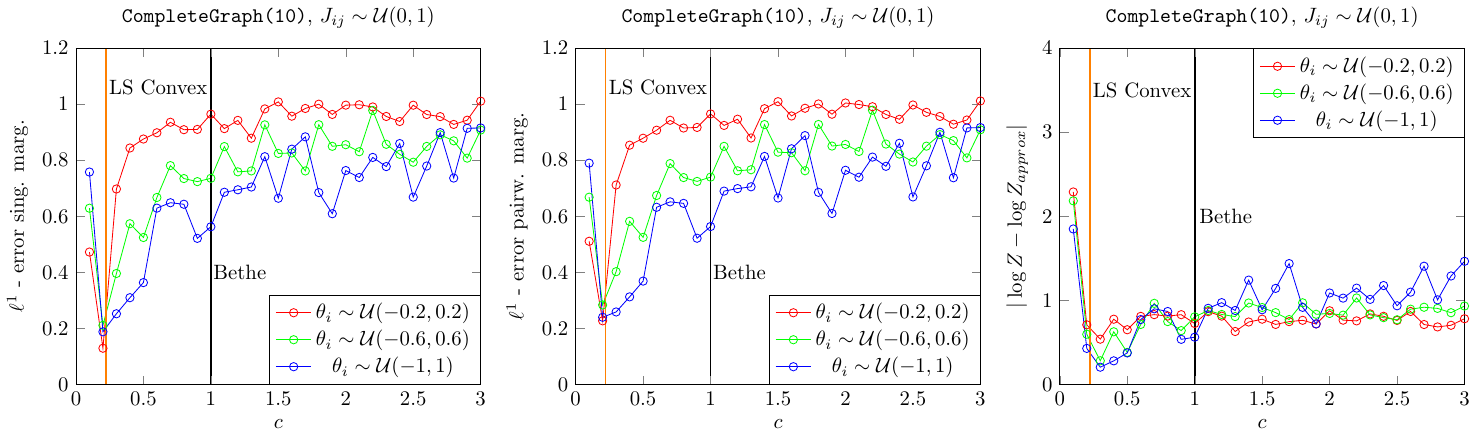}} \\
\subfigure{\includegraphics[width=0.9\linewidth]{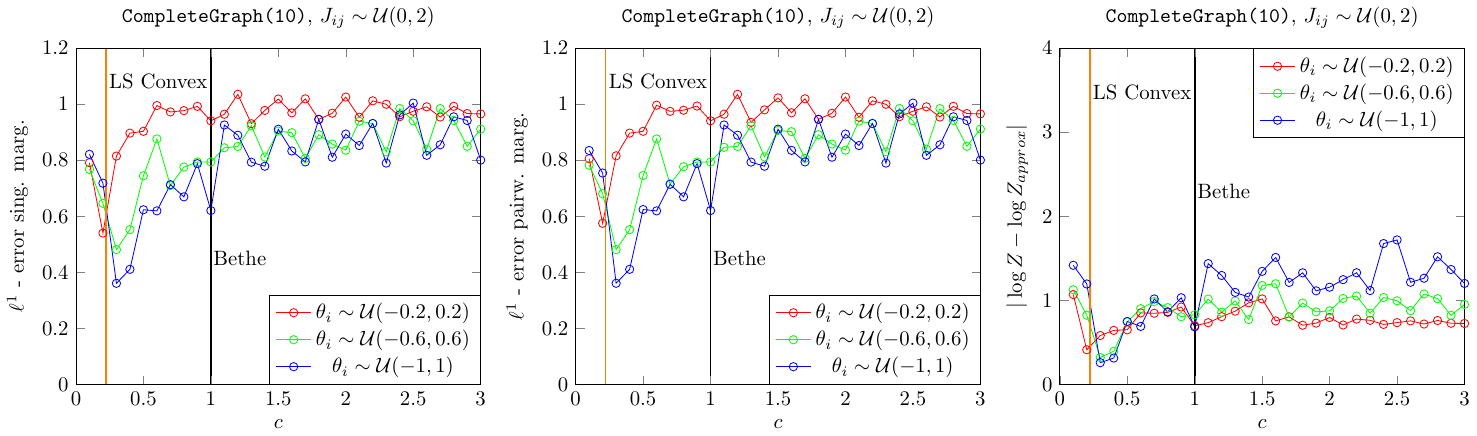}} \\
\subfigure{\includegraphics[width=0.9\linewidth]{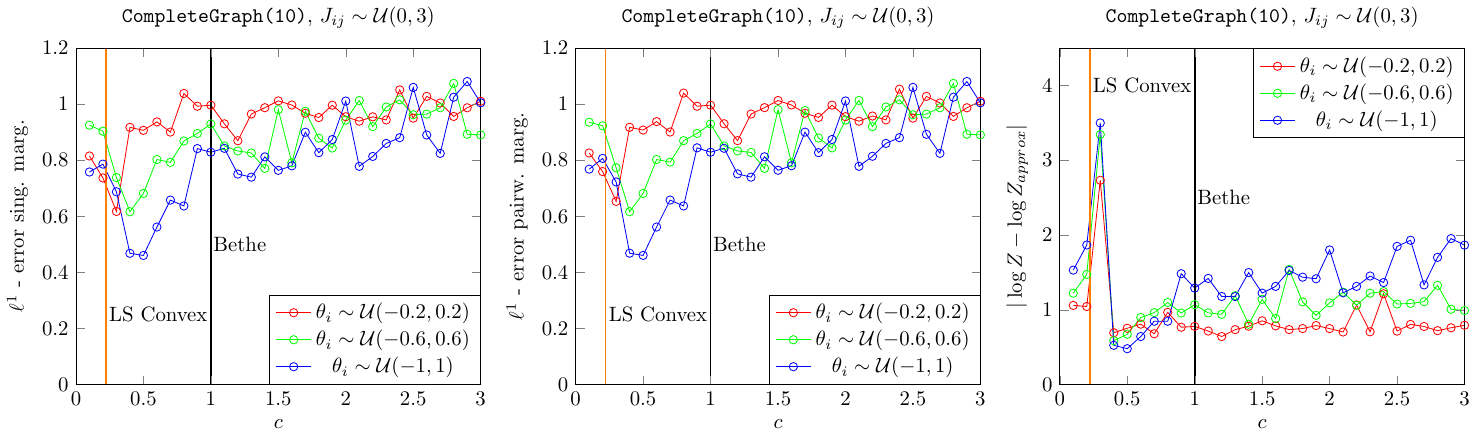}}
\caption{Approximation behavior of $\Fc$: Attractive models, complete graph on $10$ nodes.}
\label{fig:scaling_c_co10_attr}
\end{figure*}

\begin{figure*}[!t]
\centering
\subfigure{\includegraphics[width=0.9\linewidth]{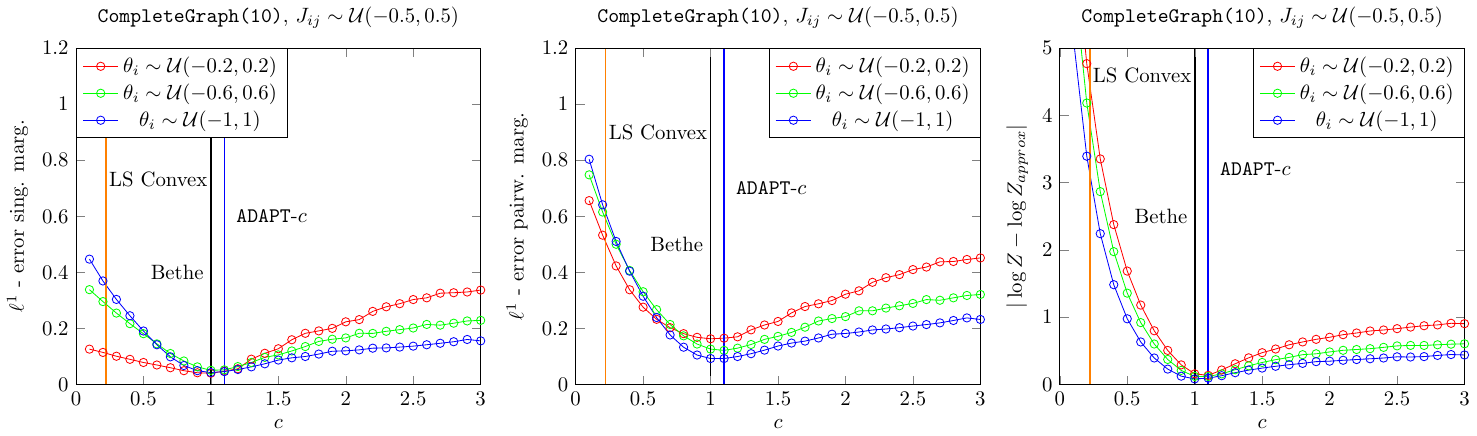}} \\
\subfigure{\includegraphics[width=0.9\linewidth]{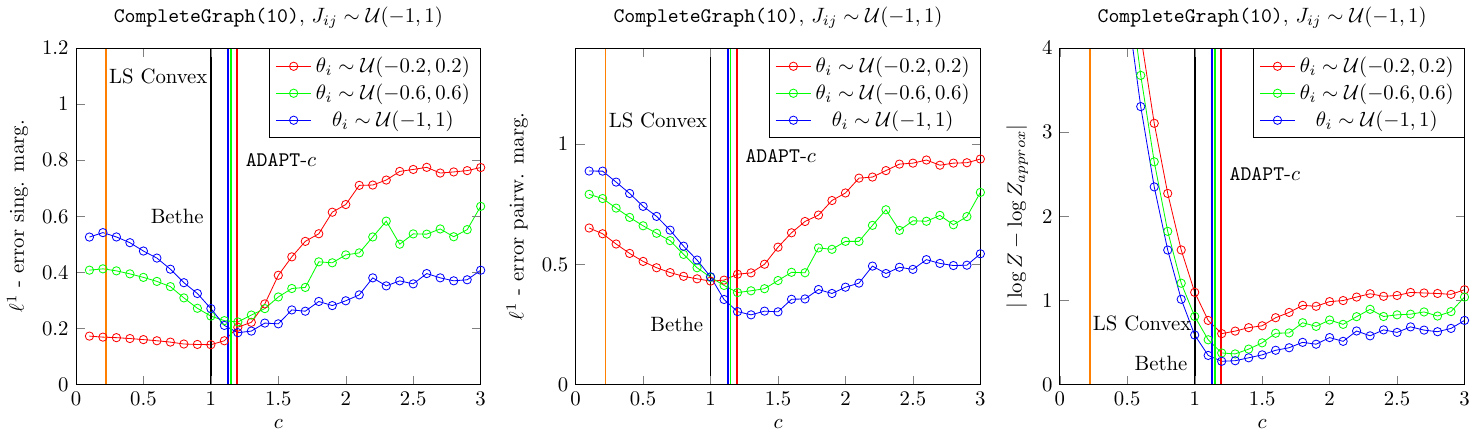}} \\
\subfigure{\includegraphics[width=0.9\linewidth]{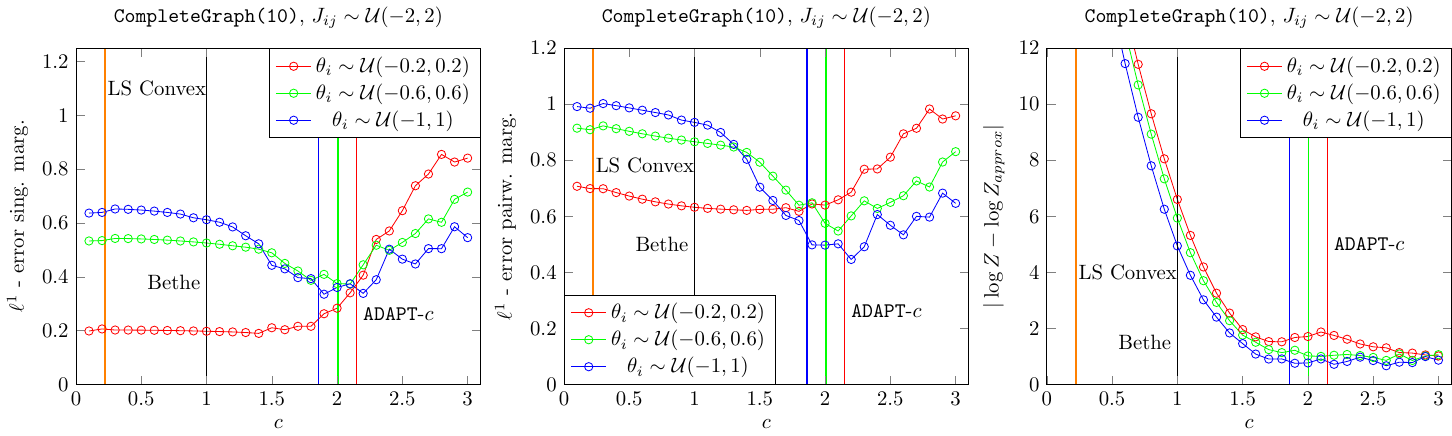}} \\
\subfigure{\includegraphics[width=0.9\linewidth]{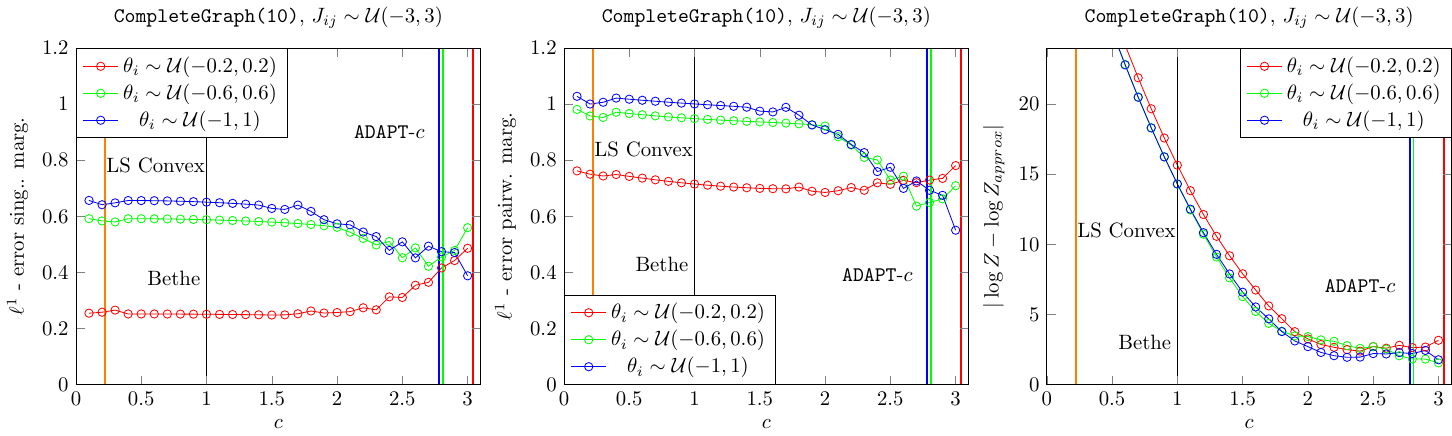}}
\caption{Approximation behavior of $\Fc$: mixed models, complete graph on $10$ nodes.}
\label{fig:scaling_c_co10_gen}
\end{figure*}

\begin{figure*}[!t]
\centering
\subfigure{\includegraphics[width=0.9\linewidth]{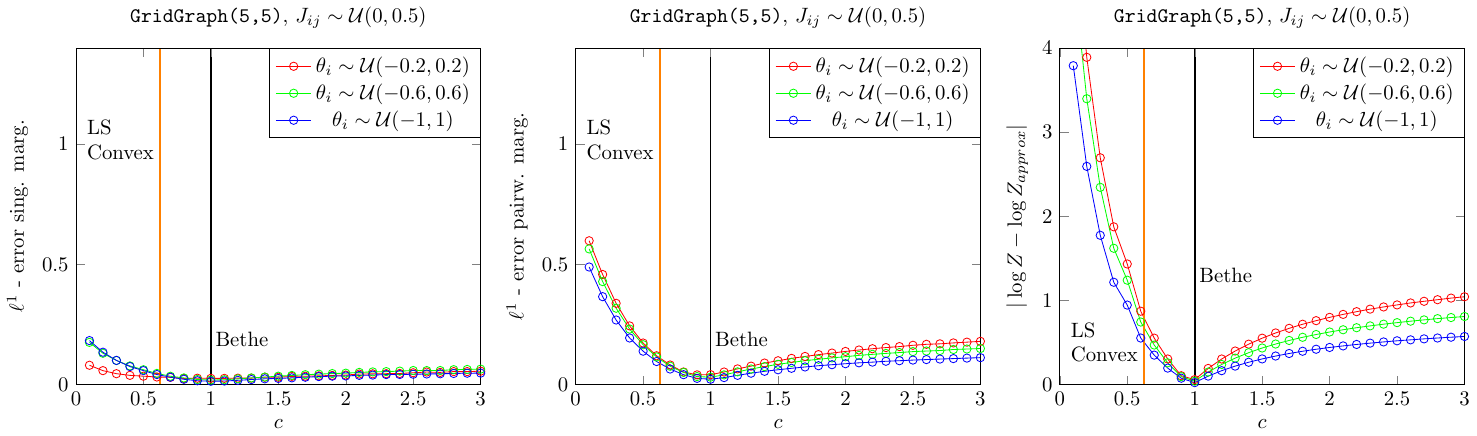}} \\
\subfigure{\includegraphics[width=0.9\linewidth]{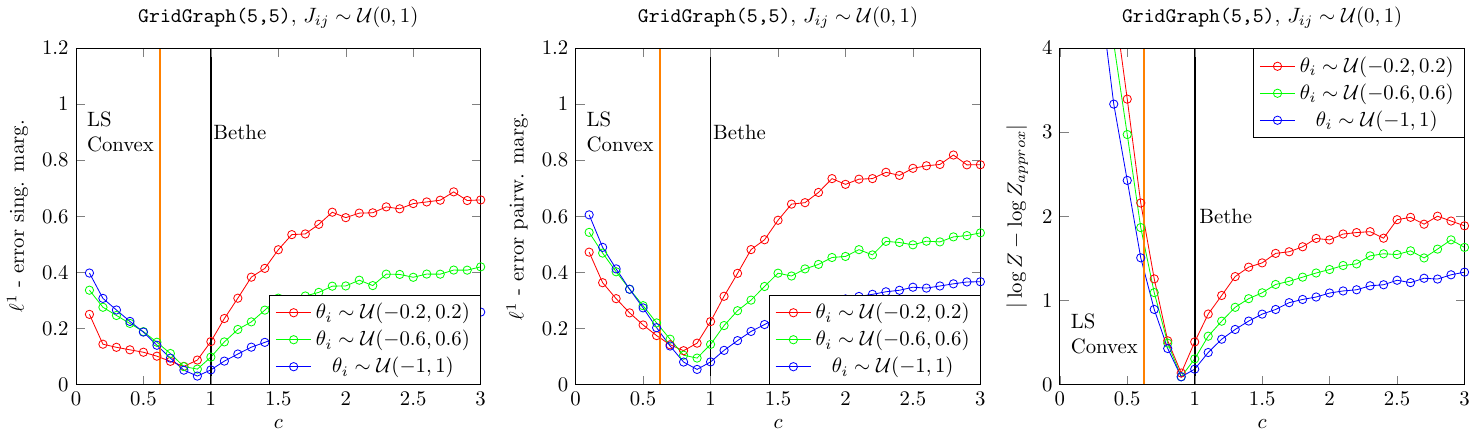}} \\
\subfigure{\includegraphics[width=0.9\linewidth]{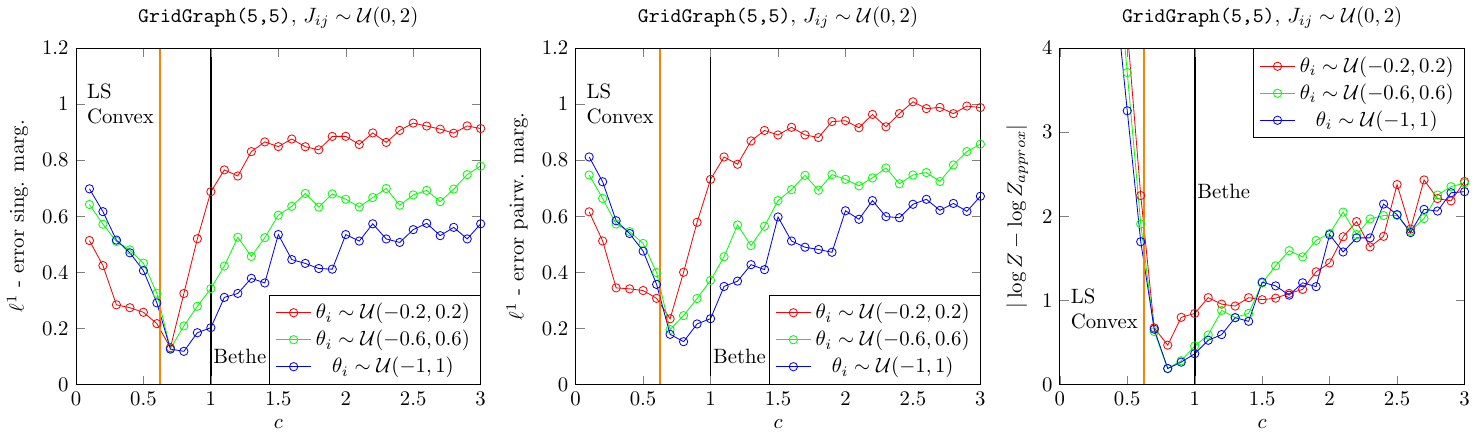}} \\
\subfigure{\includegraphics[width=0.9\linewidth]{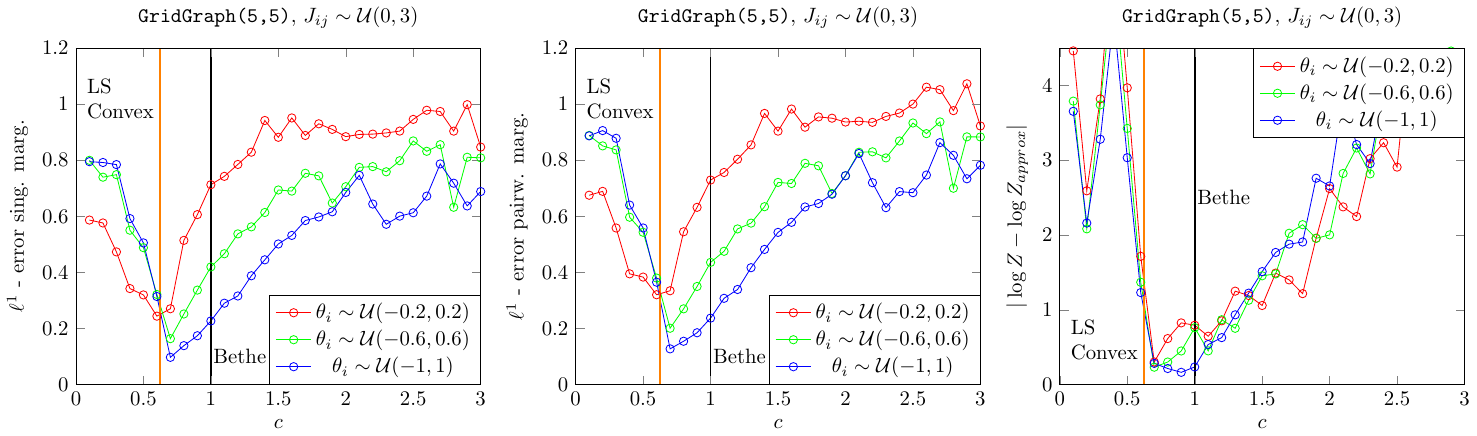}}
\caption{Approximation behavior of $\Fc$: Attractive models, grid graph $5 \times 5$ nodes.}
\label{fig:scaling_c_g5x5_attr}
\end{figure*}

\begin{figure*}[!t]
\centering
\subfigure{\includegraphics[width=0.9\linewidth]{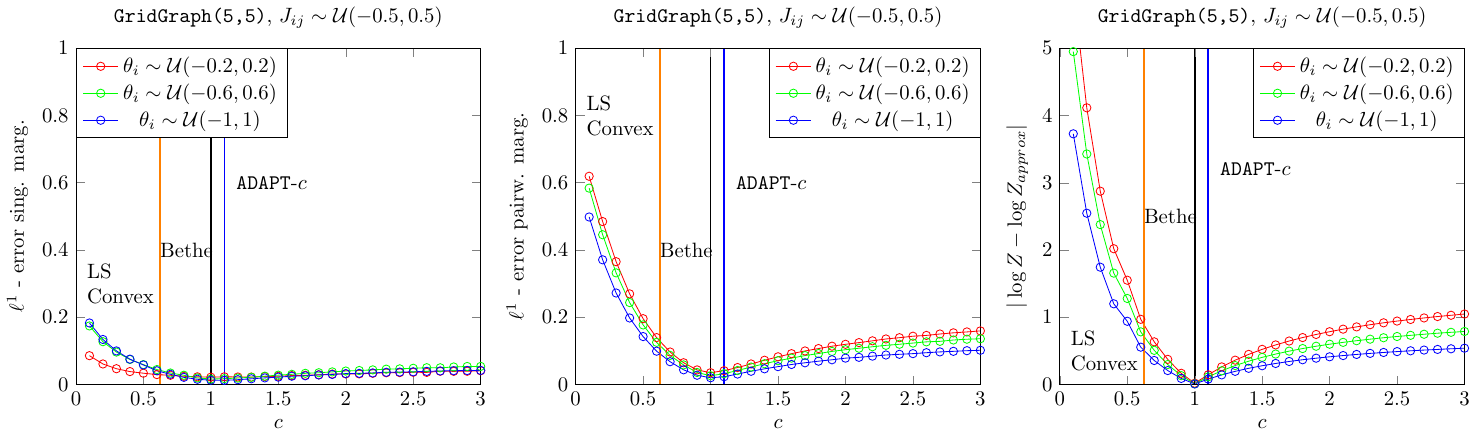}} \\
\subfigure{\includegraphics[width=0.9\linewidth]{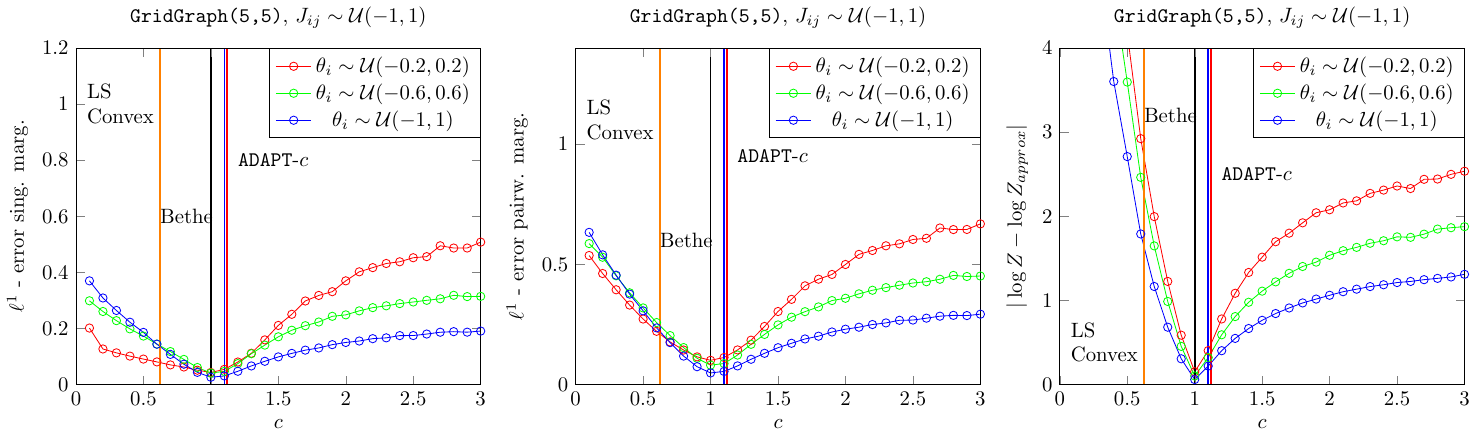}} \\
\subfigure{\includegraphics[width=0.9\linewidth]{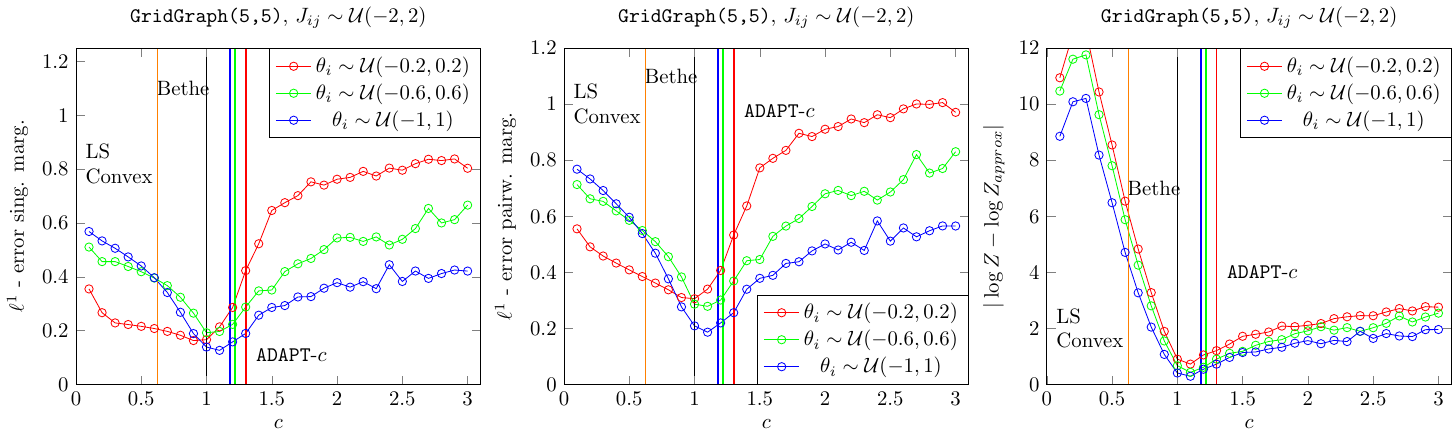}} \\
\subfigure{\includegraphics[width=0.9\linewidth]{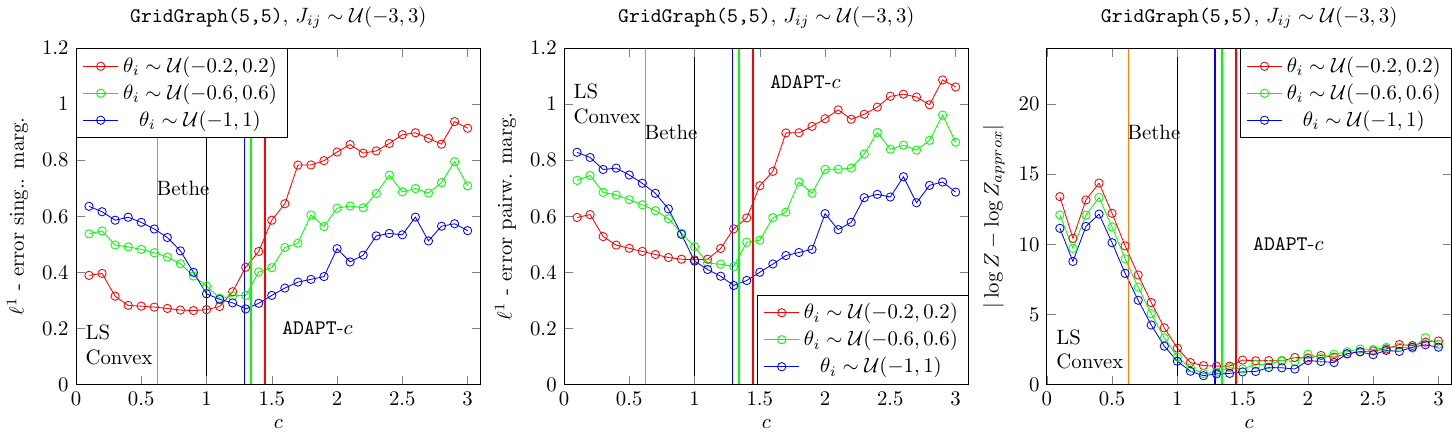}}
\caption{Approximation behavior of $\Fc$: mixed models, grid graph $5 \times 5$ nodes.}
\label{fig:scaling_c_g5x5_gen}
\end{figure*}

\begin{figure*}[!t]
\centering
\subfigure{\includegraphics[width=0.9\linewidth]{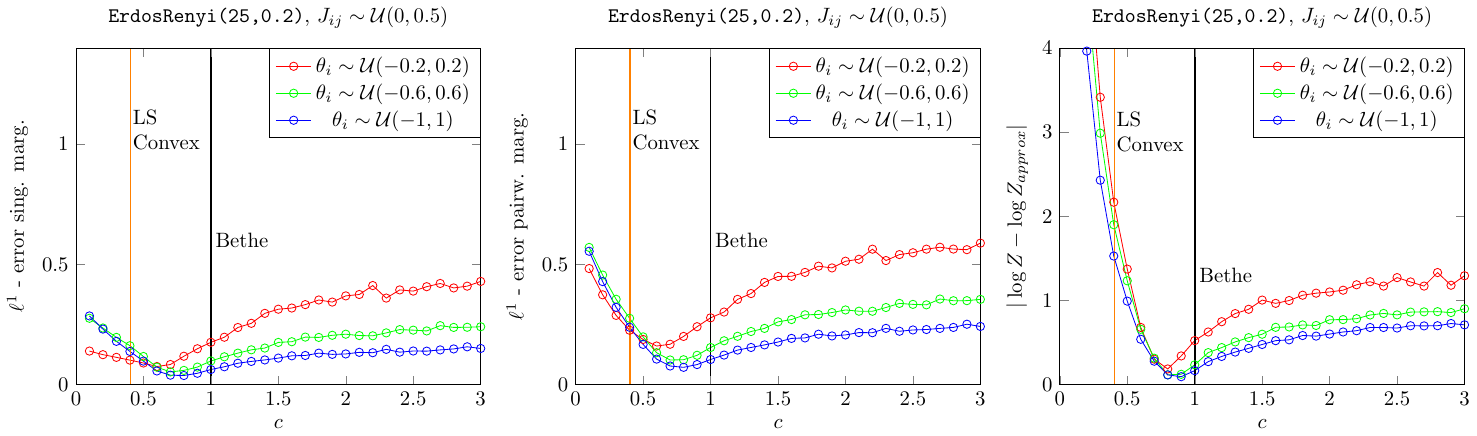}} \\
\subfigure{\includegraphics[width=0.9\linewidth]{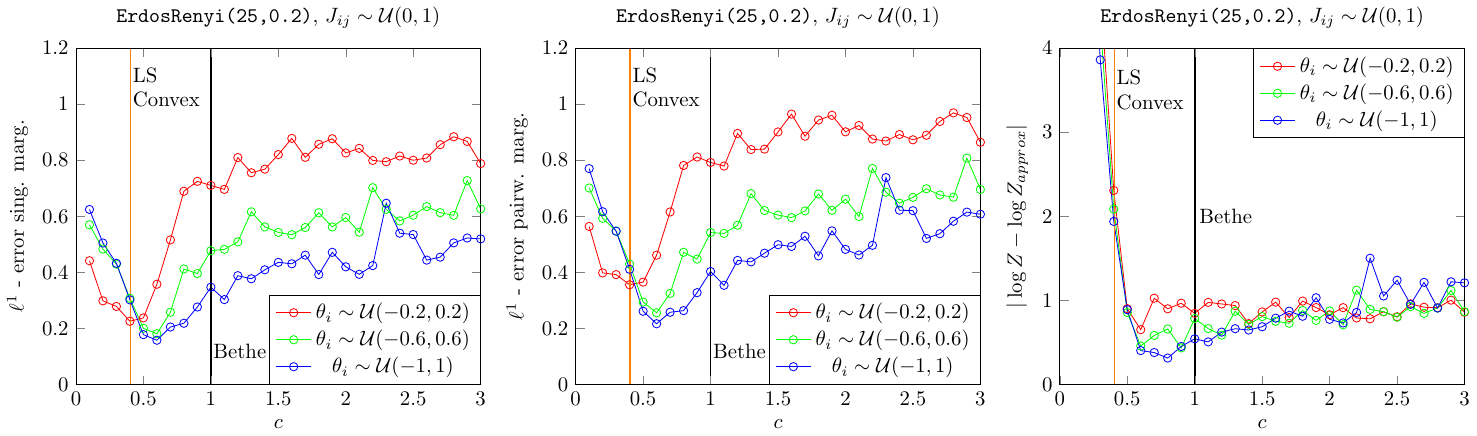}} \\
\subfigure{\includegraphics[width=0.9\linewidth]{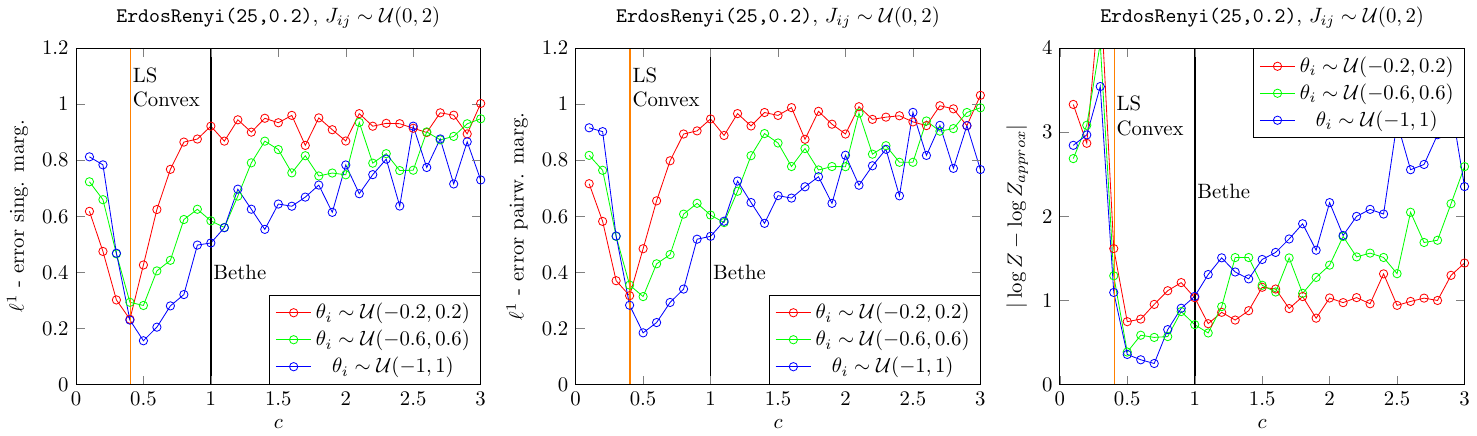}} \\
\subfigure{\includegraphics[width=0.9\linewidth]{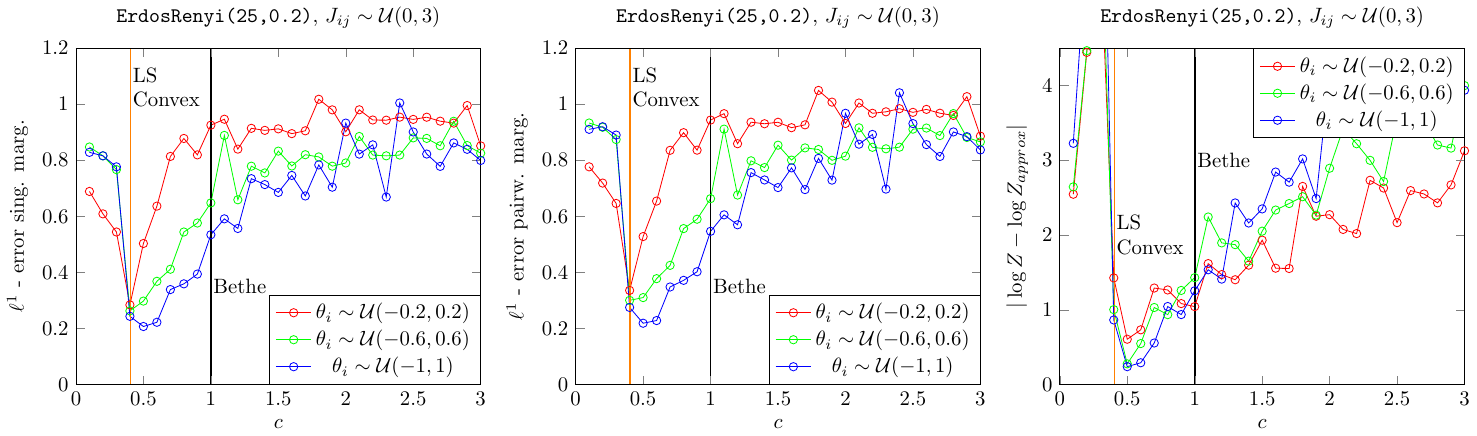}}
\caption{Approximation behavior of $\Fc$: Attractive models, Erdos-Renyi random graphs on $25$ nodes and an edge probability of $0.2$.}
\label{fig:scaling_c_er25x02_attr}
\end{figure*}

\begin{figure*}[!t]
\centering
\subfigure{\includegraphics[width=0.9\linewidth]{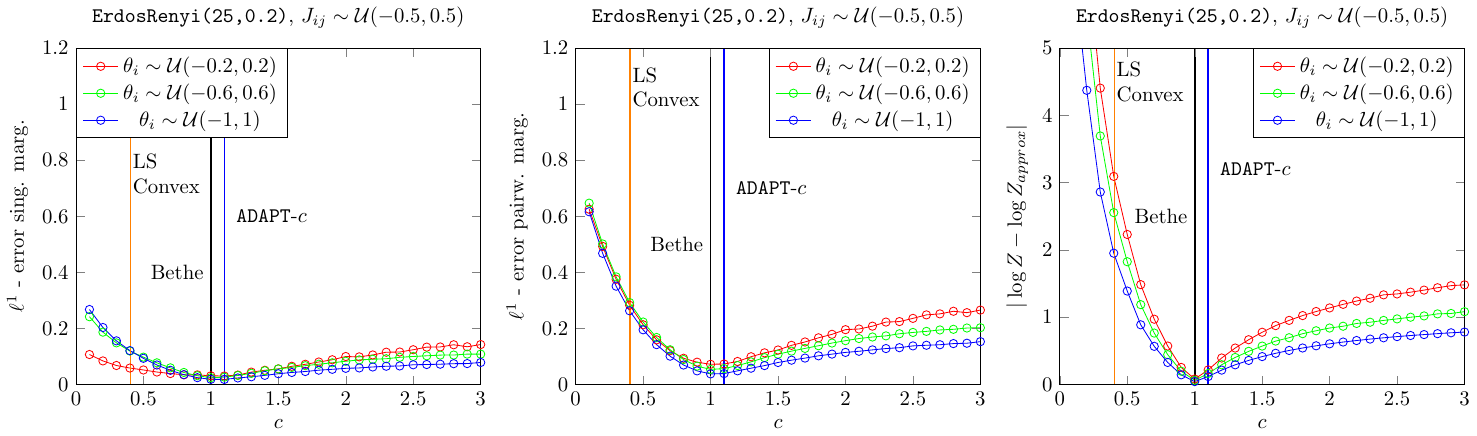}} \\
\subfigure{\includegraphics[width=0.9\linewidth]{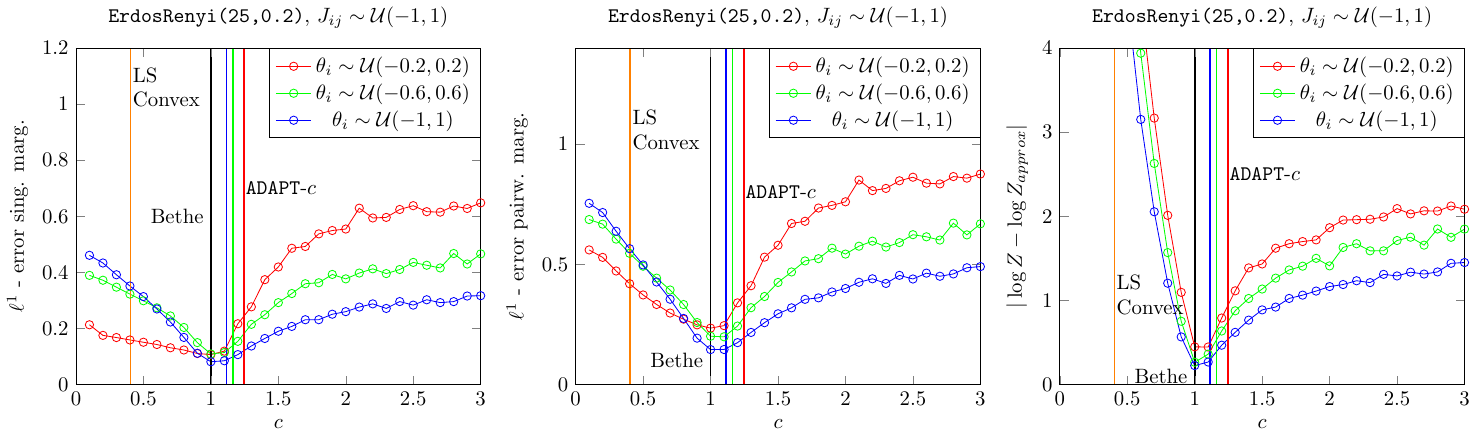}} \\
\subfigure{\includegraphics[width=0.9\linewidth]{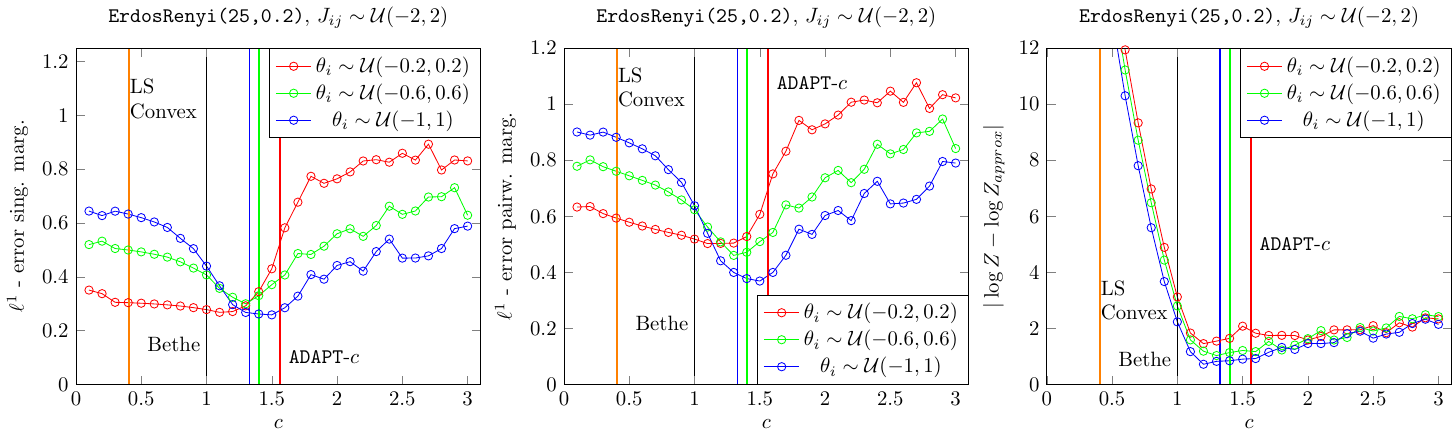}} \\
\subfigure{\includegraphics[width=0.9\linewidth]{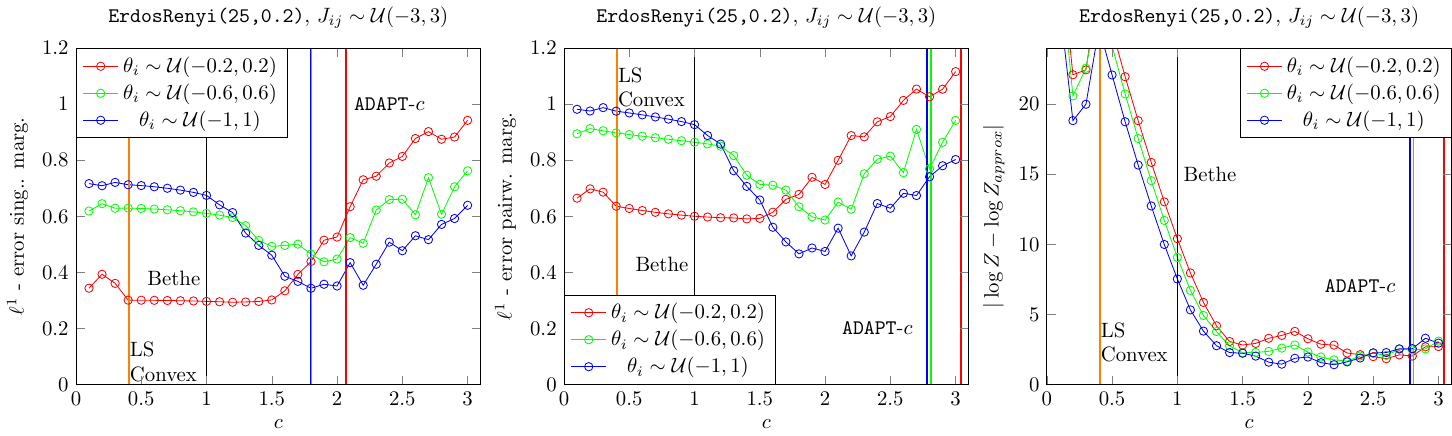}}
\caption{Approximation behavior of $\Fc$: mixed models, Erdos-Renyi random graphs on $25$ nodes and an edge probability of $0.2$.}
\label{fig:scaling_c_er25x02_gen}
\end{figure*}

\subsection{Additional evaluation of class $\Fzeta$ approximations} \label{sec:additional_class_zeta}
Figure 7-8 include analyses on the complete graph (attractive and mixed models). Figure 9-10 include analyses on the grid graph (attractive and mixed models). Figure 11-12 include analyses on the Erdos-Renyi graphs (attractive and mixed models). The results are similar as in the main paper.

\begin{figure*}[!t]
\centering
\subfigure{\includegraphics[width=0.9\linewidth]{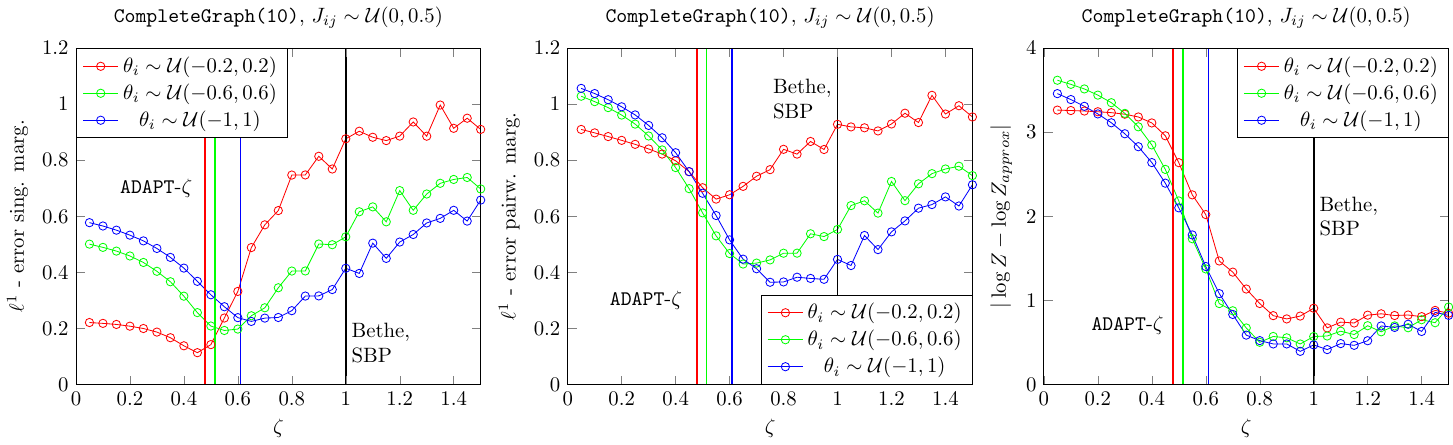}} \\
\subfigure{\includegraphics[width=0.9\linewidth]{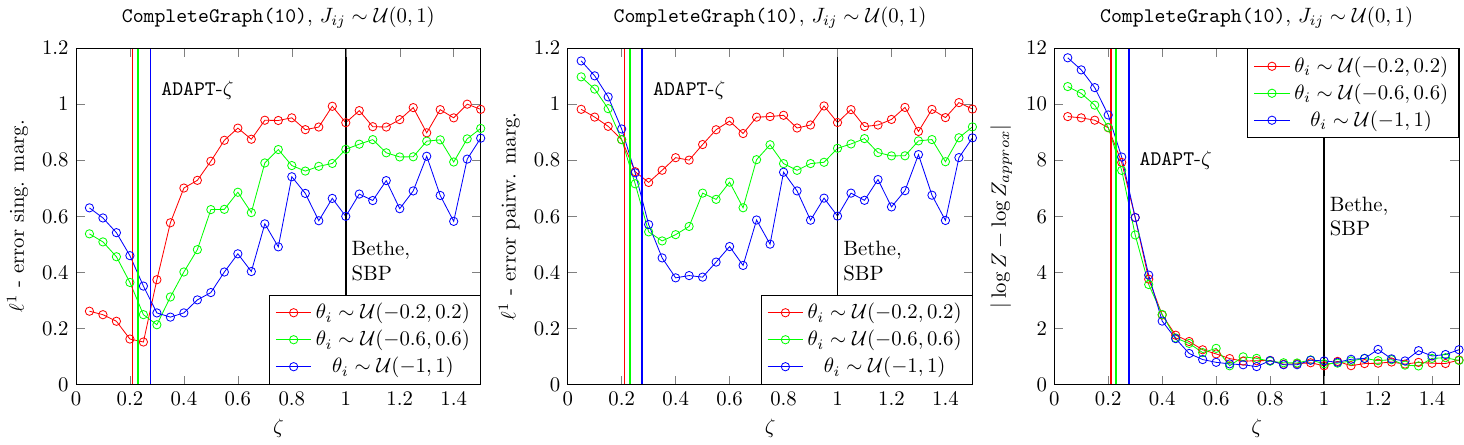}} \\
\subfigure{\includegraphics[width=0.9\linewidth]{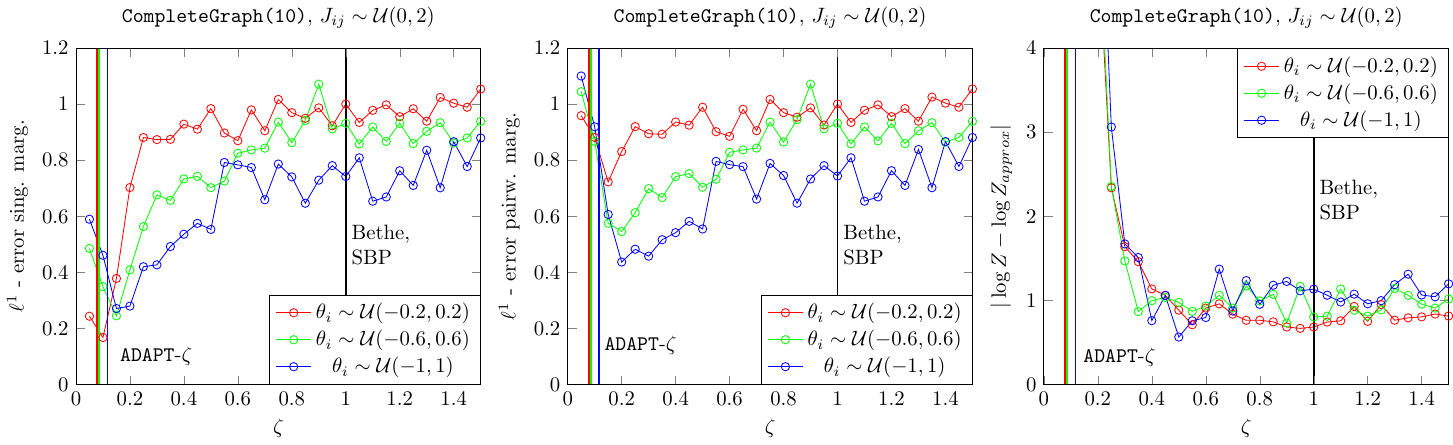}} \\
\subfigure{\includegraphics[width=0.9\linewidth]{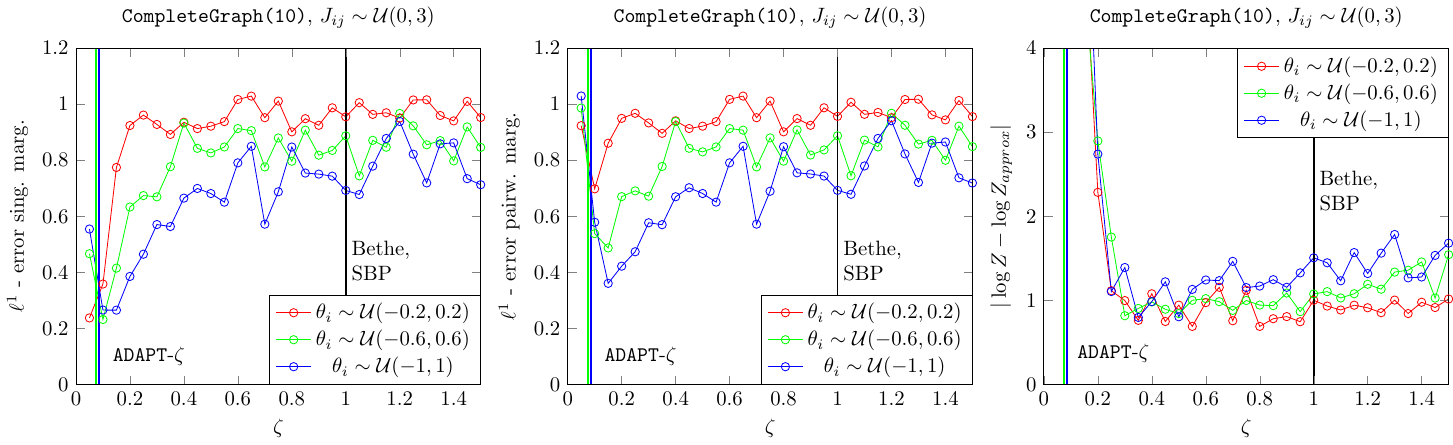}}
\caption{Approximation behavior of $\Fzeta$: Attractive models, complete graph on $10$ nodes.}
\label{fig:scaling_J_co10_attr}
\end{figure*}

\begin{figure*}[!t]
\centering
\subfigure{\includegraphics[width=0.9\linewidth]{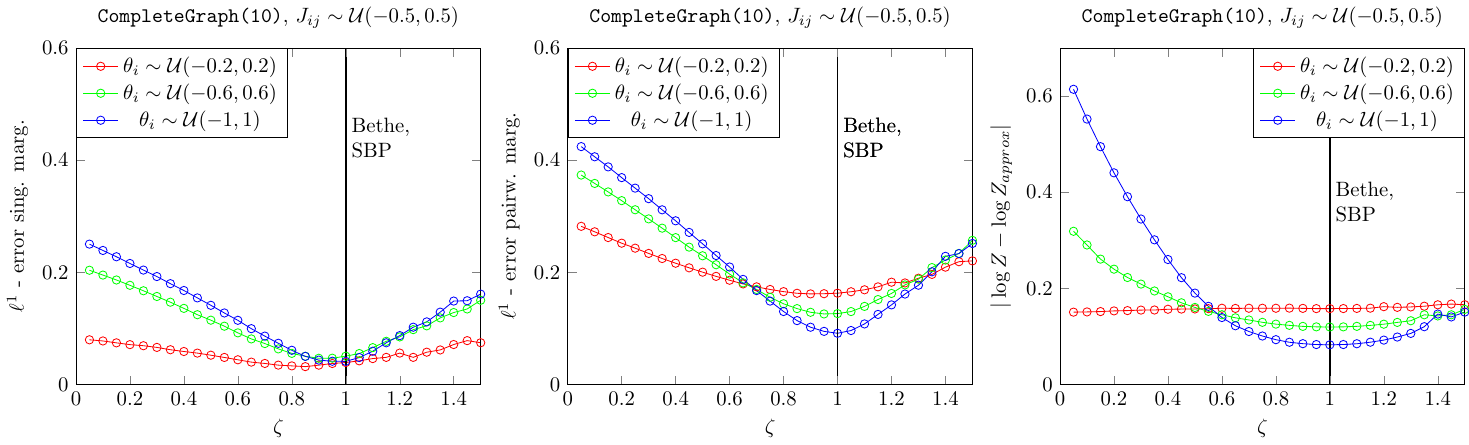}} \\
\subfigure{\includegraphics[width=0.9\linewidth]{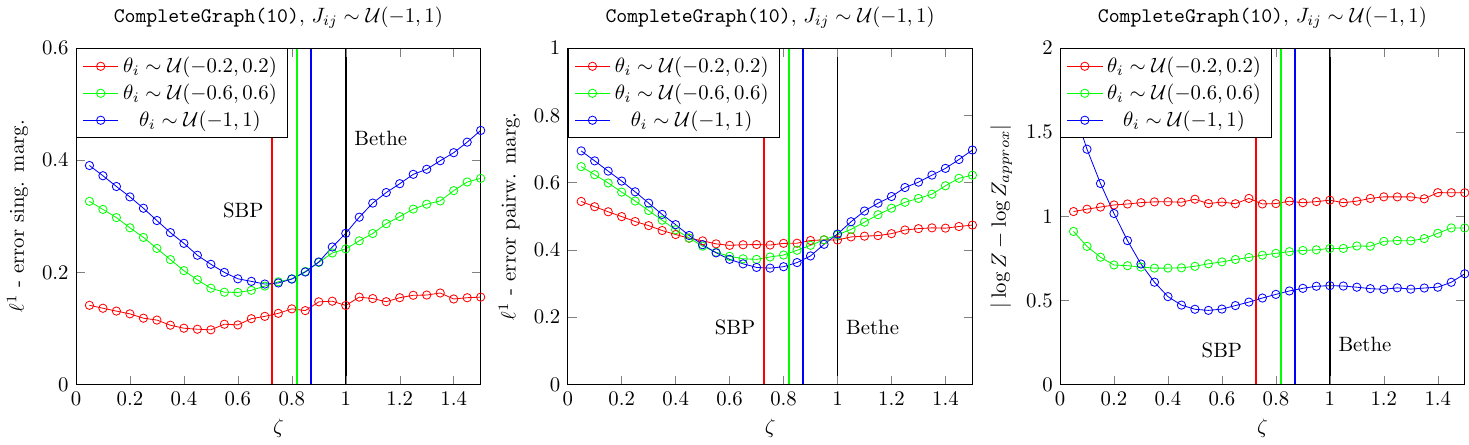}} \\
\subfigure{\includegraphics[width=0.9\linewidth]{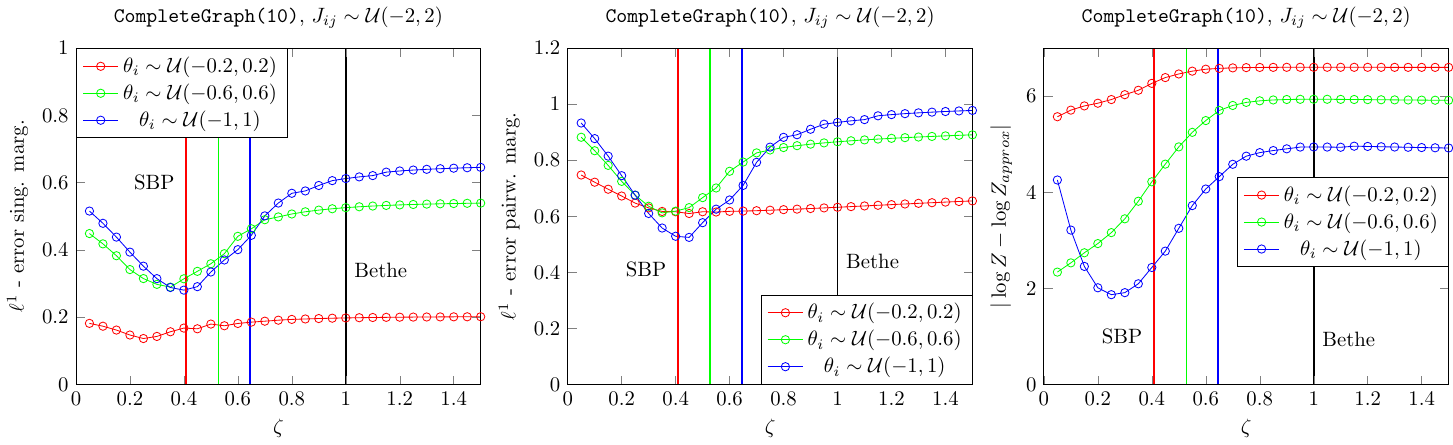}} \\
\subfigure{\includegraphics[width=0.9\linewidth]{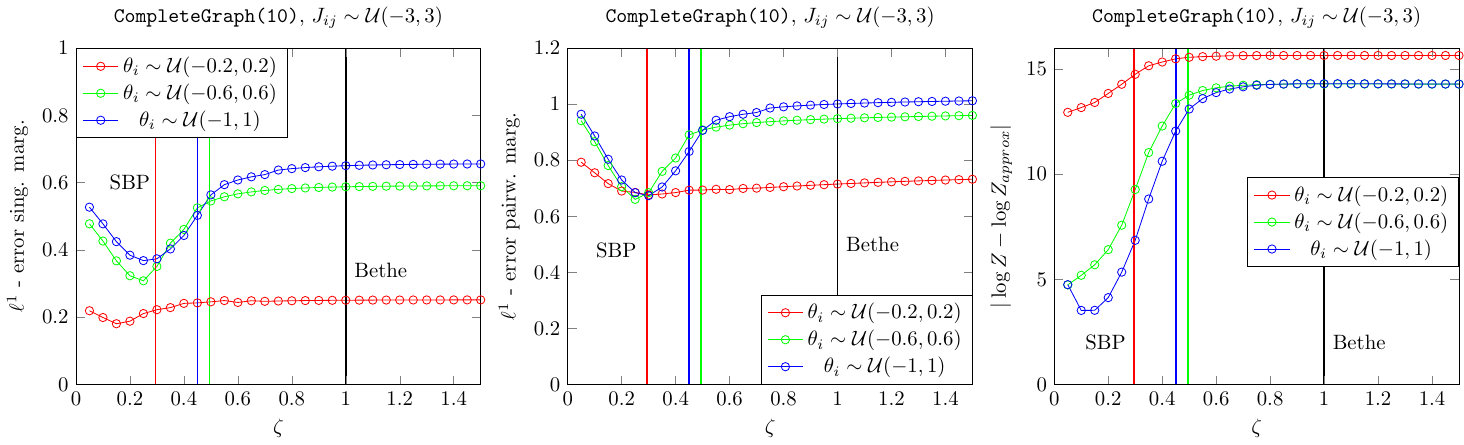}}
\caption{Approximation behavior of $\Fzeta$: Mixed models, complete graph on $10$ nodes.}
\label{fig:scaling_J_co10_gen}
\end{figure*}

\begin{figure*}[!t]
\centering
\subfigure{\includegraphics[width=0.9\linewidth]{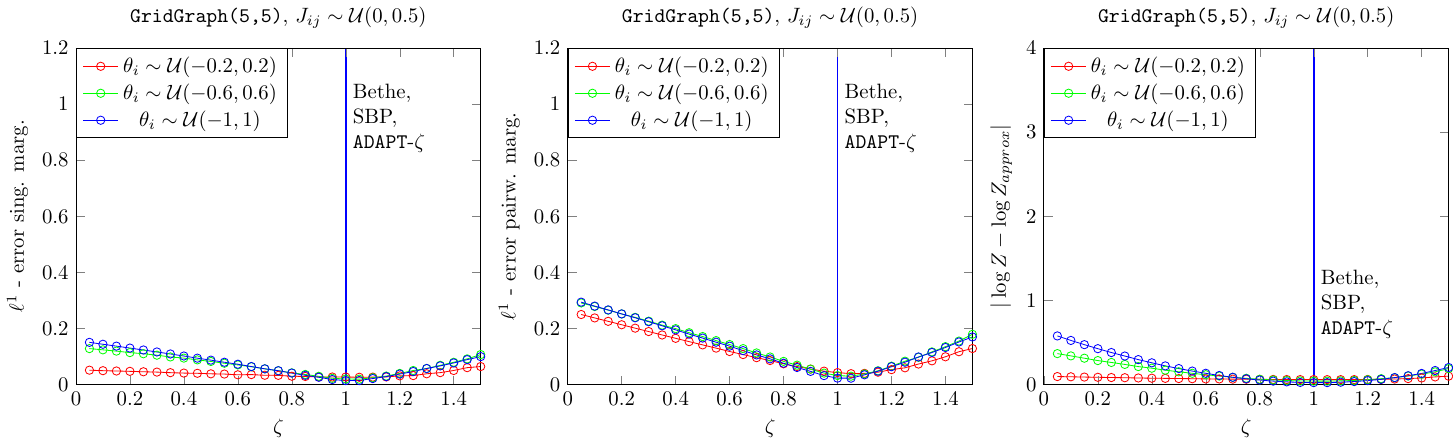}} \\
\subfigure{\includegraphics[width=0.9\linewidth]{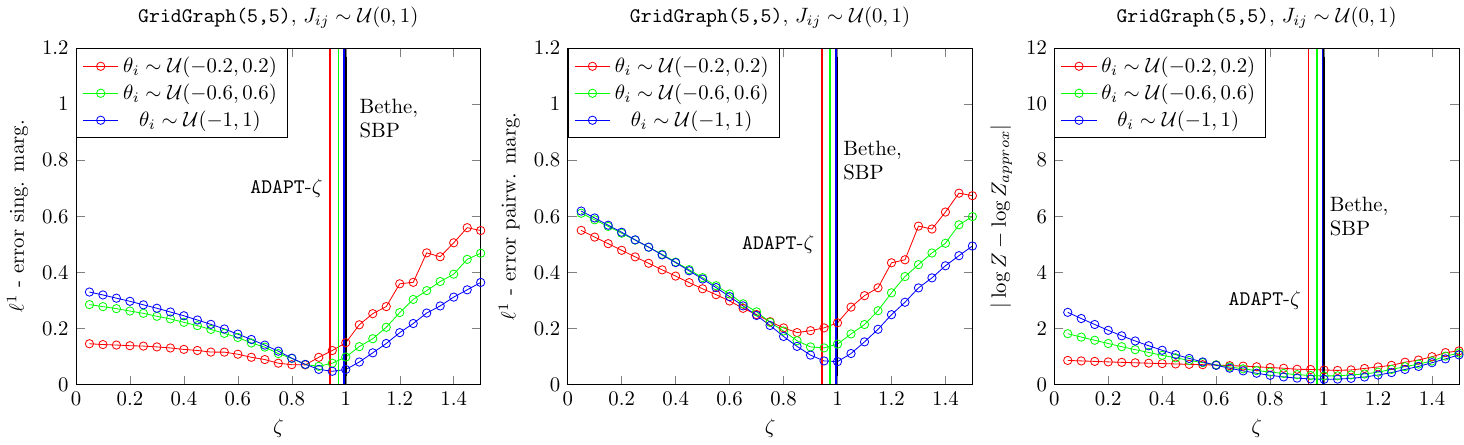}} \\
\subfigure{\includegraphics[width=0.9\linewidth]{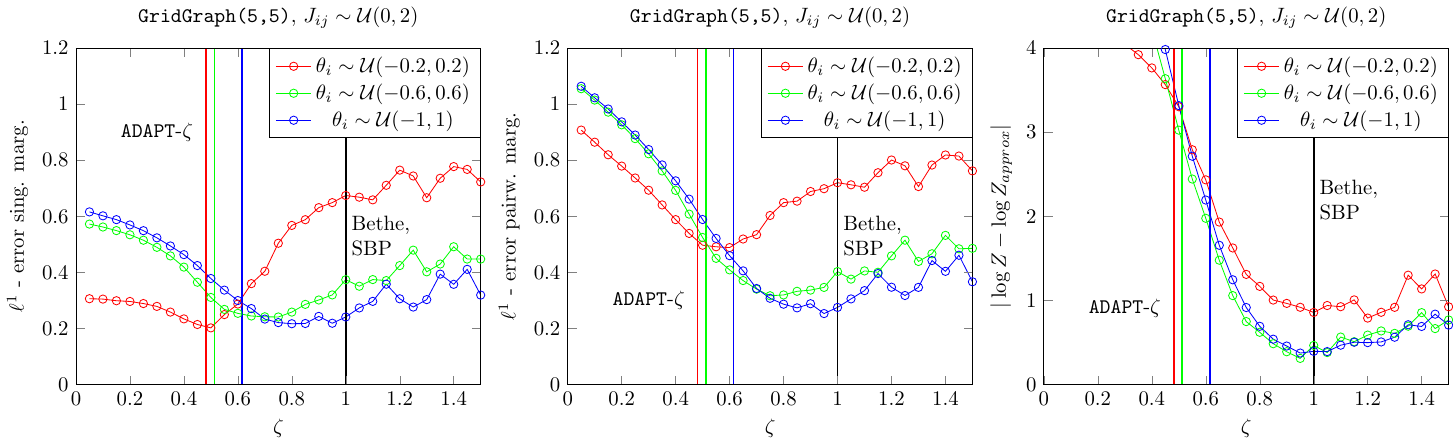}} \\
\subfigure{\includegraphics[width=0.9\linewidth]{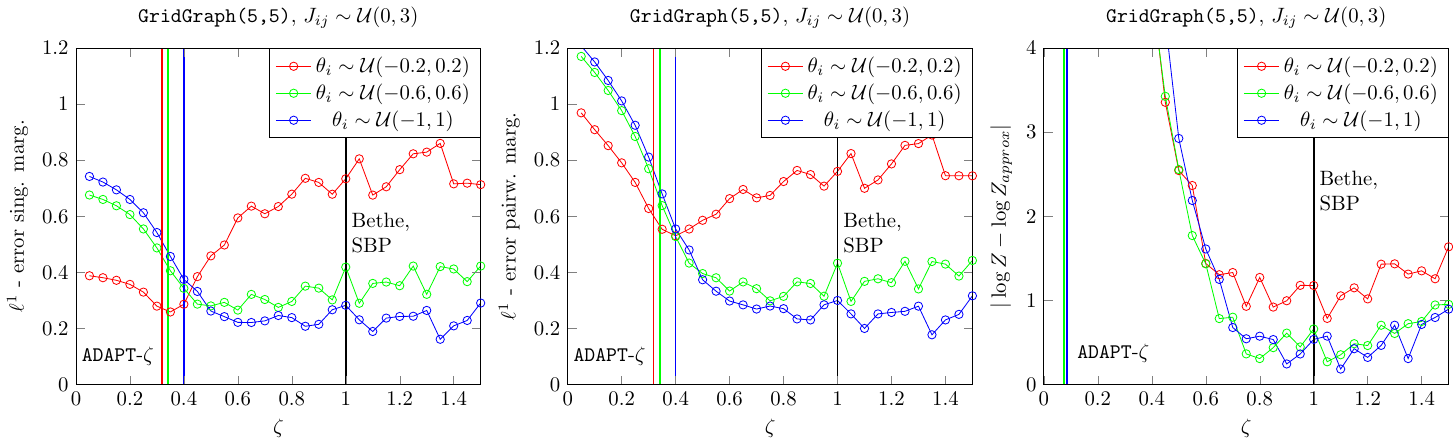}}
\caption{Approximation behavior of $\Fzeta$: Attractive models, grid graph $5 \times 5$ nodes.}
\label{fig:scaling_J_g5x5_attr}
\end{figure*}

\begin{figure*}[!t]
\centering
\subfigure{\includegraphics[width=0.9\linewidth]{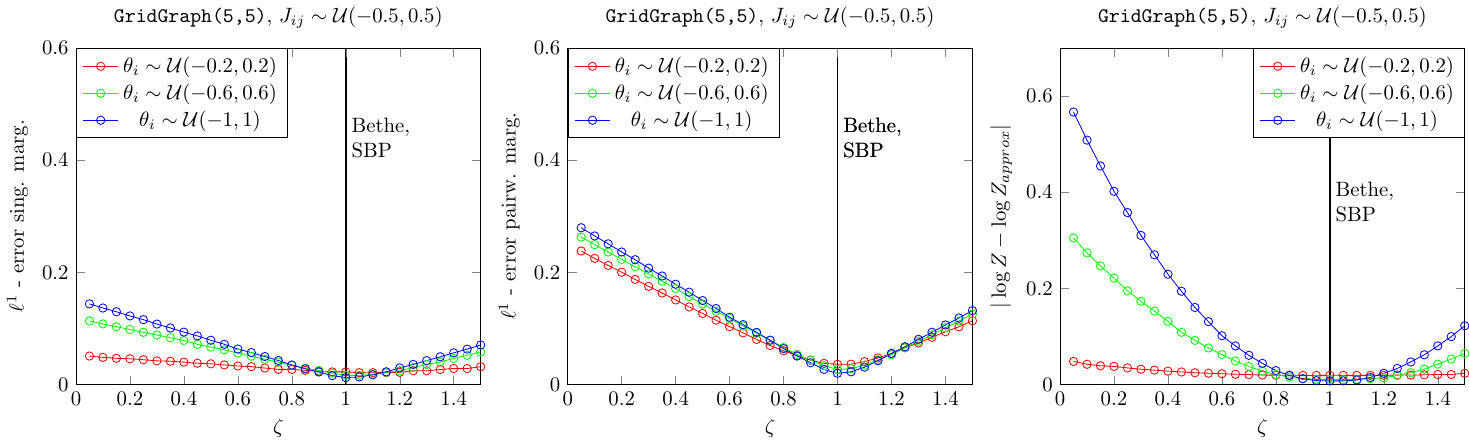}} \\
\subfigure{\includegraphics[width=0.9\linewidth]{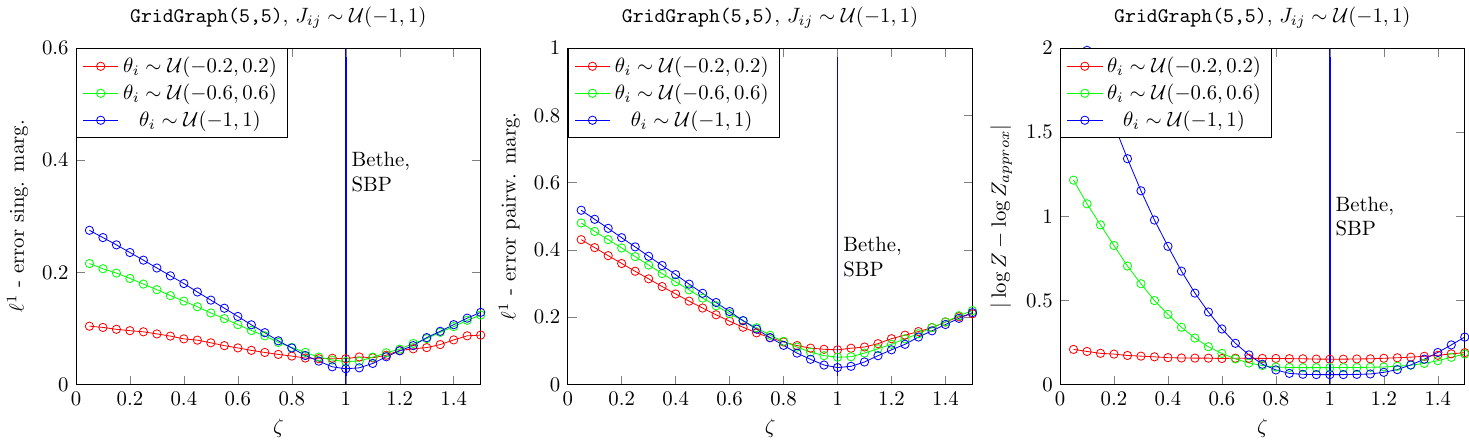}} \\
\subfigure{\includegraphics[width=0.9\linewidth]{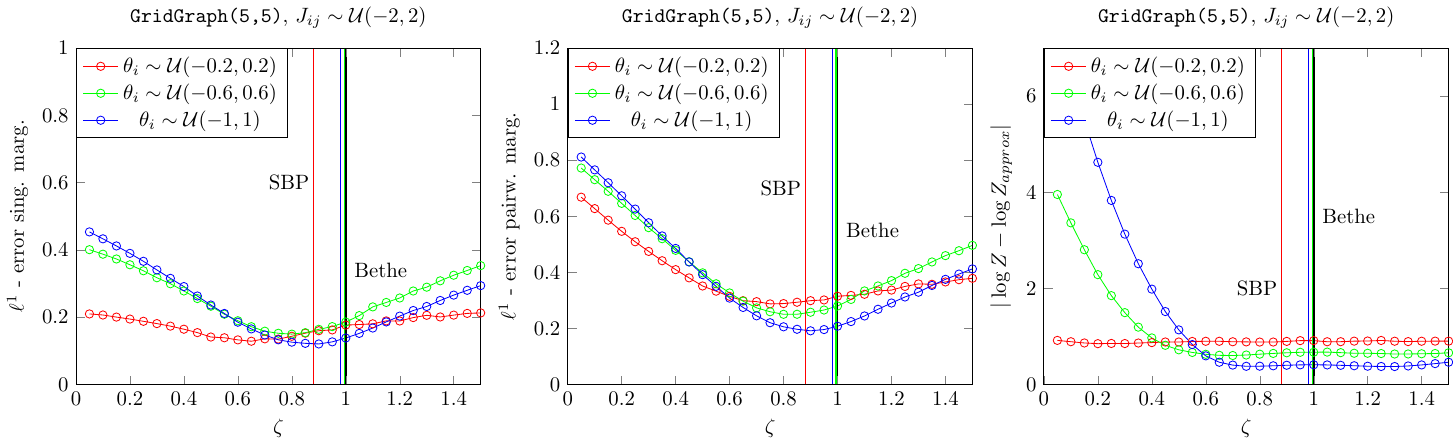}} \\
\subfigure{\includegraphics[width=0.9\linewidth]{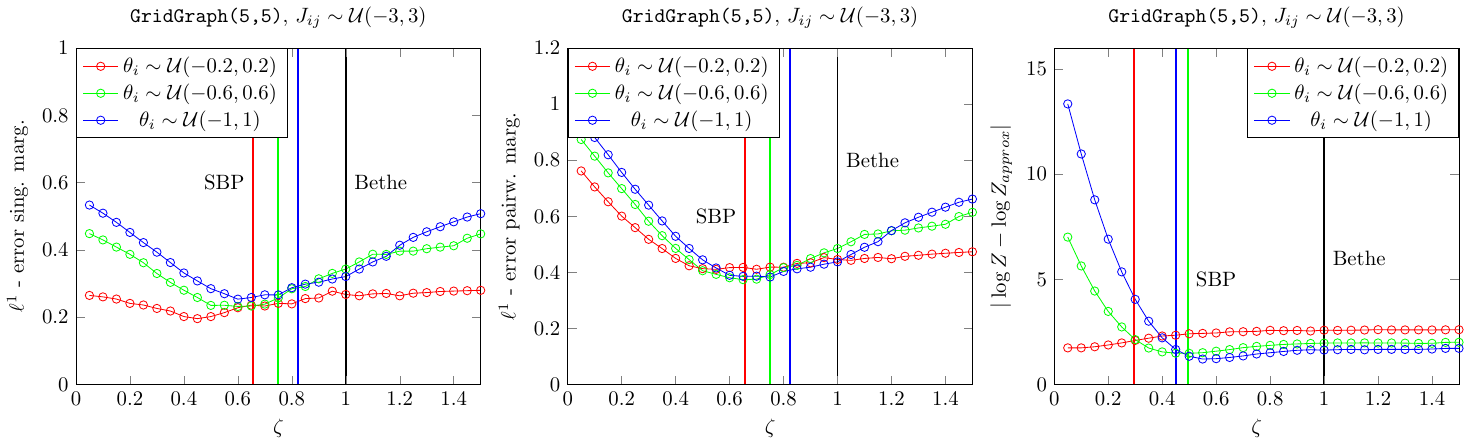}}
\caption{Approximation behavior of $\Fzeta$: Mixed models, grid graph $5 \times 5$ nodes.}
\label{fig:scaling_J_g5x5_gen}
\end{figure*}

\begin{figure*}[!t]
\centering
\subfigure{\includegraphics[width=0.9\linewidth]{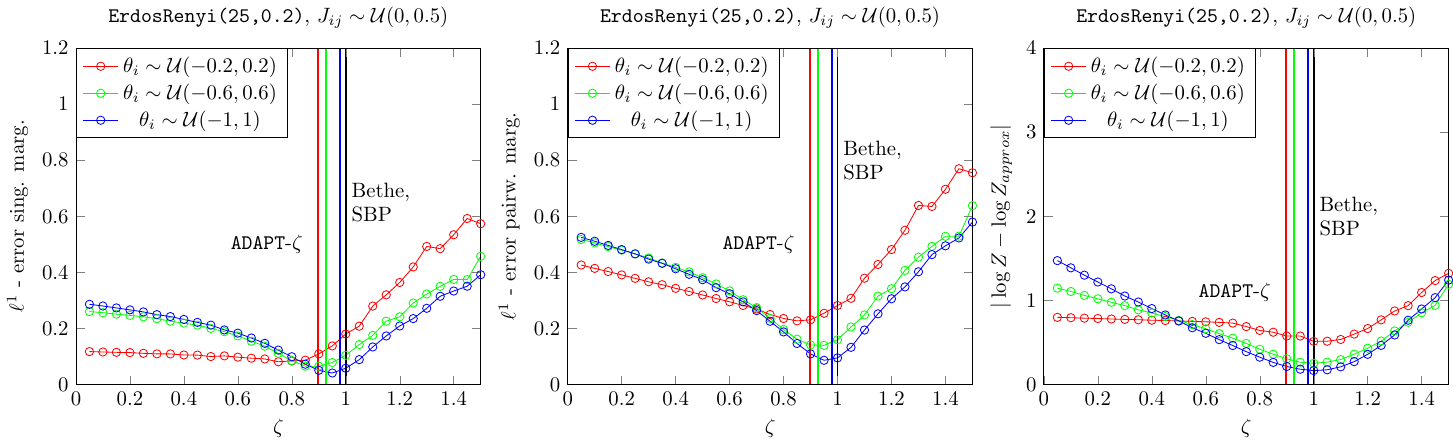}} \\
\subfigure{\includegraphics[width=0.9\linewidth]{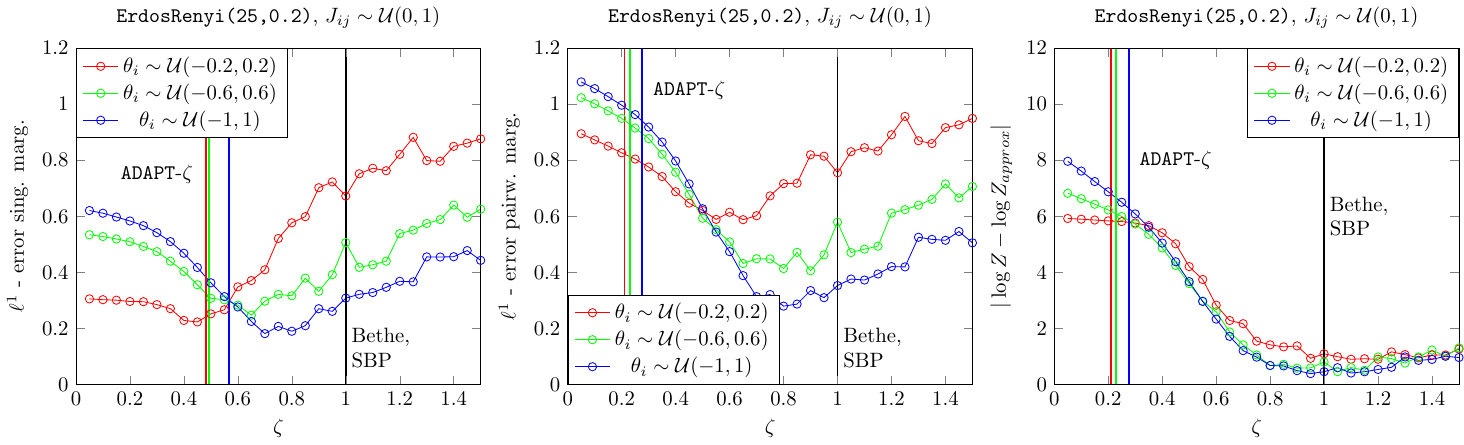}} \\
\subfigure{\includegraphics[width=0.9\linewidth]{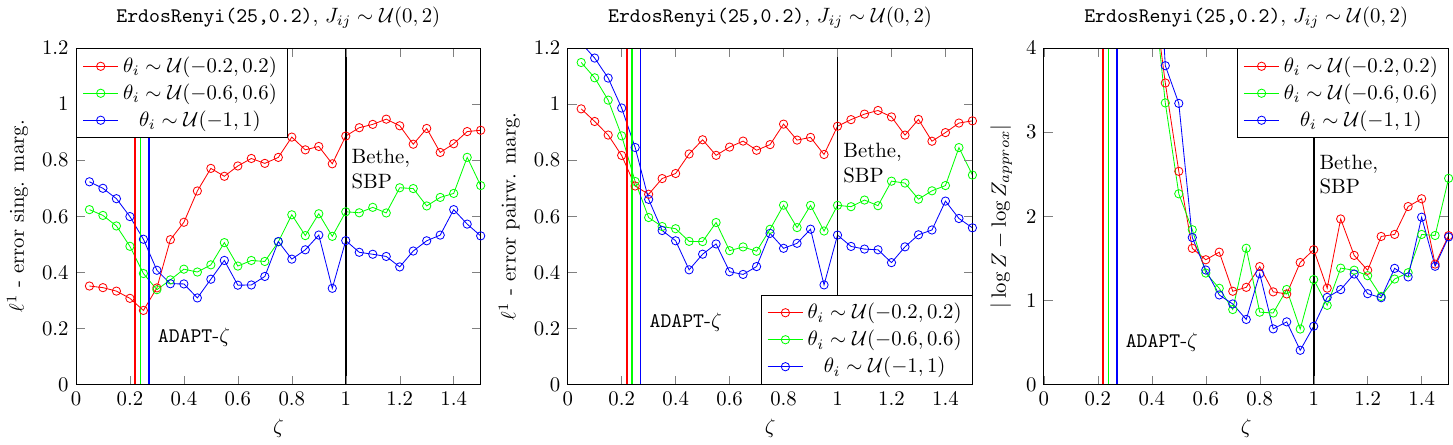}} \\
\subfigure{\includegraphics[width=0.9\linewidth]{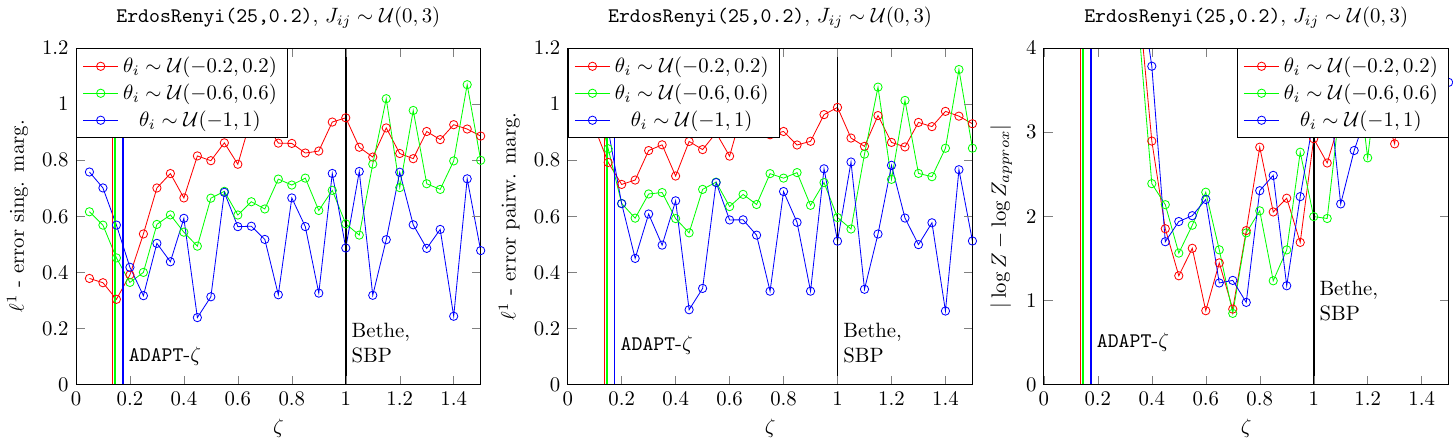}}
\caption{Approximation behavior of $\Fzeta$: Attractive models, Erdos-Renyi random graphs on $25$ nodes and an edge probability of $0.2$.}
\label{fig:scaling_J_er25x02_attr}
\end{figure*}

\begin{figure*}[!t]
\centering
\subfigure{\includegraphics[width=0.9\linewidth]{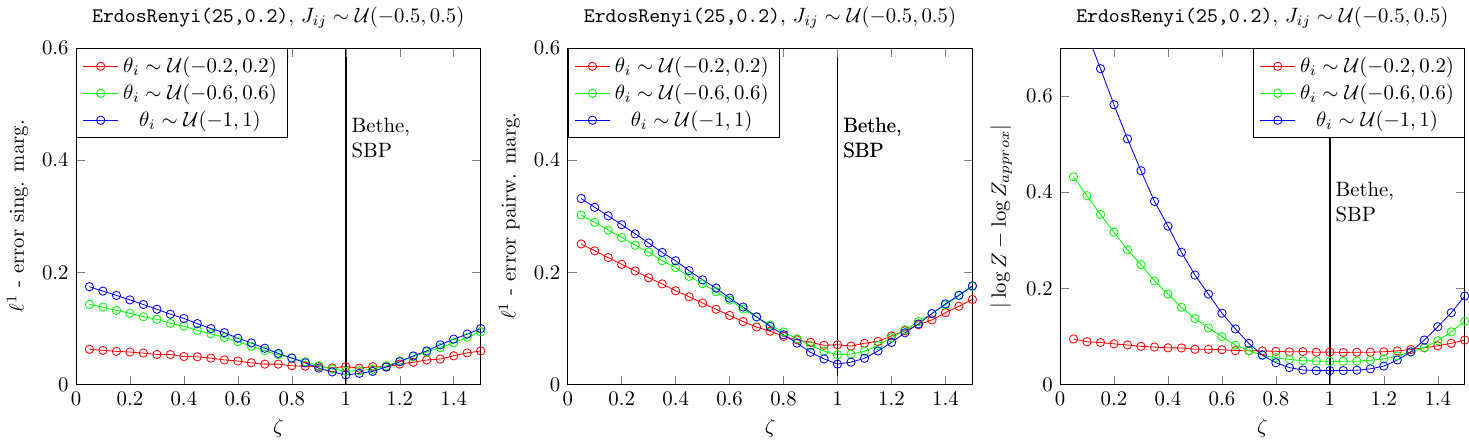}} \\
\subfigure{\includegraphics[width=0.9\linewidth]{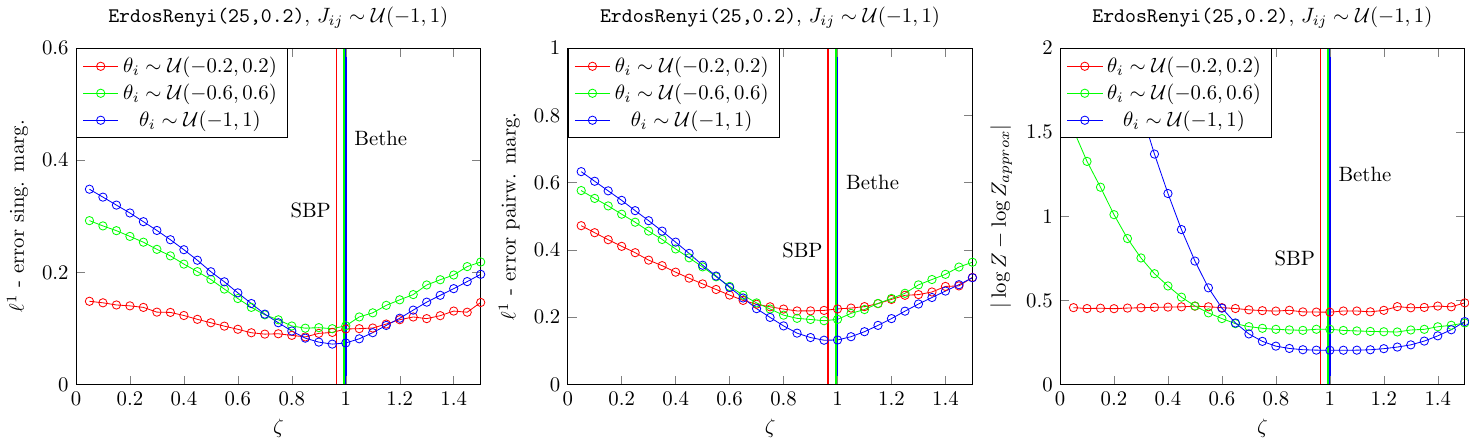}} \\
\subfigure{\includegraphics[width=0.9\linewidth]{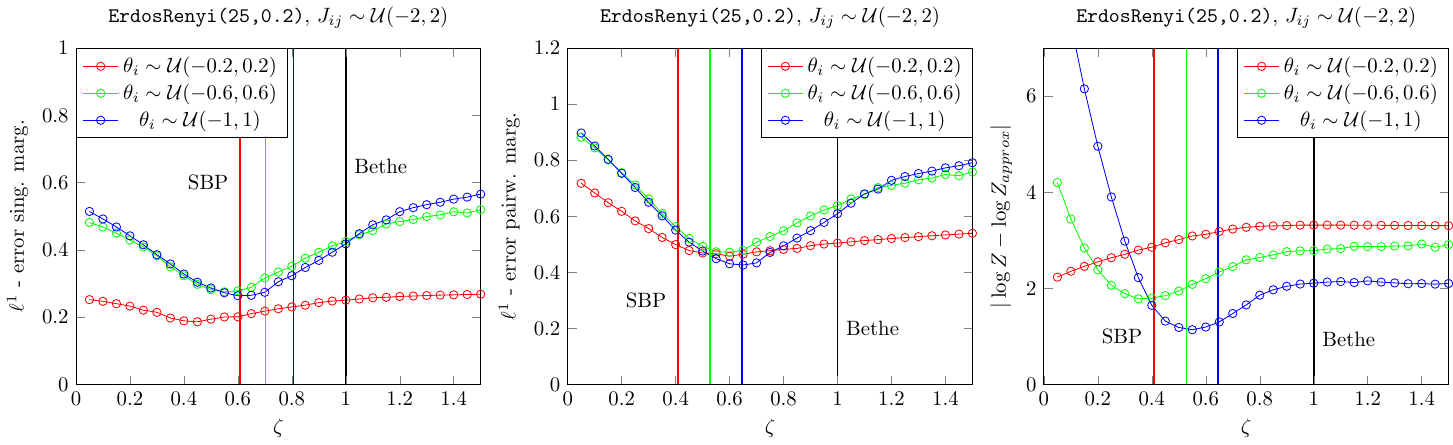}} \\
\subfigure{\includegraphics[width=0.9\linewidth]{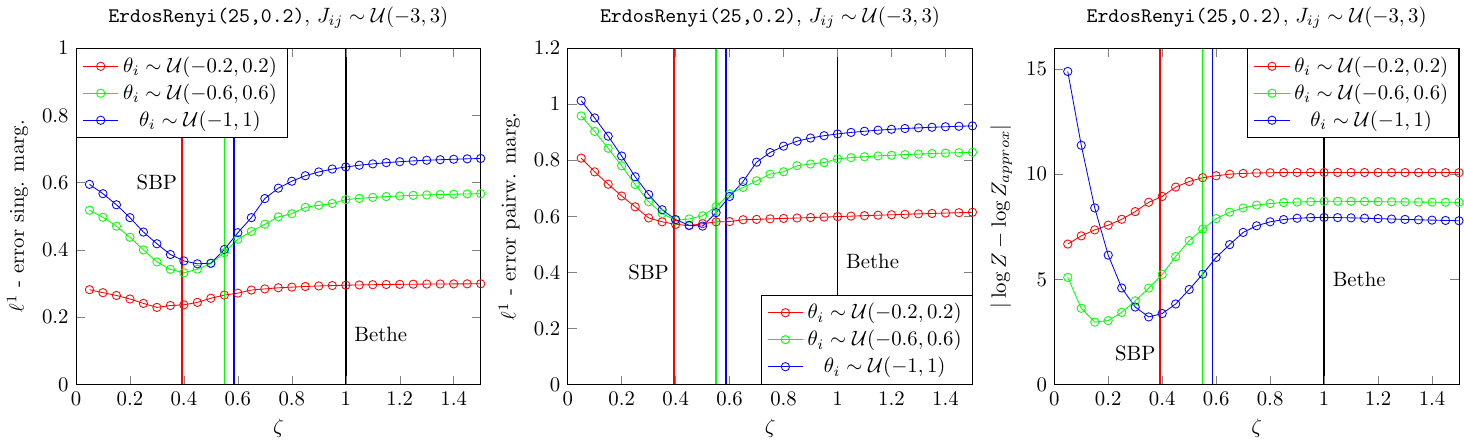}}
\caption{Approximation behavior of $\Fzeta$: Mixed models, Erdos-Renyi random graphs on $25$ nodes and an edge probability of $0.2$.}
\label{fig:scaling_J_er25x02_gen}
\end{figure*}

\section{Additional experiments}

In this section we present additional experiments in which we compare the inference algorithms introduced in the main paper to each other. As in Sec.~\ref{sec:additional_analysis_Fc_Fzeta} of this Appendix, we present additional experiments on the complete graph on $10$ vertices, on a grid graph on $5 \times 5$ vertices, and Erdos-Renyi random graphs~\citep{erdos1959random} on $25$ nodes and an edge probability of $0.2$. The experimental setup is the same as in Sec. 4 of the main paper; in particular, in particular, are averaged over $30 - 100$ individual models for each specific configuration of the potentials. In Sec.~\ref{sec:additional_experiments_attractive} we present results on attractive models, in Sec.~\ref{sec:additional_experiments_general} we present results on mixed models.

\subsection{Additional experiments for attractive models} \label{sec:additional_experiments_attractive}
In Fig. 13-15 we present additional experiments for attractive models. The results are similar as in the main paper; however, adapt-$\zeta$ loses some of its benefit in the Erdos-Renyi random graphs in which it is slightly outperformed by the LS-Convex free energies.

\begin{figure*}[]
\centering
\subfigure{\includegraphics[width=0.9\linewidth]{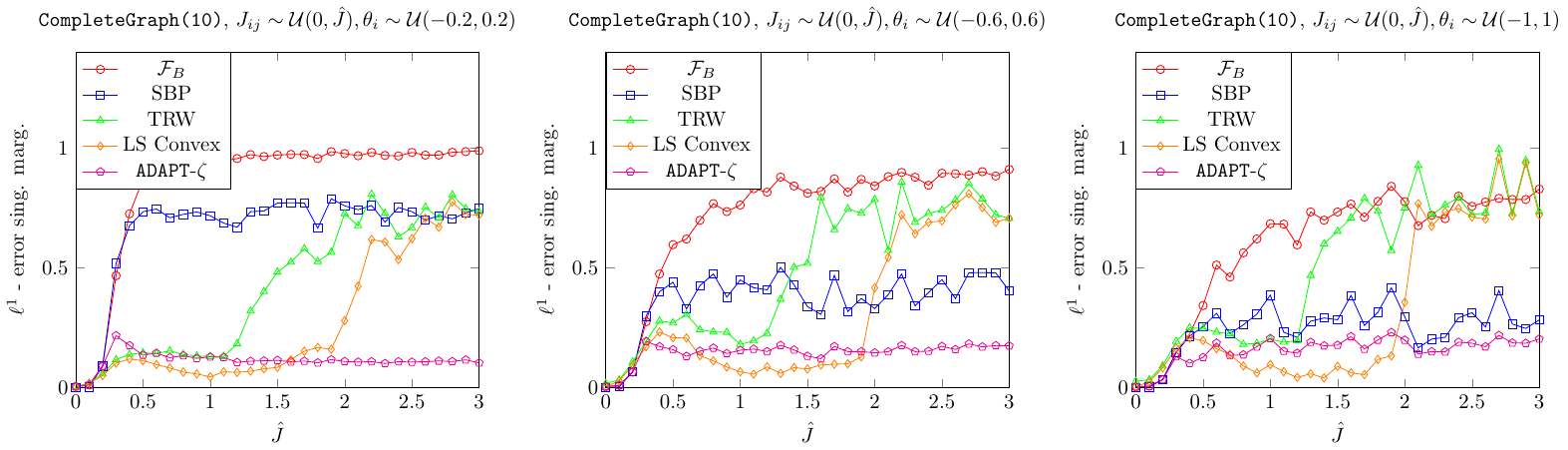}} \\
\subfigure{\includegraphics[width=0.9\linewidth]{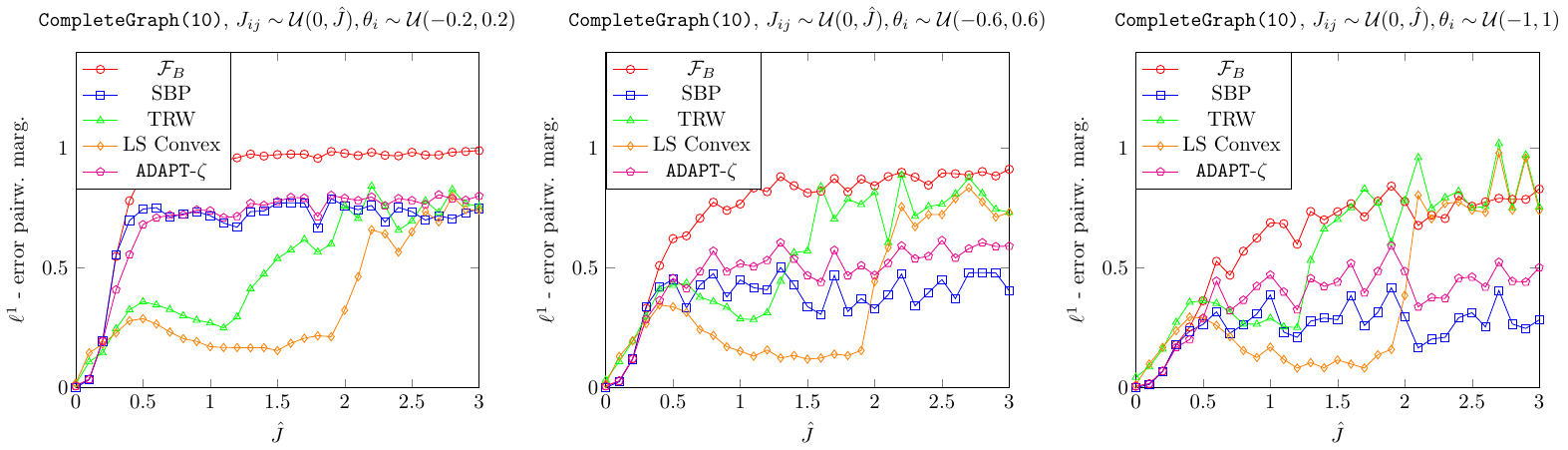}} \\
\subfigure{\includegraphics[width=0.9\linewidth]{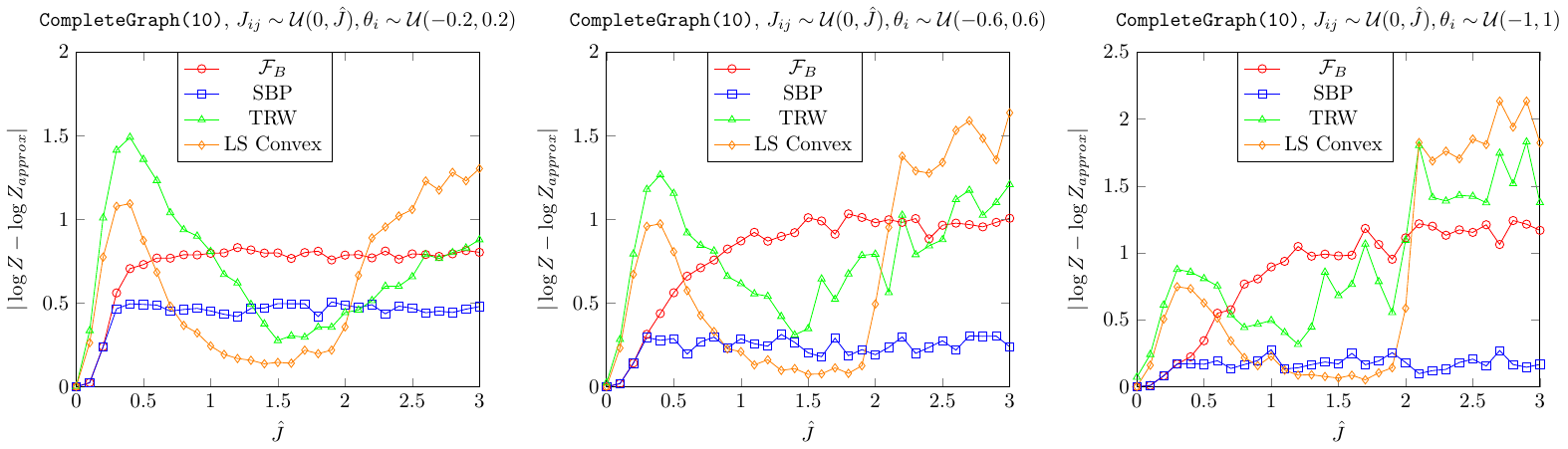}}
\caption{Algorithms Bethe ($\FB$), SBP, TRW, LS-Convex, and \texttt{ADAPT}-$\zeta$ compared on attractive models on a complete graph on $10$ nodes. First row: $l^1$- error on singleton marginals; second row: $l^1$- error on pairwise marginals; third row: absolute error on log-partition function.}
\label{fig:experiments_attractive_co10}
\end{figure*}
\newpage
\begin{figure*}[]
\centering
\subfigure{\includegraphics[width=0.9\linewidth]{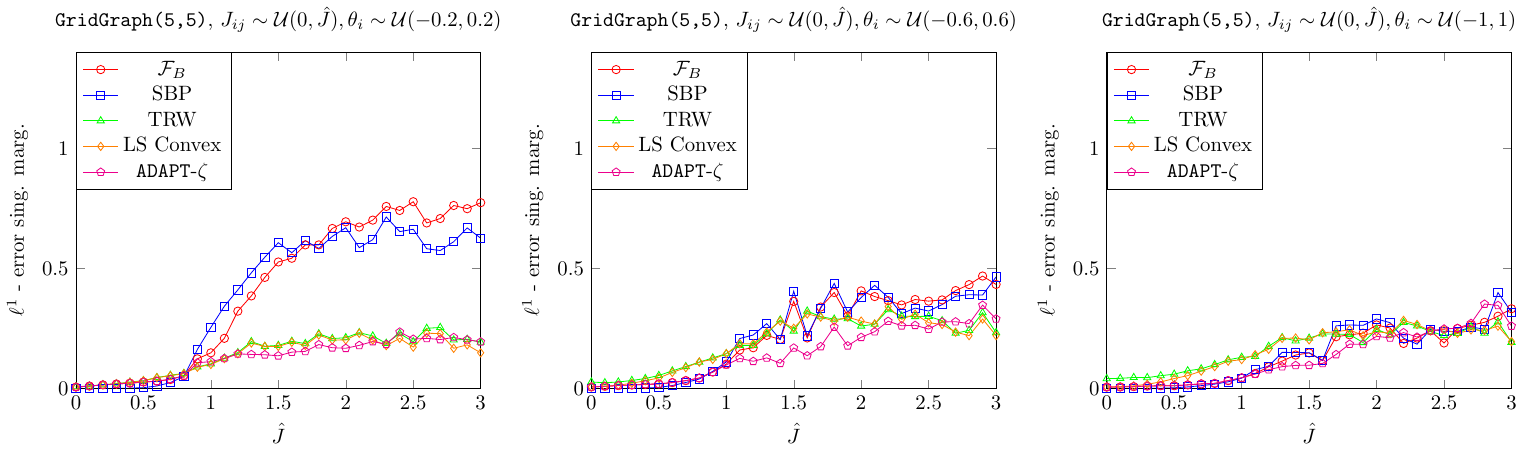}} \\
\subfigure{\includegraphics[width=0.9\linewidth]{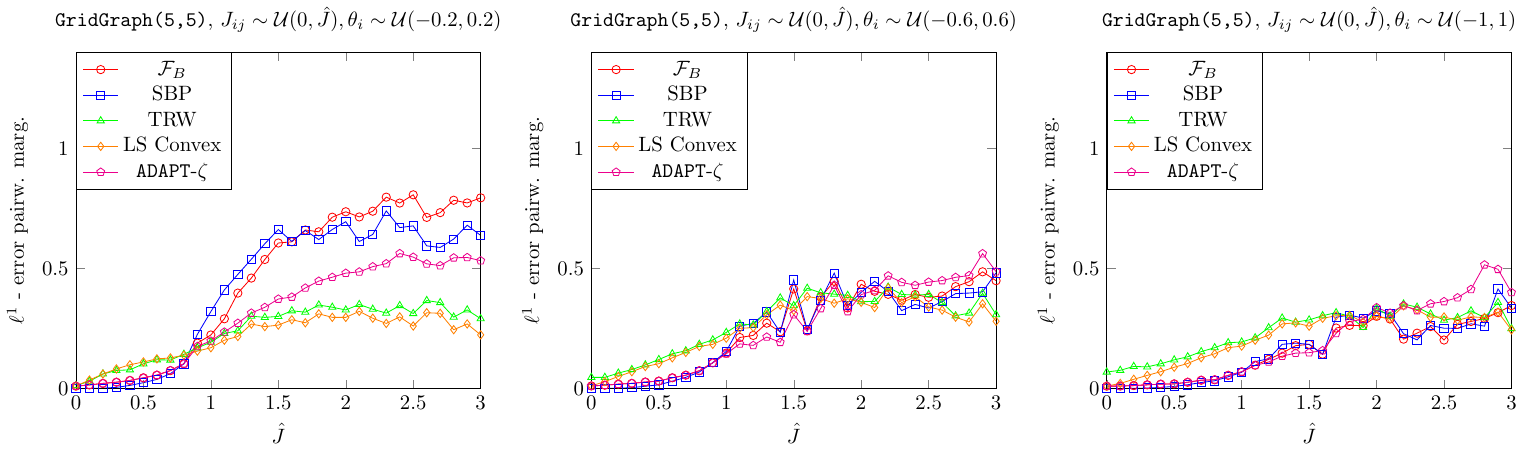}} \\
\subfigure{\includegraphics[width=0.9\linewidth]{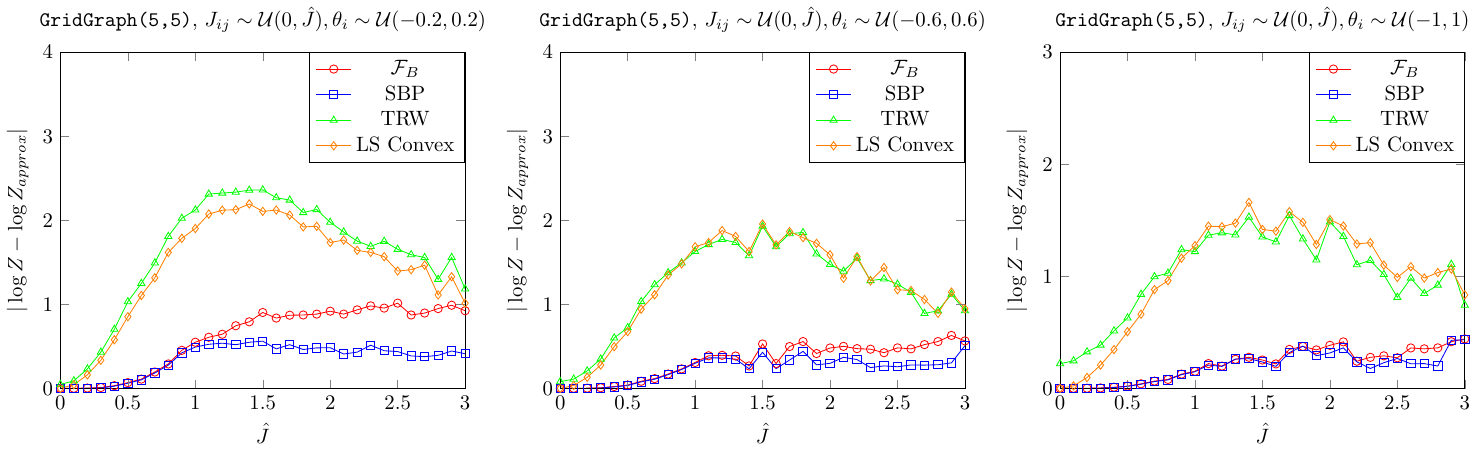}}
\caption{Algorithms Bethe ($\FB$), SBP, TRW, LS-Convex, and \texttt{ADAPT}-$\zeta$ compared on attractive models on a grid graph on $5 \times 5$ nodes. First row: $l^1$- error on singleton marginals; second row: $l^1$- error on pairwise marginals; third row: absolute error on log-partition function.}
\label{fig:experiments_attractive_g5x5}
\end{figure*}
\newpage
\begin{figure*}[]
\centering
\subfigure{\includegraphics[width=0.9\linewidth]{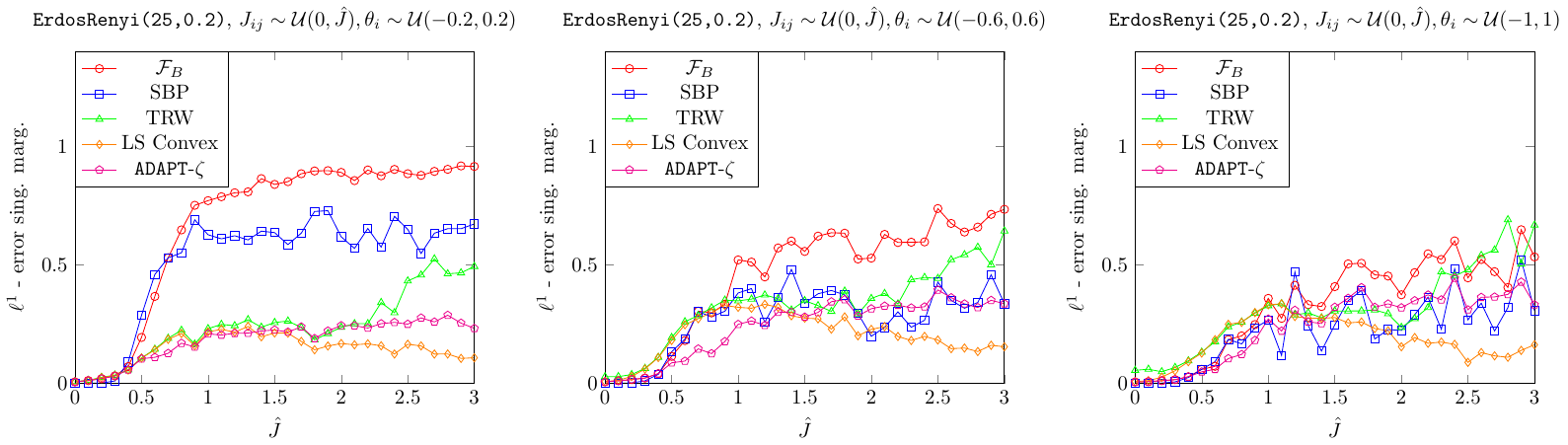}} \\
\subfigure{\includegraphics[width=0.9\linewidth]{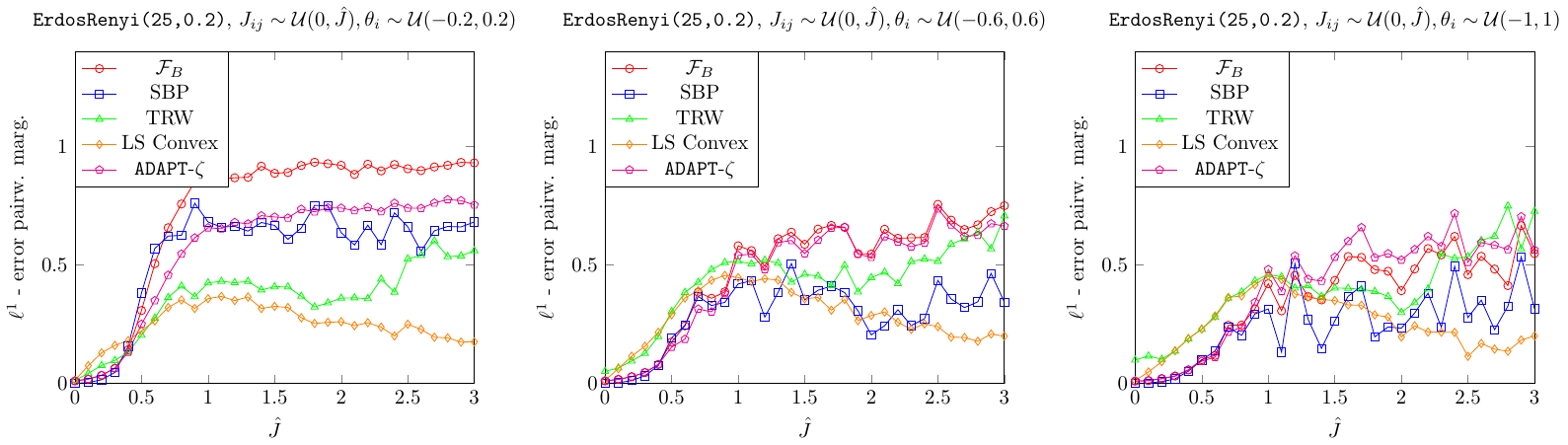}} \\
\subfigure{\includegraphics[width=0.9\linewidth]{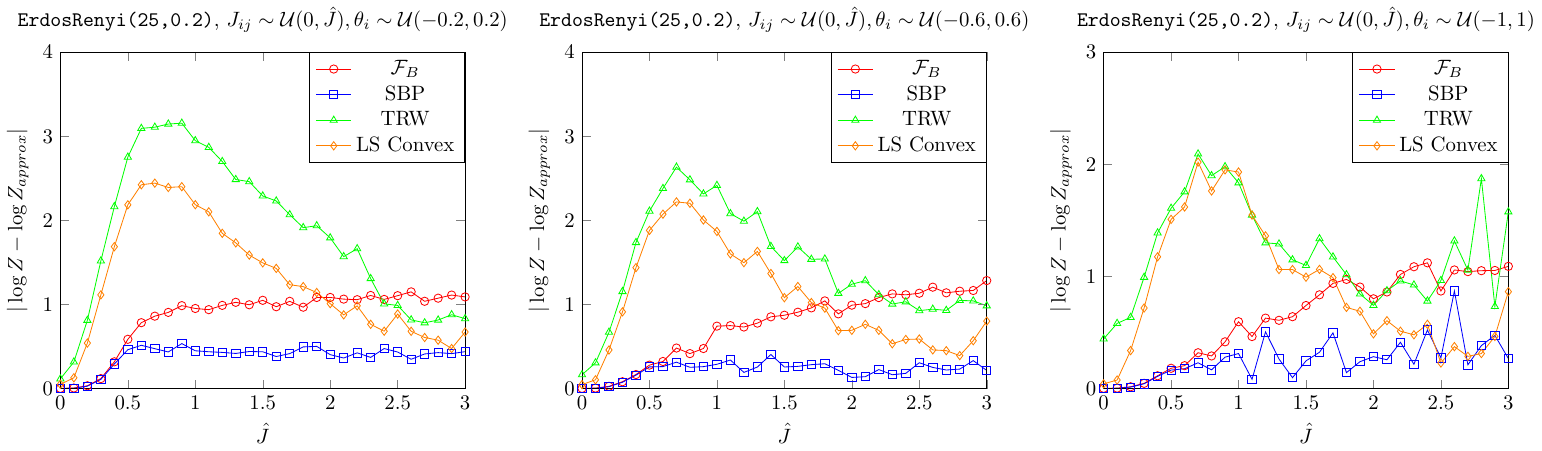}}
\caption{Algorithms Bethe ($\FB$), SBP, TRW, LS-Convex, and \texttt{ADAPT}-$\zeta$ compared on attractive models on Erdos renyi random graphs on $25$ nodes and an edge probability of $0.2$. First row: $l^1$- error on singleton marginals; second row: $l^1$- error on pairwise marginals; third row: absolute error on log-partition function.}
\label{fig:experiments_attractive_er25x02}
\end{figure*}
\newpage
\subsection{Additional experiments for mixed models} \label{sec:additional_experiments_general}
In Fig. 16-18 we present additional experiments for mixed models. The results are similar as in the main paper; in particular, \texttt{ADAPT}-$c$ is still by far superior to other methods if one aims to estimate the partition function.

\begin{figure*}[]
\centering
\subfigure{\includegraphics[width=0.9\linewidth]{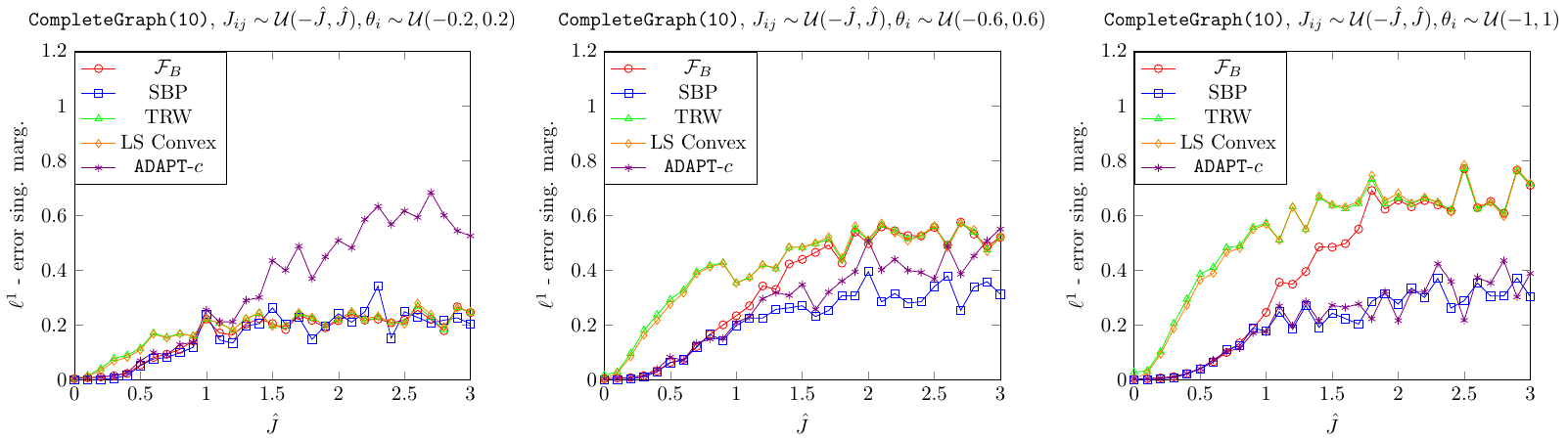}} \\
\subfigure{\includegraphics[width=0.9\linewidth]{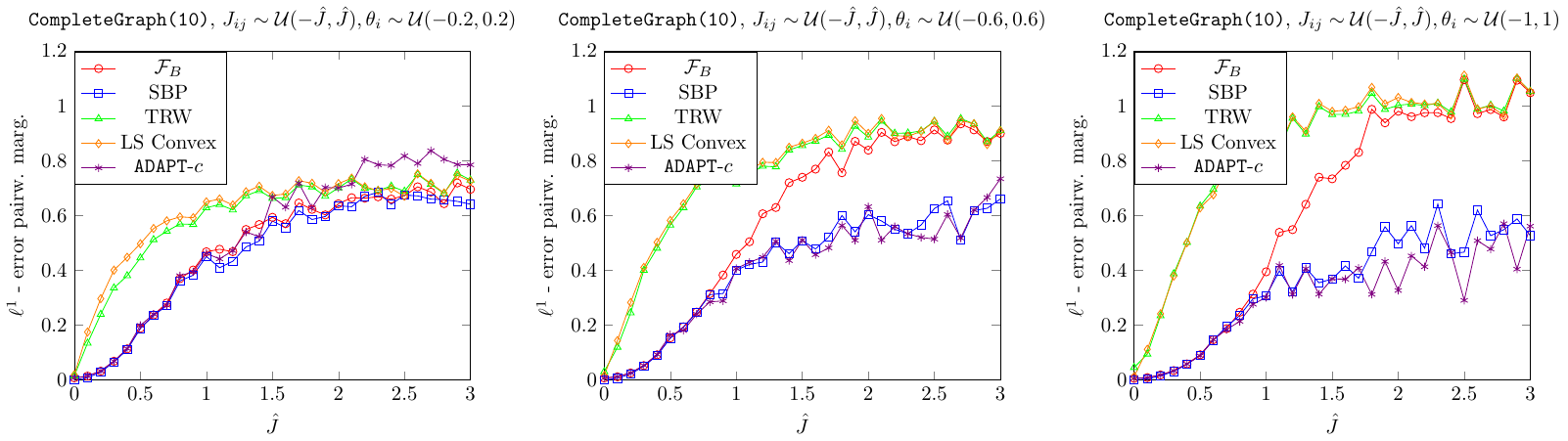}} \\
\subfigure{\includegraphics[width=0.9\linewidth]{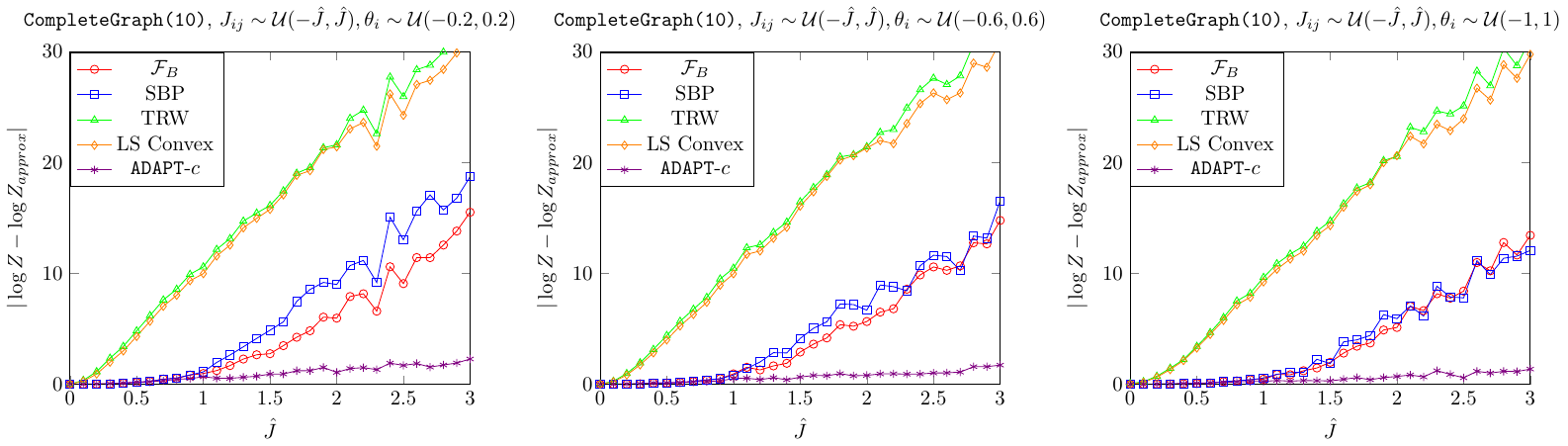}}
\caption{Algorithms Bethe ($\FB$), SBP, TRW, LS-Convex, and \texttt{ADAPT}-$c$ compared on mixed models on a complete graph on $10$ nodes. First row: $l^1$- error on singleton marginals; second row: $l^1$- error on pairwise marginals; third row: absolute error on log-partition function.}
\label{fig:experiments_general_co10}
\end{figure*}
\newpage
\begin{figure*}[]
\centering
\subfigure{\includegraphics[width=0.9\linewidth]{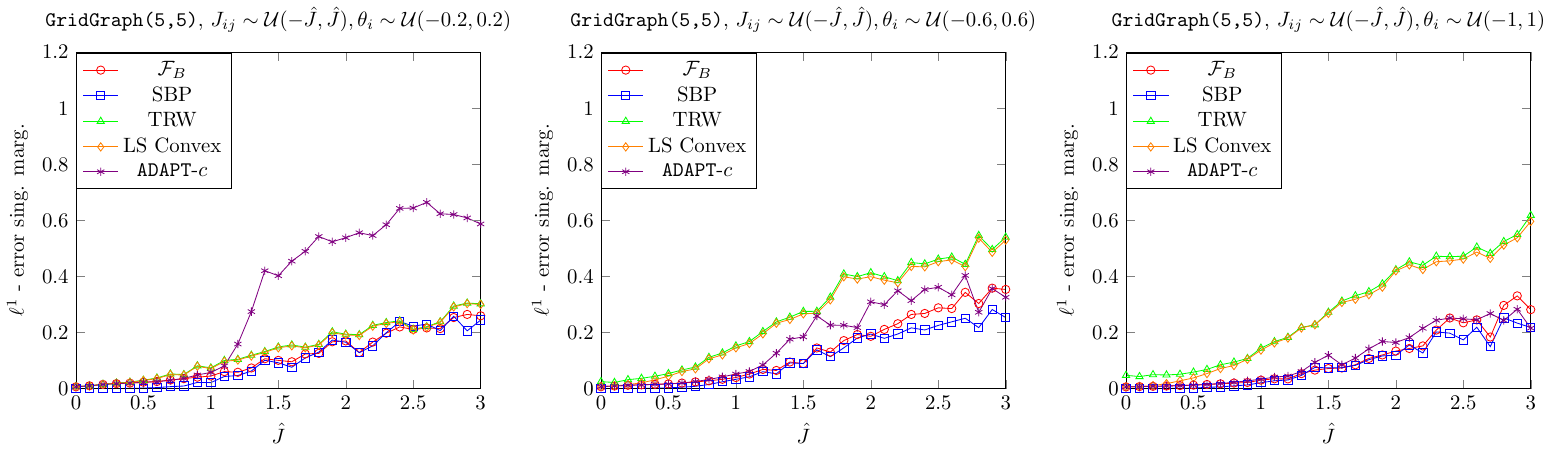}} \\
\subfigure{\includegraphics[width=0.9\linewidth]{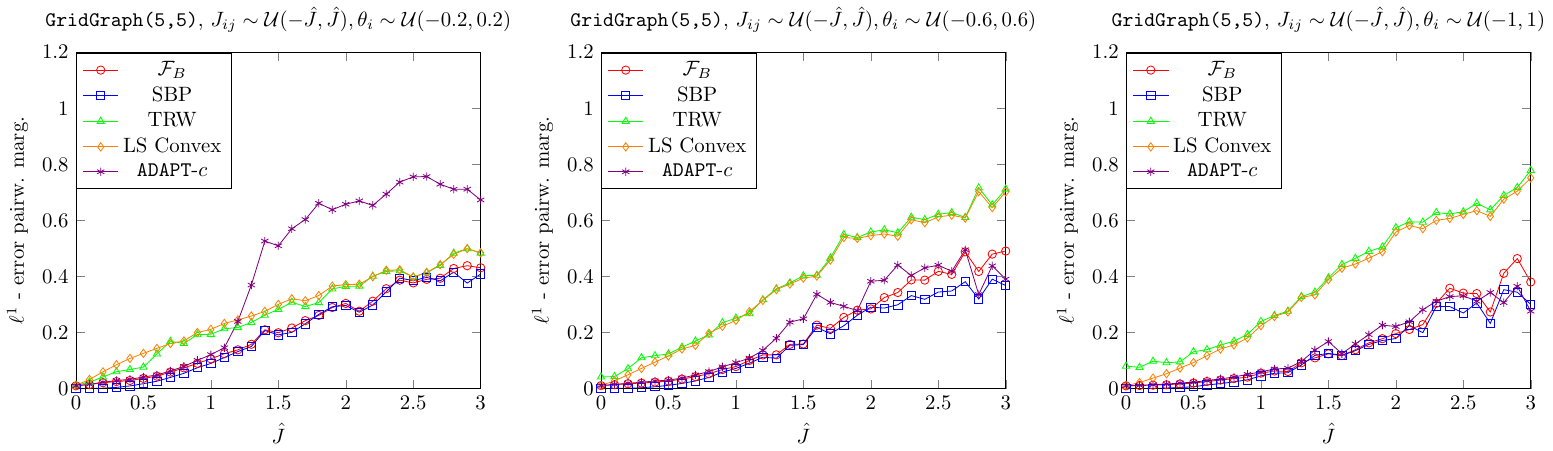}} \\
\subfigure{\includegraphics[width=0.9\linewidth]{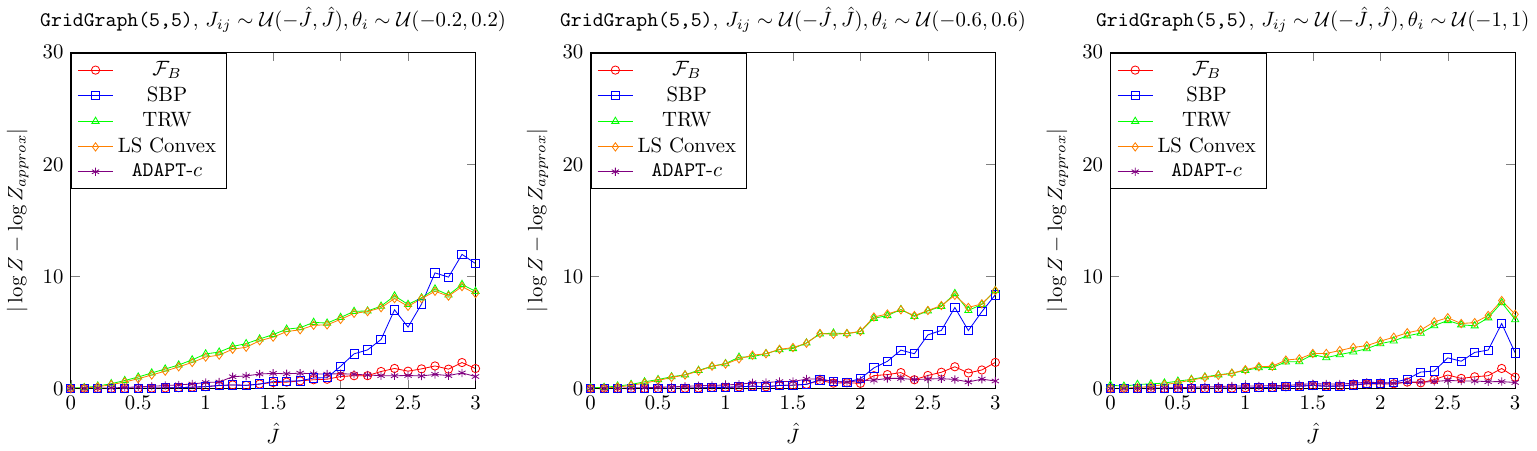}}
\caption{Algorithms Bethe ($\FB$), SBP, TRW, LS-Convex, and \texttt{ADAPT}-$c$ compared on mixed models on a grid graph on $5 \times 5$ nodes. First row: $l^1$- error on singleton marginals; second row: $l^1$- error on pairwise marginals; third row: absolute error on log-partition function.}
\label{fig:experiments_general_g5x5}
\end{figure*}
\newpage
\begin{figure*}[]
\centering
\subfigure{\includegraphics[width=0.9\linewidth]{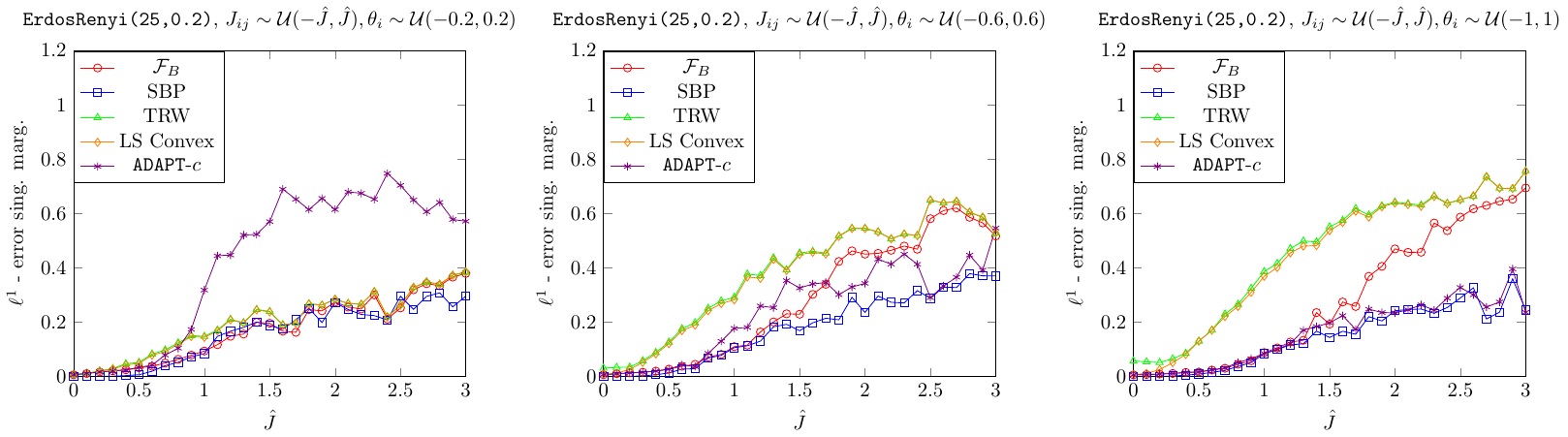}} \\
\subfigure{\includegraphics[width=0.9\linewidth]{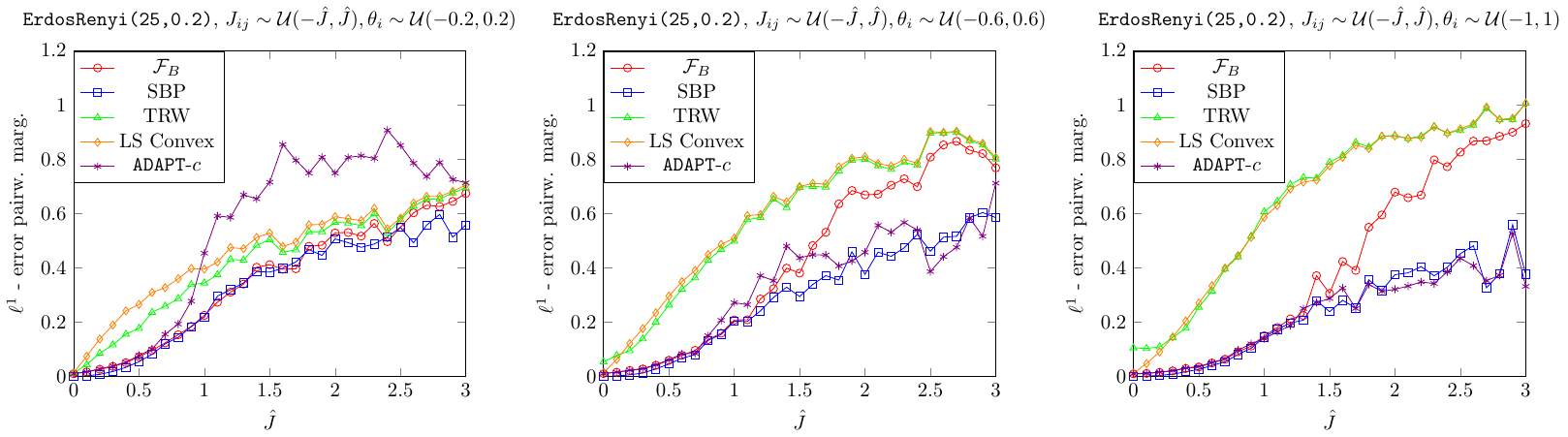}} \\
\subfigure{\includegraphics[width=0.9\linewidth]{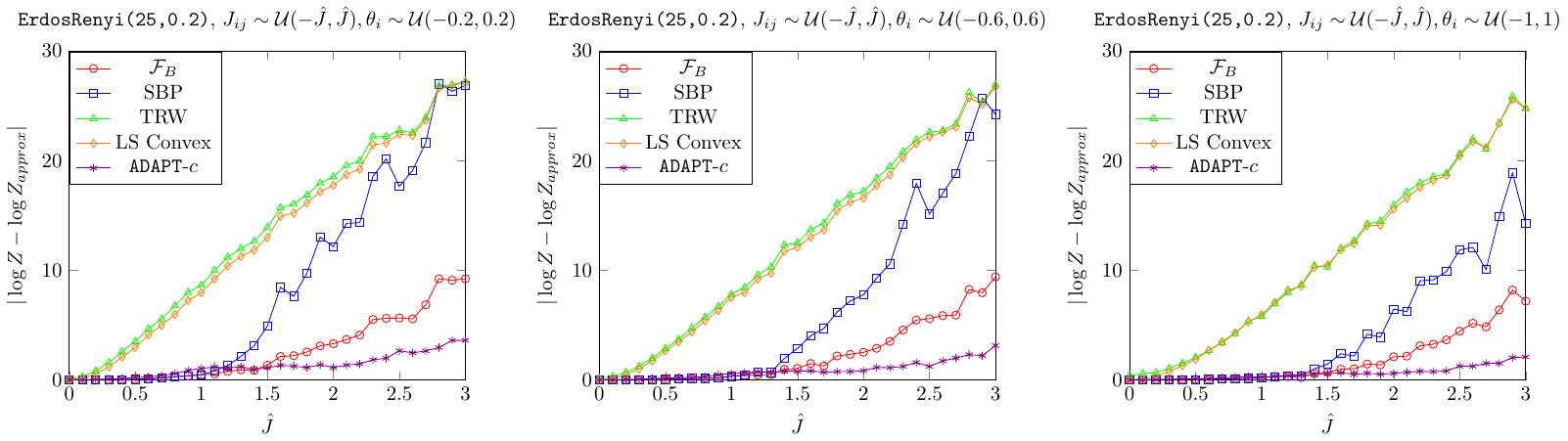}}
\caption{Algorithms Bethe ($\FB$), SBP, TRW, LS-Convex, and \texttt{ADAPT}-$c$ compared on mixed models on Erdos renyi random graphs on $25$ nodes and an edge probability of $0.2$. First row: $l^1$- error on singleton marginals; second row: $l^1$- error on pairwise marginals; third row: absolute error on log-partition function.}
\label{fig:experiments_general_er25x02}
\end{figure*}
\newpage





\newpage


\clearpage

\bibliography{Adaptive_arxiv}

\end{document}